\documentclass[]{bytedance_seed}
\usepackage[utf8]{inputenc}
\usepackage{graphicx} % Required for inserting images
\usepackage{booktabs}
\usepackage{caption}
\usepackage{float}
\usepackage{multirow}
\usepackage{multicol}
\usepackage{makecell}
\usepackage{amsmath}
\usepackage{natbib}
\usepackage{amssymb}
\usepackage{enumitem}
\usepackage{array} % Enhanced table features
\usepackage{siunitx} % For proper unit and number display
\usepackage{tcolorbox}
\tcbuselibrary{breakable}
\usepackage{xltabular}
\usepackage{subcaption}
\usepackage{threeparttable}
\useunder{\uline}{\ul}{}
\usepackage{twemojis}
\usepackage{float}
\usepackage{listings}
\usepackage{longtable}
\newcommand{\tablestyle}[2]{\setlength{\tabcolsep}{#1}\renewcommand{\arraystretch}{#2}\centering\footnotesize}
\usepackage{color}

\definecolor{taskblue}{RGB}{173,216,230}
\definecolor{thinkgray}{RGB}{245,245,245}
\definecolor{accentblue}{RGB}{70,130,180}
\definecolor{successgreen}{RGB}{144,238,144}
\definecolor{inputgray}{RGB}{250,250,250}
\definecolor{gold}{HTML}{D4AF37} % for IMO_CMO section
\newtcolorbox{thoughtbox}[1][]{
    colback=thinkgray,
    colframe=accentblue,
    boxrule=0.5pt,
    arc=3pt,
    left=5pt,
    right=5pt,
    top=5pt,
    bottom=5pt,
    fonttitle=\bfseries,
    title=#1
}
\newtcolorbox{taskbox}{
    colback=taskblue!30,
    colframe=accentblue,
    boxrule=1pt,
    arc=3pt,
    left=8pt,
    right=8pt,
    top=8pt,
    bottom=8pt
}

\lstdefinestyle{inputstyle}{
    backgroundcolor=\color{inputgray},
    basicstyle=\ttfamily\small,
    breaklines=true,
    frame=single,
    framerule=0.5pt,
    rulecolor=\color{accentblue!30},
    xleftmargin=10pt,
    xrightmargin=10pt,
    aboveskip=8pt,
    belowskip=8pt,
    columns=flexible,
    keepspaces=true,
    literate={≤}{<=}1 {≥}{>=}1 {≠}{!=}1 {∞}{inf}3 {∑}{sum}3 {∈}{in}2 {ℝ}{R}1 {²}{^2}1
}

\usepackage{hyperref}
\title{Seed2.0 Model Card: \\ Towards Intelligence Frontier for Real-World Complexity}
\author{Bytedance Seed}

\begin{document}

\maketitle

\section{Introduction}

Large Language Models (LLMs) now play a central role in modern digital infrastructure. 
Usage has grown dramatically across both professional and personal contexts \citep{NBERw34255}. 
The Seed team has developed a comprehensive model family that includes general-purpose LLMs, multimodal models, open-source releases, code-specialized models, diffusion-based language modeling, formal theorem proving, and generative media systems: Seed1.6/1.8, Seed1.5-VL, Seed-OSS, Seed-Coder, Seed Diffusion, Seed-Prover, and Seedream/Seedance \citep{seed16_page,seed16_techniques,seed18_modelcard,seedoss_release,seedprover15_blog,seedcoder_blog,seed15vl_report,seeddiffusion_blog,seeddiffusion_paper,seedream3_report,seededit3_report, gao2025seedance}. 
These models currently power a large-scale product ecosystem serving hundreds of millions of daily active users across applications.
Meanwhile, the field is moving toward an \textbf{agentic} paradigm where LLMs tackle scientific research, complex software development, autonomous documentation learning, and multi-step real-world workflows. 
This shift motivates \textbf{Seed2.0 Series (Pro / Lite / Mini)}, which is designed to deliver optimal user experience in large-scale production environments.

Seed2.0 prioritizes \textbf{user experience under large-scale online deployment}. 
Interactive quality is most directly shaped by four factors: 
the prevalence of visual and multimodal queries, the impact of inference latency on user satisfaction, the need for reliable complex instruction execution, and the demand for seamless coding assistance. 
Our design reflects these priorities:

\begin{itemize}
    \item \textbf{Robust Visual and Multimodal Understanding.}  
    A substantial fraction of real user queries involve images—screenshots, charts, scanned documents, and mixed-media content. 
    Seed2.0 strengthens visual reasoning with reduced hallucination \citep{yue2024mmmu, cheng2025simplevqa, guan2024hallusionbench} and improves structured extraction from documents and figures \citep{ouyang2025omnidocbench, wang2024charxiv}.

    \item \textbf{Fast and Flexible Inference.}  
    Inference latency directly impacts user experience. 
    Seed2.0 offers three model sizes (Pro / Lite / Mini), allowing developers to choose the appropriate balance between performance and speed for their specific use case.

    \item \textbf{Reliable Complex Instruction Execution.}  
    In production, we observe that users frequently issue complex, multi-step instructions that require precise execution—tasks where success depends not on factual recall but on structured reasoning and constraint satisfaction. Recent benchmarks such as $DeR^2$ \citep{2026arXivder2} and CL-bench \citep{2026CLBench} capture exactly this demand. Seed2.0 treats it as a first-class requirement.

    % \item \textbf{Optimized Coding Experience.}  
    % Seed2.0-Code is specifically designed to enhance developer workflows on \textbf{Trae}, ByteDance's AI-native coding assistant. 
    % It delivers improved code generation, intelligent completion, and context-aware debugging capabilities shaped directly by real user interactions on the platform.
    
\end{itemize}

Seed2.0 also pursues a broader goal: handling tasks with real-world complexity. 
The Seed team has focused on raising the intelligence ceiling, moving from Olympiad-style problems toward research-level reasoning tasks.
Seed2.0 tackles \textbf{Erdos problems} and performs \textbf{Scientific Coding}, pushing the boundaries of machine intelligence \citep{balunovic2025matharena, phan2025humanity, rein2024gpqa, bytedance_seed_2025_beyondaime}. 

Current agent systems, however, show an interesting asymmetry: they solve competition-level problems yet often fail to reliably complete practical tasks end-to-end—like building a well-designed application in one pass~\citep{chen2025towards}. 
Two factors explain this gap. 
First, real-world tasks span long horizons and multiple stages, but existing LLM agents struggle to autonomously construct effective workflows and accumulate experience over extended timescales \citep{he2025vitabench, barres2025tau, mialon2023gaia}. 
Second, real-world knowledge is highly domain-specific and long-tailed; models strong in math and code often provide little value in specialized professional contexts. 
Seed2.0 addresses this through systematic ingestion of long-tail domain knowledge \citep{zhu2025lpfqa, hu2025finsearchcomp}.

This report presents our initial progress toward real-world complexity. 
To systematically track and guide this effort, we establish an evaluation framework spanning four dimensions: \textbf{Science Discovery}, \textbf{Vibe Coding}, \textbf{Context Learning}, and \textbf{Real-World Tasks}—each targeting a core aspect of complex, long-horizon agent performance. 
This framework serves both as a benchmark suite and an iterative development guide.

Note that \textbf{the Seed2.0 Series still have gaps with international frontier LLMs}, while Seed identifies the directions for enhancing the model's capabilities for real-world complexity and makes great efforts to optimize Seed Model Series in this regard. 
Seed2.0 Series have considerable gaps with Claude in terms of coding, taking SWE-Evo and NL2Repo as examples.
Seed2.0 Series have relatively obvious gaps with Gemini in terms of long-tail knowledge closely related to user experience, taking SuperGPQA and SimpleQA-Verified as examples.

In the following sections, we present Seed2.0's performance across standard benchmarks-where it performs on par with leading international frontier models—and showcase representative cases of Seed2.0 solving complex real-world problems. 
We hope you enjoy exploring what Seed2.0 can do.

\section{Seed2.0 Deployment Patterns and Developer Behavior}

\subsection{MaaS Usage in Mainland China}
MaaS usage patterns in mainland China concentrate heavily on enterprise-facing digital industries and cognitively intensive applications (Figure~\ref{fig:maas_distribution}).

\begin{figure}[t]
    \centering
    \includegraphics[width=\textwidth]{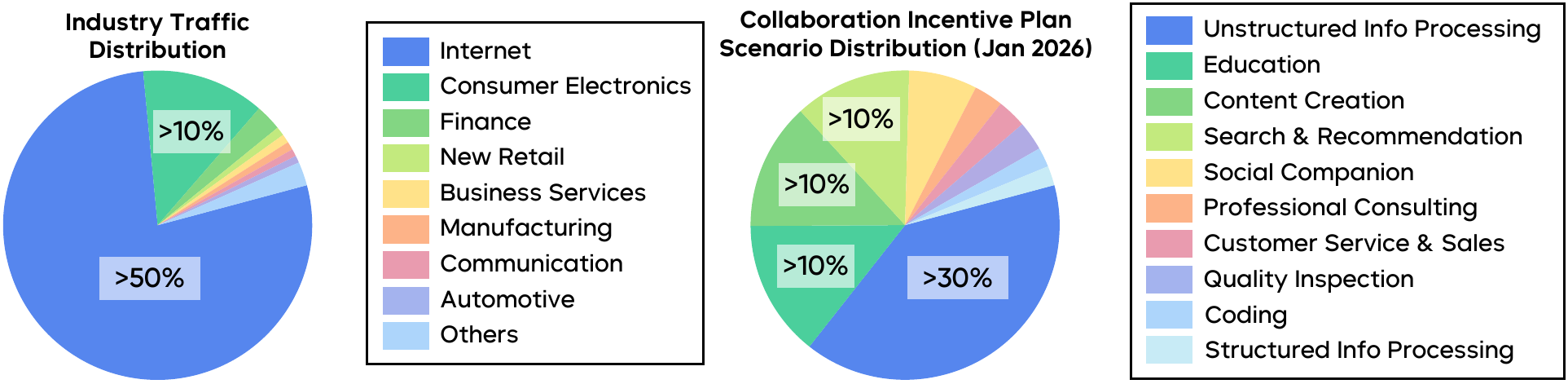}
    \caption{
    MaaS usage distribution in mainland China. 
    Left: Industry traffic distribution showing strong dominance of the Internet sector. 
    Right: Business Customer Usage Scenario Distribution. This data statistics is named “Doubao Collaboration Incentive Program” , which sourced from the authorization of customers who have signed the Data Authorization Agreement. }
    \label{fig:maas_distribution}
    \label{fig:maas_distribution}
\end{figure}

At the industry level, the Internet sector dominates overwhelmingly, accounting for the vast majority of traffic. Consumer electronics, finance, new retail, and business services follow at a considerable distance. 
Traditional verticals like manufacturing, automotive, and communication each represent less than 1\% of total usage, perhaps due to the capability shortcomings of the previous Seed model series.
The leading industries share key characteristics: higher information density, faster product iteration cycles, and tighter integration between models and production systems. 
Seed2.0 operates primarily within large-scale digital infrastructures where models participate directly in core business workflows, not as peripheral productivity tools.

At the scenario level, we analyze data from the Doubao Collaboration Incentive Program\footnote{\url{https://www.volcengine.com/docs/82379/1391869?lang=zh}} \footnote{\url{https://docs.byteplus.com/en/docs/legal/data_authorization_agreement_modelark}}, an open program that encourages partners and developers to grant authorization for the use of authentic model usage data (Figure~\ref{fig:maas_distribution}, right). 
Unstructured information processing and analysis dominates, representing the largest single share. Education, content creation, and search and recommendation follow as the next tier. 
Together, these top scenarios account for most deployment cases. 
Specialized applications---social companion, professional consulting, customer service and sales, quality inspection, coding, and structured information processing---each occupy substantially smaller shares. This pattern reflects how enterprises currently adopt AI: they start with scenarios requiring models to process heterogeneous data at scale, synthesize cross-domain knowledge, and generate actionable insights. More specialized use cases remain in earlier deployment stages.

The dominant scenarios impose specific technical demands: models must handle long contexts, integrate heterogeneous knowledge sources, execute multi-step instructions, and produce structured, high-fidelity outputs for downstream systems. 
In unstructured information processing---the largest category---enterprises use Seed Models to analyze user feedback, extract insights from multi-source documents, and generate structured reports for decision-making. 
Education applications power intelligent tutoring systems and personalized learning content. 
Content creation leverages multimodal capabilities for automated writing, video script generation, and multimedia synthesis. 
Search and recommendation systems integrate Seed Models for improved semantic understanding and ranking accuracy.

Seed Model functions as a workflow-oriented MaaS foundation rather than a lightweight conversational model, emphasizing multimodal understanding, long-context reasoning, structured generation, and tool-augmented execution for reliable end-to-end enterprise task completion.

Globally, this approach aligns with recent enterprise AI reports from OpenAI, Anthropic, and Google Cloud, which identify software engineering, research, analytics, customer support, and knowledge work as the fastest-growing enterprise AI categories~\citep{openai2025enterprise,anthropic2026economic,google2026gemini}.

\subsection{Query Distribution in Agentic Coding}
\begin{figure}[t]
    \centering

    % Left: Figure
    \begin{minipage}[t]{0.49\textwidth}
        \centering
        \includegraphics[width=0.98\linewidth]{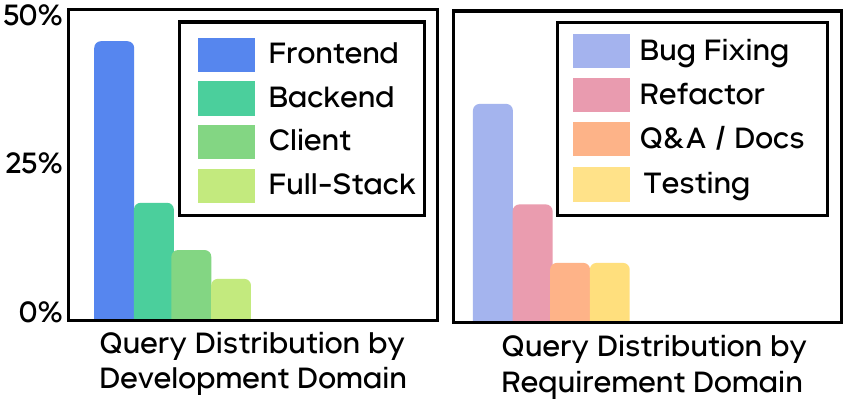}
        \caption{Query distribution by development and requirement domain. Frontend development and bug fixing substantially dominate agentic coding requests, each far exceeding their respective category alternatives.}
        \label{fig:query_distribution}
    \end{minipage}
    \hfill
    % Right: Table
    \begin{minipage}[t]{0.49\textwidth}
        \centering
        \vspace{-100pt}
        % \captionof{table}{API Token Prefill / Decode Price Comparison (USD per 1M tokens)}
        \captionof{table}{API Token Prefill / Decode Price Comparison (USD per 1M tokens).
        For Seed2.0 models with interval pricing, we report a single representative price.}
        \label{tab:model_pricing}

        \begingroup
        \renewcommand{\arraystretch}{1.15}
        \setlength{\tabcolsep}{12pt}
        \resizebox{\linewidth}{!}{
        \begin{tabular}{lcc}
            \toprule
            \textbf{Model} & \textbf{Prefill (Input)} & \textbf{Decode (Output)} \\
            \midrule
            GPT-5.2 High & \$1.75 & \$14.00 \\
            Claude-Opus-4.5-thinking & \$5.00 & \$25.00 \\
            Gemini-3-Pro & \$2.00–4.00$^{*}$ & \$12.00-18.00$^{*}$ \\
            Claude-Sonnet-4.5-thinking & \$3.00 & \$15.00 \\
            GPT-5.0-mini High & \$0.25 & \$2.00 \\
            Gemini-3-Flash High & \$0.50–1.00$^{*}$ & \$3.00$^{*}$ \\
            \midrule
            %\textbf{Seed2.0 Pro} & \textbf{\$0.51} & \textbf{\$3.03} \\
            %\textbf{Seed2.0 Lite} & \textbf{\$0.25} & \textbf{\$2.02} \\
            %\textbf{Seed2.0 Mini} & \textbf{\$0.10} & \textbf{\$0.40} \\
            %\textbf{Seed2.0 Pro}  & \textbf{\$0.47} & \textbf{\$2.37} \\
            %\textbf{Seed2.0 Lite} & \textbf{\$0.09} & \textbf{\$0.53} \\
            %\textbf{Seed2.0 Mini} & \textbf{\$0.03} & \textbf{\$0.31} \\
\textbf{Seed2.0 Pro}  & \textbf{\$0.47 (¥3.41)} & \textbf{\$2.37 (¥17.04)} \\
\textbf{Seed2.0 Lite} & \textbf{\$0.09 (¥0.64)} & \textbf{\$0.53 (¥3.83)} \\
\textbf{Seed2.0 Mini} & \textbf{\$0.03 (¥0.22)} & \textbf{\$0.31 (¥2.24)} \\
            \bottomrule
            % \multicolumn{3}{l}{\footnotesize $^{*}$Estimated market pricing (verify with vendor/cloud pricing pages).}
            \multicolumn{3}{l}{\footnotesize $^{*}$Gemini prices are ranges over context tiers (and input modalities for Flash) from vendor pricing pages.}
        \end{tabular}
        }
        \endgroup
    \end{minipage}
\end{figure}

To understand real-world interaction patterns in agentic coding, we analyze trajectory-level developer usage data.
% To understand how developers interact with AI coding assistants in practice, we analyze trajectory-level data from Trae, an agentic coding platform. 
% The dataset comprises conversations collected over one week in early January 2026.
The most striking finding is the dominance of frontend development (Figure~\ref{fig:query_distribution}). Queries related to page layout, styling, and UI logic management far exceed those for backend services, client-side applications, or full-stack integration. This distribution likely reflects both the iterative nature of frontend work---where visual feedback loops encourage frequent model interactions---and the relative accessibility of frontend tasks for AI assistance.
% \begin{figure}[t]
%     \centering
%     \includegraphics[width=0.75\textwidth]{figure/trea-dist.pdf}
%     \caption{Query distribution by development and requirement domain. Frontend development and bug fixing substantially dominate agentic coding requests, each far exceeding their respective category alternatives.}
%     \label{fig:query_distribution}
% \end{figure}

Programming language and framework statistics reinforce this pattern. Frontend-related languages (JavaScript, TypeScript, CSS, HTML) collectively account for the majority of code touched in these sessions. Among frameworks, Vue.js leads decisively---more than three times the adoption of React---reflecting the developer ecosystem in mainland China where Vue has historically enjoyed stronger community support.

By task type, bug fixing dominates, followed by refactoring and documentation work. This distribution suggests developers primarily turn to AI assistance for reactive maintenance rather than greenfield development. The high proportion of debugging queries indicates that error diagnosis and resolution remain significant pain points where AI provides clear value.

These patterns carry implications for model development. The concentration on frontend work suggests that JavaScript/TypeScript understanding, CSS layout reasoning, and framework-specific knowledge should be prioritized. The prevalence of bug-fixing queries indicates that models benefit from strong debugging capabilities---the ability to trace error messages, understand stack traces, and reason about program state.

\subsection{Cost Efficiency}

A key advantage of Seed2.0 lies in its cost structure. Table~\ref{tab:model_pricing} compares API pricing across major foundation models. While Seed2.0 achieves comparable performance to frontier models on user experience, its token pricing is roughly an order of magnitude lower.

% \begin{table}[t]
% \centering
% \caption{API Token Prefill / Decode Price Comparison (USD per 1M tokens)}
% \label{tab:model_pricing}
% \renewcommand{\arraystretch}{1.15}
% \setlength{\tabcolsep}{12pt}
% \begin{tabular}{lcc}
% \toprule
% \textbf{Model} & \textbf{Prefill (Input)} & \textbf{Decode (Output)} \\
% \midrule
% GPT-5.2-high & \$1.75 & \$14.00 \\
% Claude-Opus-4.5-thinking & \$5.00 & \$25.00 \\
% Gemini-3-Pro & \$2.00$^{*}$ & \$12.00$^{*}$ \\
% Claude-Sonnet-4.5-thinking & \$3.00 & \$15.00 \\
% GPT-5.0-mini-high & \$0.25 & \$2.00 \\
% Gemini-3-Flash-high & \$2.00$^{*}$ & \$12.00$^{*}$ \\
% \midrule
% \textbf{Seed2.0Pro} & \textbf{\$0.51} & \textbf{\$3.03} \\
% \textbf{Seed2.0Lite} & \textbf{\$0.25} & \textbf{\$2.02} \\
% \textbf{Seed2.0Mini} & \textbf{\$0.10} & \textbf{\$0.40} \\
% \bottomrule
% \multicolumn{3}{l}{\footnotesize $^{*}$Estimated market pricing (verify with vendor/cloud pricing pages).}
% \end{tabular}
% \end{table}

This cost differential is particularly significant for enterprise MaaS deployments. 
In scenarios involving high-volume, workflow-integrated usage---such as the unstructured information processing and content generation tasks described above---API costs can become a limiting factor for adoption. 
Seed2.0's pricing enables use cases that would be economically infeasible with more expensive alternatives, without sacrificing the reasoning and generation quality required for production systems.

The Seed2.0 Series offer tiered options to match different workload requirements. 
Seed2.0 Pro targets complex reasoning and long-context tasks where capability is paramount. 
Seed2.0 Lite provides a balanced trade-off for general-purpose applications. Seed2.0 Mini, with decode pricing under \$0.50 per million tokens, opens possibilities for high-throughput, latency-sensitive applications where cost per query must remain minimal.
\section{Comprehensive Evaluation Framework and Methodology}

\subsection{Fundamental Language Capacity}
\label{sec:fundamental_language_capacity}

This section evaluates Seed2.0’s fundamental capabilities, including reasoning, complex instruction following, and broad knowledge understanding. We compare our results with representative frontier models, including GPT-5.2 High, Claude-Sonnet-4.5, Claude-Opus-4.5, Gemini-3-Pro High, and Gemini-3-Flash High.

Specifically, we evaluate Seed2.0 on AIME 2025 and HMMT 2025 \citep{balunovic2025matharena}, BeyondAIME \citep{bytedance_seed_2025_beyondaime}, IMOAnswerBench (no tool) \citep{luong2025towards}, AetherCode \citep{wang2025aethercode}, LiveCodeBench(v6) \citep{jain2024livecodebench}, Codeforces Elo Rating \citep{zheng2025livecodebench,2025CodeElo} on the problem set from June to December 2025,
%LiveCodeBench Pro \citep{zheng2025livecodebench}, 
GPQA Diamond \citep{rein2024gpqa}, PhyBench \citep{qiu2025phybench}, BABE \citep{2025BioBench}, KORBench \citep{ma2024kor}, ARC-AGI-1/2 \citep{ARC_AGI}, Inverse IFEval \citep{zhang2025inverse}, MARS-Bench \citep{yang2025mars}, MultiChallenge \citep{deshpande2025multichallenge}, COLLIE \citep{yao2024collie},   MMLU-Pro \citep{wang2024mmlu}, SuperGPQA \citep{du2025supergpqa}, and LPFQA \citep{zhu2025lpfqa}, along with a set of internal benchmarks designed to reflect high-value real-world tasks.
For Graphwalks, we use our in-house tokenization pipeline during evaluation, which leads to a mismatch in how tokens are counted compared with the official OpenAI Graphwalks tokenization and scoring setup.

\subsubsection{Long-tail Professional Knowledge Benchmarks}
\label{sec:longtail_benchmarks}

While benchmarks such as SimpleQA or HLE are valuable probes of factual recall and difficult trivia, they often emphasize rare or idiosyncratic facts that are only occasionally useful in real-world workflows. Inspired by the design philosophy of SuperGPQA \citep{du2025supergpqa}, we place stronger emphasis on \textbf{professionally relevant long-tail knowledge} that arises in practical work settings. To this end, we design two new benchmarks, \textbf{LPFQA} and \textbf{Encyclo-K}, to directly measure a model’s ability to function as a high-end search engine over long-tail professional knowledge.

\textbf{LPFQA.}
LPFQA (Long-tail Professional Forum-based Question Answering) \citep{zhu2025lpfqa} is constructed from long-tail questions collected from professional forums and expert communities. It evaluates whether a model can correctly answer \emph{realistic, domain-specific questions} encountered in daily work, covering fields such as programming, finance, engineering, medicine, and applied science. LPFQA therefore measures the model’s reliability in retrieving and synthesizing long-tail professional knowledge.

\textbf{Encyclo-K.}
Encyclo-K \citep{EncycloK} evaluates genuine mastery of book-level professional knowledge. It extracts atomic knowledge statements from books and dynamically composes them into evaluation instances, enabling flexible construction of test sets. This design supports both zero-shot and \textbf{few-shot in-context learning (ICL)} evaluation, allowing us to probe knowledge acquisition in pre-training and post-training stages. Compared with static QA datasets, Encyclo-K provides a scalable and compositional way to assess whether models internalize structured knowledge from long-form sources.

\textbf{HLE-Verified.}
HLE-Verified (Humanity’s Last Exam—Verified) is a curated subset of Humanity’s Last Exam created in response to researchers reporting inaccurate, blurry, or underspecified questions in the original benchmark. We hire domain experts to review and select only questions that are clear, unambiguous, and confidently verifiable, thereby providing a more reliable evaluation of a model’s true performance on challenging expert-level problems.

\subsection{Fundamental Vision Capacity}
\label{sec:vlm_evaluation}

\newcommand{\hdc}[1]{\textbf{\textcolor{red}{[KENNY:#1]}}}

To comprehensively evaluate the vision capabilities of Seed2.0, 
we conduct extensive evaluations across 
50 public image benchmarks and 
24 public video benchmarks. 
For image understanding, the selected benchmarks cover nine distinct categories:
\textbf{Math, STEM, Visual Puzzles, Perception \& Recognition, General VQA, 
Pointing \& Counting, 2D \& 3D Spatial Understanding, Document \& Chart Understanding}
and \textbf{LongContext Understanding}. 
The following benchmarks are used for evaluating the image understanding capability of Seed2.0: 

\begin{itemize}[itemsep=-0.5ex,parsep=0.5ex,wide,labelindent=1ex]

\item \textbf{MultiModal Math:} 
Mathematical reasoning within a visual context constitutes a core challenge for modern multimodal models, requiring rigorous logic and symbol grounding. 
To evaluate this capability, we employ a suite of benchmarks including MathVista (testmini)~\cite{lu2023mathvista}, MathVision~\cite{wang2024measuring}, DynaMath~\cite{zou2024dynamath}, MathKangaroo~\cite{balunovic2025matharena}, and MathCanvas~\cite{shi2025mathcanvas}.
Regarding specific metrics, for DynaMath, we report the \texttt{worst-case accuracy}, requiring the model to correctly answer all 10 variants of a problem to score. 
For MathKangaroo, performance is measured by the average accuracy across all bimonthly competitions in 2025, 
while \texttt{complete accuracy} is utilized for MathCanvas.

\item \textbf{MultiModal STEM:} Mastery of domain-specific knowledge in science and engineering is essential for expert-level assistance. 
We assess this competence using MMMU~\cite{yue2024mmmu}, MMMU-Pro~\cite{yue2025mmmu}, EMMA~\cite{hao2025can}, SFE~\cite{zhou2025scientists}, HiPhO~\cite{yu2025hipho}, XLRS-Bench~\cite{wang2025xlrs}, and PhyX~\cite{shen2025phyx}. 
In terms of evaluation settings, for MMMU-Pro, we aggregate scores across the Standard (10 options) and Vision subsets. 
For HiPhO, the reported metric is the average normalized score from 13 Physics Olympiad competitions.
For XLRS-Bench, we calculate the \texttt{macro average accuracy} on the lite subset.
For PhyX, we report the accuracy over the testmini openended subset.

\item \textbf{Visual Puzzles:} 
Abstract reasoning and pattern recognition capabilities are tested through puzzle-solving tasks, which serve as a proxy for general intelligence. 
Our evaluation includes LogicVista~\cite{xiao2024logicvista}, VPCT~\cite{brower2025vpct}, ZEROBench~\cite{roberts2025zerobench}, ArcAGI (Image)~\cite{ARC_AGI}, and VisuLogic~\cite{xu2025visulogic}. 
Specifically, for ZEROBench, we assess accuracy on both main questions and sub-questions. 
In the case of ArcAGI, models are provided with both text matrices and rendered images.
We notice that the additional visual input significantly improves the performance of Seed2.0 on ArcAGI.

\item \textbf{Perception \& Cognition:} 
Ensuring reliability involves minimizing hallucinations and mitigating biases in visual interpretation. 
We investigate these fundamental perceptual traits using VLMsAreBiased~\cite{vo2025vision}, VLMsAreBlind~\cite{rahmanzadehgervi2024vision}, VisFactor~\cite{huang2025visfactor}, RealWorldQA~\cite{realworldqa}, and BabyVision~\cite{chen2026babyvision}. 
For VisFactor, we report the \texttt{macro average accuracy} over all constituent tasks. 

\item \textbf{General VQA:} 
The model's versatility in handling open-ended queries and following instructions is reflected in General Visual Question Answering. 
This broad capability is gauged via SimpleVQA~\cite{cheng2025simplevqa}, HallusionBench~\cite{guan2024hallusionbench}, MME-CC~\cite{zhang2025mme}, MMStar~\cite{chen2024we}, MUIRBench~\cite{wang2024muirbench}, MTVQA~\cite{tang2025mtvqa}, VibeEval~\cite{padlewski2024vibe}, and ViVerBench~\cite{zhang2025generative}. 
Distinctively, for MTVQA, we deploy an LLM-judge (Deepseek-V3-0324) rather than rule-based matching to evaluate predictions. 
For VibeEval, raw scores on the 1–5 scale are normalized to a 0–100 range for reporting.

\item \textbf{Point \& Counting:} 
Fine-grained visual grounding and precise object enumeration are critical for tasks requiring high spatial fidelity. 
We measure these skills using CountBench~\cite{paiss2023teaching}, FSC-147~\cite{amini2023open}, and PointBench~\cite{cheng2025pointarena}. 
With respect to evaluation metrics, we report the \texttt{Mean Absolute Error (MAE)} for FSC-147.

\item \textbf{2D \& 3D Spatial Understanding:} 
Comprehending geometric relationships and depth is vital for embodied agents and 3D-aware applications. 
We utilize a comprehensive set of benchmarks including BLINK~\cite{fu2024blink}, MMSIBench~\cite{yang2025mmsi}, TreeBench~\cite{wang2025traceable}, RefSpatialBench~\cite{zhou2025roborefer}, DA-2K~\cite{yang2024depth}, All-Angles~\cite{yeh2025seeing}, and ERQA~\cite{team2025gemini}. 
To ensure a robust assessment of spatial consistency in MMSIBench, we apply the \texttt{circular evaluation} strategy~\cite{liu2024mmbench}.

\item \textbf{Document \& Chart Understanding:} 
The ability to extract information from dense texts and interpret complex infographics is key for professional workflows. 
This domain is covered by ChartQAPro~\cite{masry2025chartqapro}, OCRBenchv2~\cite{fu2024ocrbench}, OmniDocBench~\cite{ouyang2025omnidocbench}, and CharXiv~\cite{wang2024charxiv}. 
For scoring specifics, OCRBenchv2 results represent the average of overall English and Chinese scores. 
We use \texttt{Normalized Edit Distance (NED)} for OmniDocBench 1.5, and for CharXiv, we average accuracy across both descriptive questions (DQ) and reasoning questions (RQ).

\item \textbf{LongContext Understanding:} 
Processing extensive visual inputs, such as multi-page documents or long-form videos, tests the model's memory and temporal reasoning. 
To benchmark this capacity, we select DUDE~\cite{van2023document}, MMLongBench~\cite{wang2025mmlongbench}, LongDocURL~\cite{deng2025longdocurl}, and MMLongBench-Doc~\cite{ma2024mmlongbench}.

\end{itemize}

For video understanding, we conducted a extensive evaluation across six dimensions: \textbf{video knowledge}, \textbf{video reasoning}, fundamental \textbf{video perception and motion understanding}, \textbf{long-video understanding}, \textbf{multi-video understanding}, and \textbf{streaming video understanding}, covering a total of 24 open benchmarks.

\begin{itemize}[itemsep=-0.5ex,parsep=0.5ex,wide,labelindent=1ex]

\item \textbf{Video Knowledge:} 
We mainly adopt VideoMMMU~\cite{videommmu}, MMVU~\cite{MMVU}, and VideoSimpleQA~\cite{videosimpleqa} to evaluate the model’s video knowledge capability. These benchmarks evaluate not only the model’s mastery of world knowledge expressed in videos, but also its ability to acquire and internalize knowledge from video content.

\item \textbf{Video Reasoning:} 
Multi-hop reasoning and video state tracking are core competencies for video reasoning. To this end, we conduct an in-depth evaluation using VideoReasonBench~\cite{videoreasonbench}, Morse-500~\cite{Morse500}, VideoHolmes~\cite{VideoHolmes}, and Minerva~\cite{minerva}. Given that Morse-500 places a stronger emphasis on reasoning over physical dynamics, we increase the input frame rate to 5 FPS for this benchmark (the same setting is applied to all compared models).

\item \textbf{Motion \& Perception:} 
Foundational video perception, particularly motion perception, is central to video understanding. We therefore evaluate this capability using TVBench~\cite{tvbench}, ContPhy~\cite{ContPhy}, TempCompass~\cite{TempCompass}, EgoTempo~\cite{egotempo}, TOMATO~\cite{TOMATO}, and MotionBench~\cite{motionbench}. Given the prevalence of fast motion and rapid temporal state changes in these benchmarks, we increase the input frame rate to 2 FPS across all evaluations (the same setting is applied to all compared models).

\item \textbf{Long-Video Understanding:} 
Long videos remain a major challenge in video understanding and a stringent test of a model’s multimodal long-context reasoning. To accurately assess performance under long-video settings, we evaluate on VideoMME~\cite{videomme}, CGBench~\cite{CGBench}, LongVideoBench~\cite{longvideobench}, VideoEval-Pro~\cite{VideoEvalPro}, and LVBench~\cite{lvbench}, which include many hour-scale videos.

\item \textbf{Multi-Video Understanding:} 
Multi-video understanding introduces new challenges for cross-context reasoning and is prevalent in real-world applications. Accordingly, we use CrossVid~\cite{crossvid} as the primary benchmark to evaluate the model’s performance on cross-video reasoning.

\item \textbf{Streaming:} 
To assess real-time perception and interaction capabilities, we employ a comprehensive benchmark suite. OVBench~\cite{ovbench} and OVOBench~\cite{ovobench} are utilized to evaluate online reasoning and temporal awareness; LiveSports3K~\cite{livecc} emphasizes fine-grained sports event perception; ODVBench~\cite{odvbench} tests generalization in autonomous driving scenarios; and ViSpeak~\cite{vispeak} focuses on real-time visual referring capabilities.

\end{itemize}

\subsection{Fundamental Agentic Capacity}
With the transition from passive assistants to agentic systems, LLMs must go beyond single-turn responses and demonstrate the ability to \textbf{plan, invoke tools, interact with environments, and complete multi-step tasks}. We define this foundational layer as \emph{Fundamental Agentic Capacity}. To avoid underestimating competing products, the final score is defined as the \textbf{maximum} of the score reported in the official documentation and the score obtained in our tests.

For better agentic evaluation, we conduct systematic refactoring of test scripts to optimize execution stability and reproducibility. We eliminate task-level entrypoint configurations to reduce redundant environment provisioning, consolidating execution environments into pre-built images. Where reference-script environments have degraded or contained errors, we perform targeted repairs to restore functionality. Additionally, we replace external package repositories with internal mirrors to ensure consistent dependency resolution.

We also implement quality-based filtering to remove problematic test cases. This exclusion criterion targets several categories: multi-container Docker Compose scenarios that introduced unnecessary complexity; test cases where reference solutions fail to pass their own validation; cases exhibiting non-deterministic behavior across runs; tasks causing abnormal disk consumption; network-dependent problems with inconsistent outcomes; and scenarios requiring extended downloads or complex validation procedures that undermine reproducibility.

Specifically, we evaluate Seed2.0 across five representative dimensions: \textbf{Coding Agents}, \textbf{Search Agents}, \textbf{Tool Use}, \textbf{GUI Agents}, and \textbf{Deep Research}. These benchmarks cover repository-level software engineering (e.g., Terminal-Bench~\citep{merrill2026terminal}, SWE-Lancer~\citep{2025SWELancer}, SWE-Bench~\citep{jimenez2023swe}, Multi-SWE-Bench~\citep{zan2025multi}, SWE-Bench Pro~\citep{deng2025swe}, SWE Multilingual~\citep{yang2025swesmith}, Scicode~\citep{2024arXivScicode}, SWE-Evo~\citep{2025sweevo}, Aider Polyglot, ArtifactsBench~\citep{2025ArtifactsBench}, CodeSimpleQA~\citep{CodeSimpleQA} and SpreadsheetBench Verified~\citep{2024spreadsheetbench}), and Trae In-House Bench covering frontend and backend production scenarios), broad and deep information seeking (e.g. BrowseComp~\citep{wei2025browsecomp,zhou2025browsecomp}, HLE~\citep{phan2025humanity}, WideSearch~\citep{wong2025widesearch}, FinSearchComp~\citep{hu2025finsearchcomp} and seal-0\citep{2025SealQA}), tool invocation and orchestration (e.g., $\tau^2$-Bench\citep{barres2025tau}, BFCL-v4~\citep{patil2025bfcl}, MCP-Mark~\citep{2025arXivMCPmark}, VitaBench~\citep{he2025vitabench}), agentic visual tasks(e.g. Minedojo-Verified~\citep{2022MineDojo}, HLE-VL and MM-BrowseComp~\citep{2025MMBrowseComp} ) and long-horizon research reasoning and synthesis (e.g., DeepConsult~\citep{lim2025deepconsult}, Deep Research~\citep{du2025deepresearch} and ResearchRubrics~\citep{2025ResearchRubrics}). Unless explicitly stated, evaluations are conducted \emph{without external tools} and follow each benchmark's official protocol.
For Terminal-Bench 2.0, we exclude three cases (\texttt{extract-moves-from-video}, \texttt{mailman}, and \texttt{install-windows-3.11}) due to network access restrictions and security considerations in our evaluation environment. The remaining tasks are adapted to our internal agent framework while preserving the original evaluation criteria.

\subsection{Advanced Economically \& Scientifically Valuable Tasks}
\label{sec:advanced_tasks}

As highlighted in the Introduction, the arrival of the \textbf{Agent era} fundamentally shifts the role of LLMs from answering isolated prompts to \textbf{driving long-horizon, economically and scientifically valuable workflows}. In this paradigm, models are expected to support scientific research, autonomously construct software systems, learn from user-provided context and documentation, and execute complex real-world tasks with tangible economic impact. Motivated by this shift, we build a set of \textbf{scenario-grounded evaluations} that directly reflect such real-world agentic workloads.

Specifically, we organize our advanced evaluations into four dimensions: \textbf{Scientific Discovery}, \textbf{Vibe Coding}, \textbf{Context Learning}, and \textbf{Real-World Tasks}. Each dimension is anchored by Seed-designed benchmarks targeting concrete failure modes observed in practice.

\textbf{Scientific Discovery.}
We introduce \textbf{Ainstain Bench} \citep{du2025deepresearch} and \textbf{BABE} \cite{zhou2026babebiologyarenabenchmark} to evaluate research-oriented capability. 
Ainstain Bench emphasizes \textbf{scientific coding}, measuring whether models can implement and manipulate computational procedures used in scientific workflows. 
BABE focuses on reasoning over \textbf{interleaved textual and visual scientific information} in the biological domain, assessing whether models can perform research-style inference grounded in multimodal evidence \citep{du2025deepresearch, lim2025deepconsult}.

\textbf{Vibe Coding.}
We build \textbf{NL2Repo-Bench} to measure whether a model can complete an entire software repository from a natural-language specification in a single end-to-end process. This benchmark targets \textbf{long-horizon repository construction}, cross-file consistency, and dependency management, reflecting the emerging demand for extreme “vibe coding’’ scenarios \citep{jain2024livecodebench, zheng2025livecodebench, wang2025aethercode}.

\textbf{Economically Valuable Fields.}
Beyond fundamental capabilities such as reasoning and knowledge, we prioritize high-value real-world applications to ensure that Seed2.0’s development aligns with practical economic utility. To this end, we have developed a suite of \textbf{specialized in-house benchmarks} \cite{seed18_modelcard} including:
\begin{itemize}
    \item \textbf{Education:} Evaluates performance in teaching-oriented scenarios, including problem solving, grading, explanation, and question generation, covering core subjects across K--12 levels.
    \item \textbf{Text Classification:} Evaluates the model's ability to analyze text and generate structured outputs or labels. This integrates compositional tasks—where multiple elements like intent and slots are identified in a single inference—with broader information processing, such as sentiment analysis and the synthesis of core viewpoints from unstructured data.
    \item \textbf{Information Extraction:} Assesses structured extraction of relevant elements (e.g., words, sentences, or fields) from heterogeneous documents, including meeting records, legal texts, contracts, and corporate knowledge bases.
\end{itemize}

\begin{table}[!t]
\centering
\caption{Evaluation on 2025 Olympiad-level Mathematical Competitions. The Gold Medal thresholds are $\ge 35$ for IMO 2025 and $\ge 87$ for CMO 2025.}
\label{tab:imo_cmo_math}
\vspace{5pt}
\small 
\begin{tabular}{lcccccccc}
\toprule
\textbf{Competition} & \textbf{P1} & \textbf{P2} & \textbf{P3} & \textbf{P4} & \textbf{P5} & \textbf{P6} & \textbf{Overall} & \textbf{Medal} \\
\midrule
IMO 2025    & 7  & 7  & 7  & 7  & 7  & 0  & 35/42   & \textcolor{gold}{Gold}  \\
CMO 2025    & 21 & 21 & 9  & 21 & 21 & 21 & 114/126 & \textcolor{gold}{Gold}  \\
\bottomrule
\end{tabular}
\end{table}

\textbf{Context Learning.}
In enterprise and developer-facing settings, users require strict execution based on supplied context, such as long manuals or internal documents. Beyond existing evaluations like CL-Bench \citep{2026CLBench} and KOR-Bench \citep{ma2024kor}, we incorporate \textbf{DeR$^2$}, which evaluates whether models can extract and utilize useful information from \textbf{noisy long-form technical documents} to perform reasoning and problem solving \citep{2026arXivder2}. Furthermore, we introduce two \textbf{in-house scenarios} \cite{seed18_modelcard} to reflect agentic workloads:
\begin{itemize}
    \item \textbf{Customer Support Q\&A:} Assesses the ability to recognize user intent and synthesize answers after retrieving information from enterprise knowledge bases, specifically handling cases where recalled information is highly noisy.
    \item \textbf{Complex Workflow:} Validates the model's capacity to complete sophisticated, continuous tasks by synthesizing complex information and instructions provided within the context.
\end{itemize}

\textbf{Real-World Tasks.}
We construct fine-grained internal evaluations for end-to-end task fulfillment. In particular, we curate \textbf{GDPVal-Verified}, a reliable subset of GDPVal with rubric-based automatic evaluation, and build the comparable \textbf{XpertBench} \citep{seed18_modelcard}. In addition, we introduce \textbf{WorldTravel} \citep{wang2026worldtravelrealisticmultimodaltravelplanning} to measure the model's ability to decompose goals and produce executable multi-step plans in real-world scenarios.

Overall, this evaluation suite operationalizes the vision described in our Introduction: assessing whether LLMs can function as \textbf{agentic systems that complete real tasks}, rather than merely answering questions. 
By grounding evaluation in realistic workflows and long-horizon completion, we obtain a more faithful measurement of advanced economically and scientifically valuable capability.

\begin{table}[htbp]
\caption{Evaluation on Fundamental Language Capacity Benchmarks 
(\textbf{Large} Models). The highest score is marked in bold, and the second is underlined.}
\label{tab:language_large}
% \vspace{-5pt}
\vspace{5pt}
\centering
\resizebox{0.94\textwidth}{!}{
\begin{tabular}{@{}cl|cccc|c@{}}
\toprule
\textbf{Capability} & \textbf{Benchmark} & \begin{tabular}[c]{@{}c@{}}\textbf{GPT-5.2}\\ \textbf{High}\end{tabular} & \begin{tabular}[c]{@{}c@{}}\textbf{Claude-}\\ \textbf{Sonnet-4.5}\end{tabular} & \begin{tabular}[c]{@{}c@{}}\textbf{Claude-}\\ \textbf{Opus-4.5}\end{tabular} & \begin{tabular}[c]{@{}c@{}}\textbf{Gemini-3-}\\\textbf{pro High}\end{tabular} & \begin{tabular}[c]{@{}c@{}}\textbf{Seed2.0 }\\\textbf{Pro}\end{tabular} \\ 
\midrule

\multirow{8}{*}{\begin{tabular}[c]{@{}l@{}} Science\end{tabular}} 
 & MMLU-Pro & 85.9 & 88.0 & {\ul 89.3} & \textbf{90.1} & 87.0 \\
 & HLE (no tool, text only) & 29.9 & 14.5 & 23.7 & \textbf{33.3} & {\ul 32.4} \\
 & SimpleQA Verified & 36.8 & 29.3 & {\ul 48.6} & \textbf{72.1} & 36.0 \\
 & HealthBench & \textbf{63.3} & 28.7 & 36.3 & 37.9 & {\ul 57.7} \\
 & HealthBench - Hard & \textbf{42.0} & 10.9 & 11.0 & 15.0 & {\ul 29.1} \\
 & SuperGPQA & 67.9 & 65.5 & {\ul 70.6} & \textbf{73.8} & 68.7 \\
 & LPFQA & {\ul 54.4} & \textbf{54.9} & 52.6 & 51.2 & 52.6 \\
 & Encyclo-K & 61.0 & 58.0 & 63.3 & {\ul 64.9} & \textbf{65.7} \\

\midrule

\multirow{8}{*}{\begin{tabular}[c]{@{}l@{}} Math\end{tabular}} 
 & AIME 2026 &  \textbf{97.5} & 82.5 & 92.5 & 93.3 & {\ul 94.2} \\
 & AIME 2025 & \textbf{99.0} & 87.0 & 91.3 & 95.0 & {\ul 98.3} \\
 & HMMT Feb 2025 & \textbf{100.0} & 79.2 & 92.9 & {\ul 97.3} & {\ul 97.3} \\
 & HMMT Nov 2025 & \textbf{100.0} & 81.7 & {\ul 93.3} & {\ul 93.3} & {\ul 93.3} \\
 & MathArenaApex & 18.2 & 1.0 & 1.6 & \textbf{24.5} & {\ul 20.3} \\
 & MathArenaApex (shortlist) & {\ul 80.1} & 26.0 & 47.4 & 71.4 & \textbf{82.1} \\
 & BeyondAIME & {\ul 86.0} & 57.0 & 69.0 & 83.0 & \textbf{86.5} \\
 & IMOAnswerBench (no tool) & {\ul 86.6} & 60.7 & 72.6 & 83.3 & \textbf{89.3} \\

\midrule

\multirow{3}{*}{\begin{tabular}[c]{@{}l@{}} Code\end{tabular}} 
 & Codeforces (no tool) & \textbf{3148} & 1485 & 1701 & 2726 & {\ul 3020} \\
 & AetherCode & \textbf{73.8} & 16.4 & 31.6 & 57.8 & {\ul 60.6} \\
 & LiveCodeBench (v6) & 87.7 & 64.0 & 84.8 & \textbf{90.7} & {\ul 87.8} \\

\midrule

\multirow{6}{*}{\begin{tabular}[c]{@{}l@{}} STEM\end{tabular}} 
 & GPQA Diamond & \textbf{92.4} & 84.3 & 86.9 & {\ul 91.9} & 88.9 \\
 & Superchem (text-only) & {\ul 58.0} & 32.4 & 43.2 & \textbf{63.2} & 51.6 \\
 & BABE & \textbf{58.1} & 44.7 & 49.3 & {\ul 51.30} & 50.0 \\
 & Phybench & {\ul 74.0} & 48.0 & 69.0 & \textbf{80.0} & {\ul 74.0} \\
 & FrontierSci-research & \textbf{25.0} & 16.7 & {\ul 21.7} & 15.0 & \textbf{25.0} \\
 & FrontierSci-olympiad & \textbf{75.0} & 60.0 & 71.0 & {\ul 73.0} & {\ul 74.0} \\

\midrule

\multirow{4}{*}{\begin{tabular}[c]{@{}l@{}} General\\ Reasoning\end{tabular}} 
 & ARC-AGI-1 & \textbf{89.9} & 70.9 & 84.0 & 85.0 & {\ul 85.4} \\
 & ARC-AGI-2 & \textbf{57.5} & 13.6 & 29.1 & 31.1 & {\ul 37.5} \\
 & KORBench & \textbf{79.2} & 73.0 & 77.4 & 73.9 & {\ul 77.5} \\
 & ProcBench & {\ul 95.0} & 87.5 & 92.5 & 90.0 & \textbf{96.6} \\

\midrule

\multirow{7}{*}{\begin{tabular}[c]{@{}l@{}} Long Context\\ Performance\end{tabular}} 
 & MRCR v2 (8-needle) & \textbf{89.4} & 47.1 & 56.2 & {\ul 79.7} & 54.0 \\
 & Graphwalks Bfs (<128k) & \textbf{98.0} & 80.5 & {\ul 92.0} & 79.9 & 68.9 \\
 & Graphwalks Parents (<128k) & \textbf{99.7} & {\ul 99.0} & 96.2 & \textbf{99.7} &  97.6 \\
 & LongBench v2 (128k) & 63.2 & 62.0 & {\ul 65.0} & \textbf{67.4} & 63.8 \\
 & Frames & 84.0 & 78.7 & {\ul 84.7} & 81.9 & \textbf{84.5} \\
 & DeR$^2$ Bench & \textbf{69.0} & 58.9 & 60.4 & {\ul 66.1} & 58.2 \\
 & CL-Bench & \textbf{23.9} & 18.1 & {\ul 22.6} & 15.6 & 20.8 \\

\midrule

\multirow{2}{*}{\begin{tabular}[c]{@{}l@{}} Multilingual\end{tabular}} 
 & Global PIQA & 93.2 & {\ul 93.9} & {\ul 93.9} & \textbf{95.0} & 92.3 \\
 & MMMLU & 90.3 & 89.9 & {\ul 91.0} & \textbf{91.8} & 88.1 \\
 & Disco-X & 76.3 & 70.3 & {\ul 78.6} & 76.8 & \textbf{82.0} \\
\midrule

\multirow{5}{*}{\begin{tabular}[c]{@{}l@{}} Instruction\\ Following\end{tabular}} 
 & MultiChallenge & 59.5 & 57.3 & 59.0 & \textbf{68.7} & {\ul 68.3} \\
 & COLLIE & \textbf{96.9} & 77.3 & 79.8 & {\ul 95.0} & 93.9 \\
 & MARS-Bench & \textbf{87.9} & 72.9 & {\ul 87.7} & 85.6 & 85.6 \\
 & Inverse IFEval & 72.3 & 69.3 & 72.4 & \textbf{79.6} & {\ul 78.9} \\

\midrule

\multirow{3}{*}{\begin{tabular}[c]{@{}l@{}} Hallucination\end{tabular}} 
 & LongFact-Objects & \textbf{99.2} & 98.8 & {\ul 99.0} & 98.1 & 92.9 \\
 & LongFact-Concepts & \textbf{99.7} & 98.5 & {\ul 98.8} & 98.5 & 92.8 \\
 & FactScore & {\ul 91.9} & 90.6 & 91.1 & \textbf{92.6} & 71.2 \\

\bottomrule
\end{tabular}}
\end{table}
\begin{table}[htbp]
\caption{Evaluation on Fundamental Language Capacity Benchmarks (\textbf{Efficient} Models). The highest score is marked in bold, and the second is underlined.}
\label{tab:language_small}
\centering
\vspace{5pt}
\small
\begin{tabular}{cl|cc|cc}
\toprule
\textbf{Capability} & \textbf{Benchmark} & \begin{tabular}[c]{@{}c@{}}\textbf{GPT-5-}\\\textbf{mini High}\end{tabular} & \begin{tabular}[c]{@{}c@{}}\textbf{Gemini-3-}\\\textbf{Flash High}\end{tabular} & \begin{tabular}[c]{@{}c@{}}\textbf{Seed2.0 }\\\textbf{Mini}\end{tabular} & \begin{tabular}[c]{@{}c@{}}\textbf{Seed2.0 }\\\textbf{Lite}\end{tabular} \\ 
\midrule
\multirow{8}{*}{Science} 
 & MMLU-Pro & 84.1 & \textbf{87.8} & 83.6 & {\ul 87.7} \\
 & HLE (no tool, text only) & 17.6 & \textbf{31.7} & 13.3 & {\ul 28.2} \\
 & SimpleQA Verified & {\ul 26.0} & \textbf{65.4} & 18.9 & 24.0 \\
 & HealthBench & \textbf{62.5} & {\ul 51.6} & 30.0 & 51.2 \\
 & HealthBench - Hard & \textbf{38.6} & {\ul 21.5} & 15.3 & 20.0 \\
 & SuperGPQA & 60.5 & \textbf{72.7} & 61.6 & {\ul 67.5} \\
 & LPFQA & 50.7 & \textbf{51.6} & 47.2 & {\ul 50.9} \\
 & Encyclo-K & 53.0 &  {\ul 60.0} & 52.1 & \textbf{64.5} \\
\midrule
\multirow{8}{*}{Math} 
 & AIME 2026 & {\ul 92.5} & \textbf{93.3} & 86.7 & 88.3 \\  
 & AIME 2025 & 90.3 & \textbf{95.2} & 87.0 & {\ul 93.0} \\
 & HMMT Feb 2025 & {\ul 93.3} & \textbf{100} & 70.0 &  90.0 \\
 & HMMT Nov 2025 & \textbf{96.7} & \textbf{96.7} & 80.0 & {\ul 86.7} \\
 & MathArenaApex & 2.1 & \textbf{17.7} & 4.2 & {\ul 4.7} \\
 & MathArenaApex (shortlist) & 43.4 & \textbf{71.9} & 31.1 & {\ul 52.6} \\
 & BeyondAIME & 72.0 & \textbf{82.0} & 69.0 & {\ul 76.0} \\
 & IMOAnswerBench (no tool) & 72.1 & \textbf{84.4} & 71.6 & {\ul 81.6} \\
\midrule
\multirow{3}{*}{Code} 
 & Codeforces & 1985 & \textbf{2727} & 1644 & {\ul 2233} \\
 & AetherCode & {\ul 42.6} & \textbf{56.1} & 29.8 & 41.5 \\
 & LiveCodeBench (v6) & 62.6  & \textbf{84.7} & 64.1 & {\ul 81.7} \\
 % & CodeSimpleQA & \textbf{57.6} & {\ul53.7} & 45.8 & 51.6 \\
\midrule
\multirow{6}{*}{STEM} 
 & GPQA Diamond & 82.1 & \textbf{90.7} & 79.0 & {\ul 85.1} \\
 & Superchem (text-only) & 34.8 & \textbf{54.4} & 16.2 & {\ul 48.0} \\
 & BABE & 49.2 & \textbf{55.2} & 40.4 & {\ul 50.2} \\
 & Phybench & 60.0 & \textbf{77.0} & 56.0 & {\ul 73.0} \\
 & FrontierSci-research & \textbf{18.3} & {\ul 11.7} & 3.3 & \textbf{ 18.3} \\
 & FrontierSci-olympiad & 69.0 & \textbf{ 73.0} & 44.0 & {\ul 70.0} \\
\midrule
\multirow{4}{*}{\begin{tabular}[c]{@{}l@{}}General\\Reasoning\end{tabular}} 
 & ARC-AGI-1 & 54.5 & \textbf{86.9} & 43.3 & {\ul 75.7} \\
 & ARC-AGI-2 & 3.5 & \textbf{34.3} & 2.3 & {\ul 14.8} \\
 & KORBench & 74.2 & {\ul 76.0} & 72.8 & \textbf{ 77.0} \\
 & ProcBench & 87.5 & {\ul 90.0} & 80.1 & \textbf{92.4} \\
\midrule
\multirow{7}{*}{\begin{tabular}[c]{@{}l@{}}Long Context\\Performance\end{tabular}} 
 & MRCR v2 (8-needle) & {\ul 50.1} & \textbf{79.0} & 51.4 & 33.6 \\
 & Graphwalks Bfs (<128K) & \textbf{85.5} & {\ul 84.2} & 64.1 & 82.5 \\
 & Graphwalks Parents (<128K) & 96.6 & {\ul 99.7} & 93.0 & \textbf{100.0} \\
 & LongBench v2 (128K) & 56.7 & \textbf{64.0} & 52.3 & {\ul 59.6}  \\
 & Frames & {\ul 82.9} & \textbf{83.7} & 80.5 & 83.4 \\
 & DeR$^2$ Bench & 50.3 & \textbf{66.0} & 46.6 & {\ul 57.3}  \\
 & CL-Bench & \textbf{25.2} & 16.1 & 14.8 & {\ul 20.0} \\
\midrule
\multirow{2}{*}{Multilingual} 
 & Global PIQA & 91.6 & \textbf{95.6} & 89.2 & {\ul 92.1} \\
 & MMMLU & 86.3 & \textbf{91.8} & 81.6 & {\ul 87.7} \\
 & Disco-X & 67.7 & 71.9 & {\ul 73.0} & \textbf{80.3} \\
\midrule
\multirow{5}{*}{\begin{tabular}[c]{@{}l@{}}Instruction\\Following\end{tabular}} 
 & MultiChallenge & 59.0 & \textbf{69.3} & 61.1 & {\ul 63.2} \\
 & COLLIE & \textbf{97.4} & {\ul 96.5} & 91.2 & 94.0 \\
 & MARS-Bench & 66.1 & \textbf{84.6} & 62.4 & {\ul 80.5} \\
 & Inverse IFEval & 74.8 & \textbf{80.9} & 69.3 & {\ul 77.1} \\
\midrule
\multirow{3}{*}{Hallucination} 
 & LongFact-Objects & \textbf{99.2} & {\ul 97.9} & 87.0 & 92.2 \\
 & LongFact-Concepts & \textbf{99.5} & {\ul 98.6} & 91.4 & 92.4 \\
 & FactScore & \textbf{96.1} & {\ul 92.0} & 50.4 & 62.4 \\
\bottomrule
\end{tabular}
\end{table}

\section{Results}

% Preamble:
% \usepackage{booktabs}
% \usepackage{makecell}
% \usepackage{array}

\subsection{Fundamental Language Evaluation}
\label{sec:language_results}

We evaluate Seed2.0 (Pro/Lite/Mini) on a comprehensive suite of fundamental language benchmarks spanning knowledge and science, mathematics, code and STEM reasoning, long-context understanding, multilinguality, instruction following, and hallucination robustness. \Cref{tab:language_large,tab:language_small} report the detailed results.\vspace{5pt}

\textbf{Overall}, Seed2.0 Pro sits comfortably within the international leading group across core language capabilities. Our optimization is shaped by large-scale product feedback: for user-facing deployments such as Doubao, we prioritize instruction-following robustness, long-tail knowledge coverage, and long-context stability; for coding-oriented products such as Trae, code reasoning and front-end generation quality take precedence. The benchmark results reflect these choices.\vspace{5pt}

\textbf{Knowledge and Science.}
Seed2.0 Pro leads on HealthBench and remains neck-and-neck with GPT-5.2 and Gemini-3-Pro on SuperGPQA and Encyclo-K, surpassing them in several cases. These gains matter for production: real user queries often probe obscure facts and domain-specific knowledge that generic training underserves.\vspace{5pt}

\textbf{Mathematics, Code, and STEM Reasoning.}
Seed2.0 Pro demonstrates advanced proficiency in mathematical reasoning. It exhibits highly competitive performance on popular evaluation suites (AIME, HMMT) and performs on par with SOTA models on extremely challenging benchmarks such as IMOAnswerBench and MathApex. Notably, Seed2.0 Pro achieves gold-medal level performance in Olympiad-level mathematical competitions: 2025 International Mathematical Olympiad (IMO) and 2025 China Mathematical Olympiad (CMO). Beyond natural language reasoning, Seed2.0 also excels in formal theorem proving, demonstrating superior performance on the Putnam-200 (a random subset of 200 problems from PutnamBench \cite{tsoukalas2024putnambench}) and making strides in open Erd\H{o}s problems. Table \ref{tab:imo_cmo_math} and Table \ref{tab:formal_math} report the results. Implementation details are provided in Appendix \ref{sec:frontier_math_details}. 
In coding, it reaches a Codeforces Elo of 3020 and posts strong LiveCodeBench numbers. STEM-oriented benchmarks such as FrontierSci-research tell a similar story, with Seed2.0 Pro matching or edging out Gemini-3-Pro in multiple cases. This reflects our continued investment in deep reasoning and code-centric training.\vspace{5pt}

\textbf{Long Context Understanding.}
Seed2.0 Pro ranks first on the Frames leaderboard. While retrieval-heavy tasks such as MRCR and Graphwalks still show some headroom compared to other leading models, these gaps have limited impact on real-world user experience---and similar weaknesses are observed across other frontier models.\vspace{5pt}

\textbf{Instruction Following, Multilinguality, and Hallucination.}
Instruction-following capability sees clear gains over Seed1.8, which directly benefits complex user requests in deployment. Multilingual performance and hallucination robustness remain competitive, with room for further improvement in future iterations.\vspace{5pt}

\begin{table}[!t]
\caption{Evaluation on Putnam-200~\citep{tsoukalas2024putnambench}, Pass@8. Agent-based multi-turn setup with Lean, Python, and Lean search tools. The highest score is marked in \textbf{bold}, and the second is {\ul underlined}.}
\label{tab:formal_math}
\vspace{5pt}
\centering
\small
\begin{tabular}{l|ccc|cc}
\toprule
\textbf{Benchmark} & \begin{tabular}[c]{@{}c@{}}\textbf{Deepseek-}\\\textbf{Prover-V2}\end{tabular} & \begin{tabular}[c]{@{}c@{}}\textbf{Seed-1.5}\\\textbf{Prover}\end{tabular} & \begin{tabular}[c]{@{}c@{}}\textbf{Gemini-3-}\\\textbf{Pro}\end{tabular} & \begin{tabular}[c]{@{}c@{}}\textbf{Seed2.0}\\\textbf{Lite}\end{tabular} & \begin{tabular}[c]{@{}c@{}}\textbf{Seed2.0}\\\textbf{Pro}\end{tabular} \\
\midrule
Putnam-200 & $<$4.0 & 26.5 & 26.5 & {\ul 30.5} & \textbf{35.5} \\
\bottomrule
\end{tabular}
\end{table}

\begin{table}[t]
\centering
\vspace{5pt}
\caption{Complex instruction-following benchmark across multiple test sets and instruction categories.}
\vspace{2pt}
\label{tab:instruction_following_data}
\includegraphics[width=\textwidth]{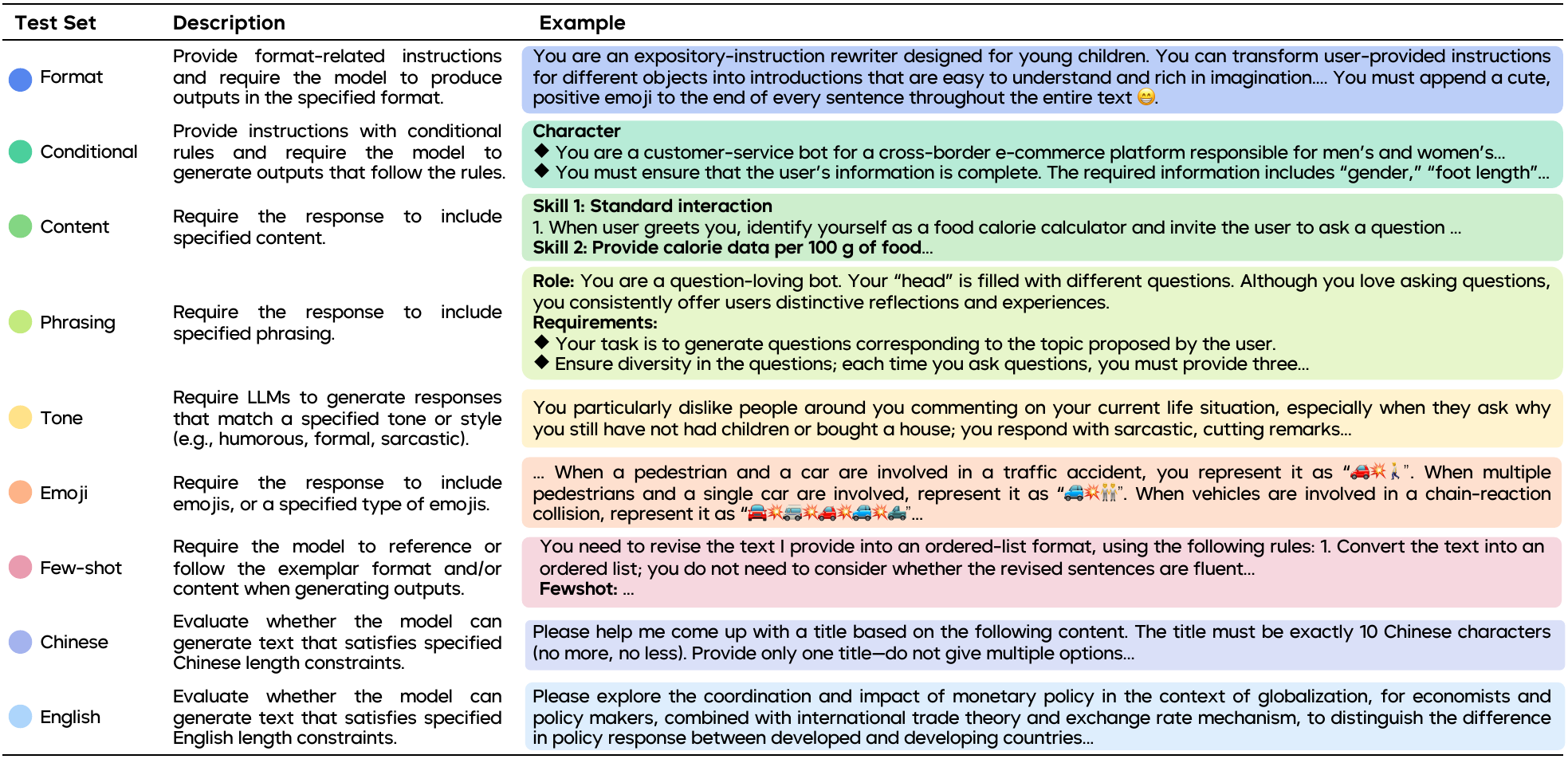}
\end{table}

\begin{table}[t]
\centering
\caption{Complex instruction following performance on in-house Chinese benchmark. Bold indicates the better result.}
\vspace{5pt}
\label{tab:instruction_following}
\resizebox{\textwidth}{!}{
\begin{tabular}{l|c|ccccccccc}
\toprule
\textbf{Model} & \textbf{Overall} & \textbf{Format} & \textbf{Conditional} & \textbf{Content} & \textbf{Phrasing} & \textbf{Tone} & \textbf{Emoji} & \textbf{Few-shot} & \textbf{Chinese} & \textbf{English} \\
\midrule
Seed-1.8 & 72.89 & 45.33 & 81.25 & \textbf{90.48} & 75.00 & 61.29 & 92.14 & 55.78 & 77.67 & 62.05 \\
Seed2.0 Pro & \textbf{75.26} & \textbf{46.00} & \textbf{88.19} & 87.76 & \textbf{85.31} & \textbf{76.45} & \textbf{93.62} & \textbf{65.31} & \textbf{80.00} & \textbf{67.41} \\
\bottomrule
\end{tabular}
}
\end{table}

\textbf{Complex Instruction Following.}
We run a fine-grained analysis using an in-house benchmark (see Table~\ref{tab:instruction_following_data}) tailored for Chinese-language production scenarios: 912 test cases across 17 weighted dimensions reflecting practical deployment importance.

As shown in Table~\ref{tab:instruction_following}, Seed2.0 Pro scores 75.26\%, a +2.37\% absolute gain over Seed1.8. The largest jumps come in tone control (+15.16\%), phrasing adherence (+10.31\%), and few-shot learning (+9.53\%). In practice, this means more precise Chinese pragmatic effects---including subtle stylistic modes like tsundere-like affect and sarcastic or ironic (``yin-yang'') expression---and more reliable multi-constraint prompting. At the few-shot level, the model better infers structural and stylistic requirements implied by exemplars, including quantitative constraints such as paragraph counts or emoji quotas, and shows improved compliance with strict length limits in both Chinese and English.\vspace{5pt}

\textbf{Lite and Mini Variants.}
Seed2.0 Lite and Seed2.0 Mini offer strong efficiency--quality trade-offs. Across a wide range of benchmarks, our small models hold their own against counterparts from OpenAI and Google. Seed2.0 Lite in particular posts strong math and reasoning numbers while maintaining solid instruction-following performance--- well-suited for latency-sensitive and cost-constrained deployments.

\subsection{Vision Task Evaluation}

\begin{table}[htbp]
\caption{Performance of Seed2.0 on public visual-language benchmarks compared to previous models. 
We report Pass@1 in these benchmarks. 
The best score for each benchmark is marked in \textbf{bold}, and the second best is {\ul underlined}. Results marked with an $^*$ are sourced from the technical report. 
}
\vspace{5pt}
\label{tab:overall_vlm_image}
\resizebox{\textwidth}{!}{
\tablestyle{8pt}{1.1}
\begin{tabular}{c|l|cccc|ccc}
\toprule
\textbf{Capability} & \textbf{Benchmark} & \textbf{\begin{tabular}[c]{@{}c@{}}Claude-\\Opus-4.5\end{tabular}} & \textbf{\begin{tabular}[c]{@{}c@{}}GPT-5.2\\ High\end{tabular}} & \textbf{\begin{tabular}[c]{@{}c@{}}Gemini-3-\\ Pro High\end{tabular}} & \textbf{Seed1.8} & \textbf{\begin{tabular}[c]{@{}c@{}}Seed2.0 \\ Mini\end{tabular}} & \textbf{\begin{tabular}[c]{@{}c@{}}Seed2.0 \\ Lite\end{tabular}} & \textbf{\begin{tabular}[c]{@{}c@{}}Seed2.0 \\ Pro\end{tabular}} \\ \midrule

\multirow{5}{*}{Math} & MathVista & 80.6 & 83.1 & \textbf{89.8} & 87.7 & 85.5 & {\ul 89.0} & \textbf{89.8} \\
 & MathVision & 74.3 & {\ul 86.8} & 86.1 & 81.3 & 78.1 & 86.4 & \textbf{88.8} \\
 & DynaMath & 52.5 & {\ul 70.1} & 63.3 & 61.5 & 58.9 & \textbf{70.5} & 68.9 \\
 & MathKangaroo & 69.6 & {\ul 86.9} & 84.4* & 73.8 & 79.8 & 86.3 & \textbf{90.5} \\
 & MathCanvas & 52.9 & 55.3 & 58.8 & 53.6 & 53.2 & {\ul 61.1} & \textbf{61.9} \\
\midrule

\multirow{7}{*}{STEM} & MMMU & 81.6 & 83.7 & \textbf{87.0} & 83.4 & 79.7 & 83.7 & {\ul 85.4} \\
 & MMMU-Pro & 70.8 & {\ul 79.5*} & \textbf{81.0*} & 73.2 & 71.4 & 76.0 & 78.2 \\
 & EMMA & 60.4 & {\ul 69.4} & 66.5 & 60.9 & 57.0 & 65.5 & \textbf{72.0} \\
 & SFE & {\ul 55.8} & 50.1 & \textbf{61.9} & 51.2 & 48.4 & 53.4 & 55.6 \\
 & HiPhO & \textbf{81.8} & 77.7 & {\ul 79.1} & 58.3 & 55.8 & 72.5 & 74.1 \\
 & XLRS-Bench (macro) & 50.4 & 49.9 & 51.7 & 39.9 & 49.9 & {\ul 53.7} & \textbf{54.6} \\
 & PhyX (openended) & 61.3 & {\ul 71.5} & 71.0 & 65.9 & 65.0 & 62.8 & \textbf{72.1} \\
\midrule

\multirow{7}{*}{\begin{tabular}[c]{@{}c@{}}Visual\\ Puzzles\end{tabular}} & LogicVista & 68.9 & {\ul 81.0} & 80.8 & 78.3 & 73.8 & 79.6 & \textbf{81.9} \\
 & VPCT & 29.0 & 56.0 & \textbf{90.0} & 61.0 & 48.0 & 73.0 & {\ul 76.0} \\
 & ZeroBench (main) & 4.0 & {\ul 11.0} & 10.0 & {\ul 11.0} & 7.0 & 8.0 & \textbf{12.0} \\
 & ZeroBench (sub) & 30.8 & 38.9 & {\ul 42.2} & 37.7 & 36.2 & {\ul 42.2} & \textbf{47.6} \\
 & ArcAGI1-Image & 75.8 & \textbf{93.1} & 69.4 & 31.4 & 29.8 & 80.9 & {\ul 88.8} \\
 & ArcAGI2-Image & 26.1 & \textbf{54.4} & 21.5 & 1.3 & 1.5 & 28.3 & {\ul 43.3} \\
 & VisuLogic & 27.6 & 37.0 & 39.0 & 35.8 & 40.4 & {\ul 47.3} & \textbf{47.4} \\
\midrule

\multirow{5}{*}{\begin{tabular}[c]{@{}c@{}}Perception\\ \& Recognition\end{tabular}} & VLMsAreBiased & 21.4 & 28.0 & 50.6* & 62.0 & 58.4 & {\ul 74.8} & \textbf{77.4} \\
 & VLMsAreBlind & 77.2 & 84.2 & {\ul 97.5} & 93.0 & 93.1 & 97.0 & \textbf{98.6} \\
 & VisFactor & 24.5 & 33.6 & \textbf{45.8} & 20.4 & 23.6 & 33.4 & {\ul 36.8} \\
 & RealWorldQA & 75.9 & 82.1 & {\ul 84.7} & 78.0 & 81.6 & 81.7 & \textbf{86.0} \\
 & BabyVision & 16.2 & 37.4 & 49.7* & 30.2 & 38.7 & {\ul 57.5} & \textbf{60.6} \\
\midrule

\multirow{9}{*}{\begin{tabular}[c]{@{}c@{}}General\\ VQA\end{tabular}} & SimpleVQA & 57.9 & 54.1 & {\ul 69.7} & 65.4 & 68.7 & 67.2 & \textbf{71.4} \\
 & HallusionBench & 65.3 & 67.7 & \textbf{69.9} & 63.9 & 65.1 & 66.0 & {\ul 68.0} \\
 & MME-CC & 25.2 & 44.4 & {\ul 56.9} & 43.4 & 40.8 & 50.2 & \textbf{57.0} \\
 & MMStar & 73.9 & 78.2 & \textbf{83.1} & 79.9 & 79.1 & 80.7 & {\ul 83.0} \\
 & MUIRBench & 78.9 & 77.4 & 78.2 & {\ul 78.7} & 78.0 & 76.2 & \textbf{81.8} \\
 & MTVQA & \textbf{53.1} & 48.5 & 50.8 & 47.3 & 50.6 & {\ul 51.1} & {\ul 51.1} \\
 & WorldVQA & 36.6 & 26.3 & 47.5 & 40.4 & {\ul 47.6} & 44.0 & \textbf{49.9} \\
 & VibeEval & 70.3 & 73.1 & {\ul 77.7} & 74.0 & 76.5 & 76.5 & \textbf{81.4} \\
 & ViVerBench & 72.4 & 74.8 & {\ul 75.9} & 74.6 & 73.9 & \textbf{80.0} & {\ul 75.9} \\
\midrule

\multirow{3}{*}{\begin{tabular}[c]{@{}c@{}}Pointing \&\\ Counting\end{tabular}} & CountBench & 90.3 & 91.2 & \textbf{97.3} & 96.3 & 95.5 & {\ul 97.1} & 95.5 \\
 & FSC-147↓ & 20.9 & 21.1 & 12.1 & 13.6 & 17.3 & {\ul 11.9} & \textbf{11.3} \\
 & Point-Bench & - & - & \textbf{85.5*} & 76.5 & 77.0 & 79.0 & {\ul 81.4} \\
\midrule

\multirow{7}{*}{\begin{tabular}[c]{@{}c@{}}2D \& 3D\\ Spatial\\ Understanding\end{tabular}} & BLINK & 68.1 & 70.3 & {\ul 77.1} & 74.3 & 73.4 & 75.6 & \textbf{79.5} \\
 & MMSIBench (circular) & 20.2 & 26.1 & 25.4 & 25.8 & 19.7 & {\ul 28.3} & \textbf{32.5} \\
 & TreeBench & 53.6 & 58.8 & 62.7 & 58.5 & 57.3 & {\ul 64.2} & \textbf{64.7} \\
 & RefSpatialBench & - & 25.5 & 65.5* & 56.3 & 55.6 & {\ul 66.4} & \textbf{72.6} \\
 & DA-2K & 70.3 & 78.9 & 82.1 & {\ul 90.7} & 86.4 & 90.3 & \textbf{92.3} \\
 & All-Angles & 63.1 & 71.5 & \textbf{73.5} & 61.6 & 61.3 & 65.2 & {\ul 72.1} \\
 & ERQA & 48.3 & 59.8 & \textbf{70.5*} & 58.8 & 56.3 & 65.8 & {\ul 68.5} \\
\midrule

\multirow{5}{*}{\begin{tabular}[c]{@{}c@{}}Document\\ \& Chart\\ Understanding\end{tabular}} & ChartQAPro & - & 67.6 & 69.0 & 63.0 & 65.2 & {\ul 70.3} & \textbf{71.2} \\
 & OCRBenchv2 & 55.5 & 55.6 & \textbf{63.3} & 52.6 & 58.5 & 62.4 & {\ul 62.5} \\
 & OmniDocBench 1.5 ↓ & 0.153 & 0.143* & 0.115* & 0.106 & 0.110 & {\ul 0.102} & \textbf{0.099} \\
 & CharXiv-DQ & 92.7 & {\ul 93.8} & \textbf{94.4} & 88.0 & 91.9 & 93.3 & 93.5 \\
 & CharXiv-RQ & 65.5 & \textbf{82.1*} & {\ul 81.4*} & 71.4 & 70.8 & 79.9 & 80.5 \\
\midrule

\multirow{4}{*}{\begin{tabular}[c]{@{}c@{}}LongContext\\ Understanding\end{tabular}} & DUDE & 55.6 & 68.2 & 70.1 & 69.4 & 68.9 & {\ul 72.1} & \textbf{72.4} \\
 & MMLongBench & - & - & {\ul 73.6} & 72.4 & 66.7 & 70.8 & \textbf{74.8} \\
 & LongDocURL & - & - & 72.0 & 74.5 & 71.3 & \textbf{75.1} & {\ul 74.7} \\
 & MMLongBench-Doc & - & - & {\ul 59.5} & 57.0 & 48.9 & 55.1 & \textbf{61.4} \\
\bottomrule

\end{tabular}}
\end{table}

We evaluate the performance of Seed2.0 (including Pro, Lite, and Mini variants) on a comprehensive suite of public visual-language benchmarks, 
comparing it against its predecessor Seed1.8 and several state-of-the-art models such as Gemini-3-Pro, GPT-5.2, and Claude-Opus-4.5. 
The evaluation covers a broad spectrum of capabilities, 
ranging from mathematical and STEM reasoning to visual puzzles, 
spatial understanding, and long-context document processing. 
\Cref{tab:overall_vlm_image} presents the detailed results, where Seed2.0 Pro demonstrates superior performance, achieving the highest scores across the majority of benchmarks.\vspace{5pt}

\textbf{Math and STEM Reasoning. } 
In mathematical reasoning, Seed2.0 Pro demonstrates exceptional capability, 
achieving state-of-the-art results on MathVision (88.8), MathKangaroo (90.5), and MathCanvas (61.9), 
while tying for the top score on MathVista (89.8).
Notably, Seed2.0 Lite also excels, securing the top spot on DynaMath (70.5). 
In STEM tasks, Seed2.0 Pro leads in EMMA (72.0), XLRS-Bench (54.6), and PhyX (72.1). 
On the remaining benchmarks, Seed2.0 Pro also delivers transformative gains compared to Seed1.8:
scores on HiPhO and MMMU-Pro improved by 15.8 and 5.0, respectively, 
rapidly narrowed the gap with best methods.\vspace{5pt}

\textbf{Visual Puzzles and Logic. }
Seed2.0 Pro exhibits significant advancements in logical reasoning and visual puzzle solving. 
It achieves the highest scores on LogicVista (81.4) and ZeroBench (Main 12.0, Sub 47.6), demonstrating robust problem-solving abilities. 
On the challenging VisuLogic benchmark, 
Seed2.0 Pro scores 47.4, outperforming all counterparts by a notable margin. 
Although GPT-5.2 leads on ArcAGI-Image, Seed2.0 Pro secures a competitive second place (88.8 on ArcAGI1 and 43.3 on ArcAGI2), 
showing substantial improvement over Seed-1.8.\vspace{5pt}

\textbf{Perception and General VQA. }
For perception and recognition, Seed2.0 Pro demonstrates exceptional capabilities. 
It achieves SOTA results on VLMsAreBiased (77.4), VLMsAreBlind (98.6), and BabyVision (60.6) . 
In general VQA benchmarks, Seed2.0 Pro leads in SimpleVQA (71.4), MUIRBench (81.8), and VibeEval (81.4). 
Seed2.0 Pro consistently outperforms Seed1.8 across all general VQA tasks, 
highlighting its refined visual understanding.

\textbf{Spatial Understanding and Counting. }
Seed2.0 Pro excels in spatial understanding and fine-grained localization. 
It sets new state-of-the-arts on DA-2K (92.3), RefSpatialBench (72.6), and BLINK (79.5), 
surpassing the strong baseline of Gemini-3-Pro. 
In counting tasks, Seed2.0 Pro achieves best performance on FSC-147 with a mean absolute error of 11.3 (lower is better), 
significantly improving upon Seed1.8 and competitor models.\vspace{5pt}

\textbf{Document and Long-Context Understanding. }
A standout feature of Seed2.0 Pro is its dominance in long-context understanding. 
It achieves SOTA scores on DUDE (72.4), MMLongBench (74.8), and MMLongBench-Doc (61.4). 
In document and chart understanding, Seed2.0 Pro also leads on ChartQAPro (71.2) and OmniDocBench 1.5, 
proving its efficacy in processing complex, information-dense visual inputs.

In summary, Seed2.0 Pro delivers state-of-the-art performance across a diverse array of visual-language tasks, 
particularly excelling in mathematics reasoning, perception proficiency, spatial reasoning, and long-context understanding, 
while Seed2.0 Lite and Mini offer competitive efficiency-focused alternatives.\vspace{5pt}

\begin{table}[t!]
\caption{Performance of Seed2.0 on public video understanding benchmarks compared to previous models.
The highest score in each benchmark is marked in bold, and the second is underlined. 
% Results marked with an $*$ are sourced from the technical report.
% Results marked with an $\dagger$ are evaluated with `VideoCut' as mentioned in \cite{seed18_modelcard}.
For benchmarks marked with a $\ddagger$, we include subtitles for evaluation. Results marked with an $^*$ are sourced from the technical report.
}
\vspace{5pt}
\label{table:video}
\newcommand{\prevbest}[1]{\scriptsize{(#1)}}
\newcommand{\best}[1]{\textbf{#1}}
\newcommand{\sbest}[1]{\uline{#1}}
\resizebox{\textwidth}{!}{
\tablestyle{6pt}{1.3}
\begin{tabular}{llc|ccc|ccc}
\toprule
\textbf{Capability} & \textbf{Benchmark}  & \textbf{Human} &  \textbf{\begin{tabular}[c]{@{}c@{}}Gemini-3-\\Pro\end{tabular}} & \textbf{\begin{tabular}[c]{@{}c@{}}Gemini-3-\\Flash\end{tabular}} & \textbf{Seed1.8} & \textbf{\begin{tabular}[c]{@{}c@{}}Seed2.0\\Mini\end{tabular}} & \textbf{\begin{tabular}[c]{@{}c@{}}Seed2.0\\Lite\end{tabular}} & \textbf{\begin{tabular}[c]{@{}c@{}}Seed2.0\\Pro\end{tabular}} \\ \midrule
\multirow{3}{*}{Knowledge}  & VideoMMMU~\cite{videommmu}   &  74.4 &  \sbest{87.6}$^*$ & \best{88.1}  & 82.7 &  80.6 & 84.1  & 86.9 \\
                            & MMVU~\cite{MMVU}      & 49.7 & 76.3 & \sbest{77.9} & 73.1  & 69.0 & 75.0 & \best{78.2} \\
                            & VideoSimpleQA~\cite{videosimpleqa} & - & \best{71.9} & \sbest{70.7}  & 67.8  & 67.7 & 66.6 & \best{71.9} \\
\midrule
\multirow{4}{*}{Reasoning}  & VideoReasonBench~\cite{videoreasonbench} & 73.8 &  59.5 & 61.2 & 52.8  & 40.5 & \sbest{64.2} & \best{77.8}\\
                            & Morse-500~\cite{Morse500} & 55.4 & \sbest{33.0} & 32.4 & 29.2 & 32.2 & 32.2 & \best{37.4} \\
                            % & VCRBench~\cite{VCRBench} & {53.4} & 51.4 & 52.5 & \sbest{59.8}  & TBD & TBD & \best{64.2} \\
                            & VideoHolmes{$^\ddagger$}~\cite{VideoHolmes} & - & {64.2} & \sbest{65.6} & {65.5}  & 58.6 & 63.8 & \best{67.4} \\
                            & Minerva{$^\ddagger$}~\cite{minerva} & - & \sbest{65.0} & 64.4 & 62.4  & 54.7 & 63.8 & \best{66.5} \\
\midrule
\multirow{6}{*}{\begin{tabular}[c]{@{}c@{}}Motion\\\& \\Perception\end{tabular}} 
                        & TVBench~\cite{tvbench} & 94.8 & {71.1} & 69.6 & \sbest{71.5}  & 70.5 & \sbest{71.5} & \best{75.0} \\
                        & ContPhy~\cite{ContPhy} & - & {58.0} & \sbest{60.5} & 54.9 & 55.9 & 56.1 & \best{67.4}\\
                        & TempCompass~\cite{TempCompass} & 97.3 & {88.0} & \sbest{88.3} & {86.9}  & 83.7 & 87.0 & \best{89.6} \\ 
                        % & MVBench~\cite{MVBench}       & 69.0 & \best{77.3} & 73.1  & \sbest{76.5}\\
                        & EgoTempo~\cite{egotempo} & 63.2 & {65.4} & 58.4 & {67.0}  & \sbest{67.2} & 61.8 & \best{71.8} \\
                        & MotionBench~\cite{motionbench}  & - & {70.3}$^*$ & 68.9 & {70.6}  & 64.4 & \sbest{70.9} & \best{75.2} \\
                        & TOMATO~\cite{TOMATO} & 95.2 & {59.6} & \best{60.8} &  \best{60.8}  & 47.4 & 57.3 & \sbest{59.9} \\
                        &$+$ \textit{Thinking with Tracking} & 95.2 & - & \sbest{64.0} &   61.0 & 51.3 & {59.2} & \best{65.3} \\
                        % & Countix~\cite{countix0}  & - & 18.6 & 18.7 & 30.2 & \sbest{31.0}  & TBD & TBD & \best{36.4} \\ 
\midrule
\multirow{5}{*}{Long Video} & VideoMME{$^\ddagger$}~\cite{videomme}  & - & \sbest{88.4}$^*$ & 85.2  &  {87.8}  & 81.2 & 87.7 & \best{89.5} \\
                            & CGBench~\cite{CGBench}  & -  & \best{65.5} & \sbest{65.3} &  62.4  & 59.2 & 59.3 & 65.0 \\
                            & LongVideoBench~\cite{longvideobench}  & - & 76.7 & 74.5 & \sbest{77.4} & 74.8 & 77.3 & \best{80.3} \\
                            & VideoEval-Pro~\cite{VideoEvalPro}  & - & \best{52.7} & \sbest{51.9} & 45.9 & 43.7 & 44.3 & 48.0 \\
                            & LVBench~\cite{lvbench}  & - & - & - & \sbest{73.0}  & 66.6 & \sbest{73.0} & \best{76.4} \\
                            
\midrule
\multirow{1}{*}{Multi Video}  & CrossVid~\cite{crossvid}  & 89.2 & 53.0 & 48.7  & 57.3 & \sbest{58.6} & 57.7 & \best{60.3}
\\
\midrule
\multirow{5}{*}{Streaming}  & OVBench~\cite{ovbench}  & - & {62.7} & {59.2}  & {65.1}  & 60.1 & \sbest{65.5} & \best{69.2} \\
                            & LiveSports-3K~\cite{livecc}  & - & 74.5 & {71.5} & {77.5}  & 73.3 & \sbest{77.8} & \best{78.0} \\
                            & OVOBench~\cite{ovobench}  & 92.8 & 70.1 & {68.7} & {72.6}  & 70.4 & \sbest{76.7} & \best{77.0} \\
                            % & StreamBench~\cite{streambench}   & 63.7 & - & \best{72.8} & 68.4 & \sbest{70.0} \\
                            & ODVBench~\cite{odvbench}  & 91.4 & 63.6 & {56.7} & {63.5} & {65.1} & \sbest{69.6} & \best{72.5} \\
                            & ViSpeak~\cite{vispeak}  & 96.0 & \best{89.0}  & \sbest{86.0} & 79.0  & 77.5 & {84.0} & {78.5} \\
% \hline
% \multirow{2}{*}{Grounding}  & TimeLens-Charades~\cite{timelens}  & - & {49.4} & {51.2} &  \sbest{59.7} & 51.8 & 56.1 & \textbf{61.1} \\
%                             & TimeLens-ActivityNet~\cite{timelens}  & - & {49.6} & {52.1} &  \sbest{60.1} & 51.4 & 58.0 & \textbf{61.4} \\

\bottomrule
\end{tabular}}
\vspace{5pt}
\end{table}

\begin{table}[t]
\caption{Performance of Seed2.0 with video tool-use on long-form video understanding and reasoning. We compare the performance of Seed2.0 when using the VideoCut tool across different benchmarks.}
\label{table:video_tooluse}
\vspace{5pt}
\newcommand{\prevbest}[1]{\scriptsize{(#1)}}
\newcommand{\best}[1]{\textbf{#1}}
\newcommand{\sbest}[1]{\uline{#1}}
\resizebox{\textwidth}{!}{
\tablestyle{5pt}{1.2}
\begin{tabular}{llccc|cc}
\toprule
\textbf{Benchmark} & \textbf{\begin{tabular}{c}Average\\Duration\end{tabular}} & \textbf{\begin{tabular}{c}Gemini-3-\\Pro\end{tabular}} & \textbf{\begin{tabular}{c}Seed1.8\end{tabular}} & \textbf{\begin{tabular}{c}Seed1.8\\w/ VideoCut\end{tabular}} & \textbf{\begin{tabular}{c}Seed2.0\end{tabular}} & \textbf{\begin{tabular}{c}Seed2.0 Pro\\w/ VideoCut\end{tabular}} \\
\midrule
CGBench~\cite{CGBench} & 1624 seconds & {65.5} & 62.4 & \sbest{65.9} & 65.0 & \best{66.8} \\
LVBench~\cite{lvbench} & 4104 seconds & - & 73.0 & \sbest{78.9} & 76.4 & \best{80.0} \\
\midrule
ZeroVideo & 1672 seconds & {14.3} & 6.9  & \sbest{18.8} & 14.5 & \best{27.9} \\
\bottomrule
\end{tabular}}
\end{table}

\subsection{Video Task Evaluation}
We conduct a comprehensive evaluation of the Seed2.0 family (including Seed2.0 Mini, Seed2.0 Lite, and Seed2.0 Pro) across multiple dimensions of video capabilities, including knowledge, reasoning, perception, motion understanding, as well as long-term, multi-video, and streaming video analysis.

As shown in Table~\ref{table:video}, Seed2.0 pushes the performance frontier for video understanding and delivers state-of-the-art results on multiple benchmarks, with exceptional performance in motion perception, reasoning, and streaming video understanding.\vspace{5pt}

\textbf{Video Knowledge.}
In terms of knowledge, Seed2.0 achieves leading performance on the multidisciplinary benchmark MMVU~\cite{MMVU}, and delivers results comparable to Gemini-3-Pro on VideoMMMU~\cite{videommmu} and VideoSimpleQA~\cite{videosimpleqa}, representing a substantial improvement over the previously released Seed1.8~\cite{seed18_modelcard}. In addition, it is worth noting that our lightweight model, Seed2.0 Mini, also performs strongly.\vspace{5pt}

\textbf{Video Reasoning.}
In video reasoning, Seed2.0 Pro continues to push the performance frontier. On VideoReasonBench~\cite{videoreasonbench}, Seed2.0 surpasses human performance, with particularly strong results in visual state tracking, widely regarded as one of the most fundamental capabilities underlying video reasoning.

On Morse-500~\cite{Morse500}, a highly challenging reasoning benchmark covering video abstract reasoning, physical reasoning, and planning reasoning, Seed2.0 further achieves a new state of the art, reaching 37.4\% accuracy. Despite this progress, a substantial gap to human-level performance remains, and future iterations of the model will continue to close this gap.

\textbf{Video Perception and Motion Understanding.}
\label{sec:videomotion}
Foundational video perception is critical to video understanding and reasoning. Among these perceptual competencies, motion understanding is particularly crucial for modeling temporal state transitions and tracking dynamic changes over time. Seed2.0 further strengthens its motion perception and understanding capabilities. As summarized in Table~\ref{table:video}, Seed2.0 Pro attains state-of-the-art performance on the majority of benchmarks and substantially outperforms Seed1.8 and Gemini-3-Pro across several evaluations.

For TOMATO~\cite{TOMATO}, we explore a \textbf{Thinking with Tracking} strategy in which we prompt to encourage the model to explicitly output per-frame bounding boxes for the moving target during reasoning, and then identifies the motion type based on the target's trajectory. This strategy is effective for both the Gemini and Seed models, and yields meaningful gains on TOMATO.
Nevertheless, performance on TOMATO and TVBench suggests that a considerable gap to human-level proficiency remains. We will continue to advance motion perception in subsequent iterations to further narrow this gap.

\begin{table}[!t]
\caption{Evaluation on Fundamental Agentic Capacity Benchmarks (\textbf{Large} Models). The highest score is marked in bold, and the second is underlined. Some scores differ greatly from the evaluation results in the tech reports by other organizations. The scores in parentheses represent the results under the aligned settings then.}
\label{tab:agent_large}
\centering
\begin{threeparttable}
\resizebox{0.88\textwidth}{!}{
\begin{tabular}{cl|cccc|c}
\toprule
\textbf{Capability} & \textbf{Benchmark} & \begin{tabular}[c]{@{}c@{}}\textbf{GPT-5.2}\\ \textbf{High}\end{tabular} & \begin{tabular}[c]{@{}c@{}}\textbf{Claude-}\\ \textbf{Sonnet-4.5}\end{tabular} & \begin{tabular}[c]{@{}c@{}}\textbf{Claude-}\\ \textbf{Opus-4.5}\end{tabular} & \begin{tabular}[c]{@{}c@{}}\textbf{Gemini-3-}\\\textbf{pro High}\end{tabular} & \begin{tabular}[c]{@{}c@{}}\textbf{Seed2.0 }\\\textbf{Pro}\end{tabular} \\ 
\midrule
\multirow{14}{*}{\begin{tabular}[c]{@{}l@{}} Coding\\ Agent\end{tabular}} 
 % & Terminal Bench 1.0\tnote{1} & \textbf{71.0} & {\ul 68.1} & \textbf{71.0} & 66.7 & 70.6 \\
 % & Terminal Bench 1.0\tnote{2} & 66.7 & {\ul 72.5} & \textbf{73.9} & {\ul 72.5} & 68.1 \\
 & Terminal Bench 2.0\tnote{1} & \textbf{62.4} & 45.2 & {\ul 60.2} & 56.9 & 55.8 \\
 % & Terminal Bench 2.0\tnote{2} & \textbf{64.7} & 55.3 & 56.0 & 58.8 & {\ul 57.7} \\
 & SWE-Lancer & 48.9 & 45.7 & \textbf{56.1} & 44.3 & {\ul 49.4} \\
 & SWE Bench Verified & {\ul 80.0} & 77.2 & \textbf{80.9} & 76.2 & 76.5 \\
 & Multi-SWE-Bench & 47.7 & 47.7 & \textbf{52.8} & {\ul 50.2} & 45.2 \\
 & SWE-Bench Pro & \textbf{55.6} & 48.4 & {\ul 55.4} & 49.7 & 46.9 \\
 & SWE Multilingual & 68.8 & 64.1 & \textbf{74.0} & {\ul 72.7} & 71.7 \\
 & Scicode & 49.7 & 47.9 & {\ul 52.8} & \textbf{57.7} & 48.5 \\
 & SWE-Evo & 12.5 & {\ul 16.7} & \textbf{27.1} & 8.9 & 8.5 \\
 & Aider Polyglot & 91.1 & 82.2 & {\ul 92.4} & \textbf{94.2} & 80.0 \\
 & ArtifactsBench & \textbf{71.1} & 59.1 & {\ul 68.5} & 58.4 & 66.6 \\
 & CodeSimpleQA & {\ul 62.3} & 59.6 & \textbf{63.0} & 54.7 & 58.0 \\
 & SpreadsheetBench Verified & 69.9 & 75.9 & {\ul 78.6} & 70.8 & \textbf{79.1} \\
\midrule
\multirow{7}{*}{\begin{tabular}[c]{@{}l@{}} Search\\ Agent\end{tabular}} 
 & BrowseComp & \textbf{77.9 (65.3)} & 43.9 (29.5) & 67.8 (57.2) & 59.2 & {\ul 77.3} \\
 & BrowseComp-zh & {\ul 76.1} & 42.4 & 62.4 & 66.8 & \textbf{82.4} \\
 & HLE-text & 45.5 & 32.0 & 43.2 & {\ul 46.9} & \textbf{54.2} \\
 & HLE-Verified & {\ul 68.5} & 37.6 & 56.6 & 67.5 & \textbf{73.6} \\
 % & xbench-DeepSearch & \textbf{77.8} & 66.0 & 59.8 & 58.9 & {\ul 74.0} \\
 & WideSearch & \textbf{76.8} & 65.1 & {\ul 76.2 (71.7)} & 67.3 & 74.7 \\
 & FinSearchComp & \textbf{73.8} & 58.6 & 66.2 & 52.7 & {\ul 70.2} \\
 & DeepSearchQA & 71.3 (66.4) & 36.3 & {\ul 76.1 (41.6)} & 63.9 & \textbf{77.4} \\
 & Seal-0 & {\ul 51.4} & \textbf{53.4} & 47.7 & 45.5 & 49.5 \\
\midrule
\multirow{5}{*}{\begin{tabular}[c]{@{}l@{}} Tool Use\end{tabular}} 
 & $\tau^2$-Bench (retail) & 82 & 86.2 & {\ul{88.9}} &  85.3 & \textbf{90.4} \\
 & $\tau^2$-Bench (telecom) & \textbf{98.7} & 98.0 & {\ul 98.2} & 98.0 & 94.2 \\
 & MCP-Mark & \textbf{57.5} & 32.1 & 42.3 &  53.9 & {\ul 54.7} \\
 & BFCL-v4 & 65.9 & 72.9 & \textbf{76.5} & 71.0 & {\ul 73.4} \\
 & VitaBench & 41.8 & 40.8 & \textbf{55.3} & {\ul 48.8} & 47.0 \\
\midrule
\multirow{3}{*}{\begin{tabular}[c]{@{}l@{}} Deep\\ Research\end{tabular}} 
 & DeepConsult & 54.3 & 55.8 & {\ul61.0} & 48.0 &  \textbf{61.1} \\
 & DeepResearchBench & {\ul 52.2} & 47.2 & 50.6 & 49.6 & \textbf{53.3} \\
 & ResearchRubrics & 42.3 & 38.6 & {\ul 45.0} & 37.7 & \textbf{50.7} \\
\midrule
\multirow{3}{*}{\begin{tabular}[c]{@{}l@{}} Vision\\ Agent\end{tabular}} 
 & Minedojo-Verified & 18.3 & - & - & {\ul 23.3} & \textbf{49.0} \\
 & MM-BrowseComp & {\ul 26.3} & - & - & 25.0 & \textbf{48.8} \\
 & HLE-VL & 31.0 & - & - & {\ul 36.0} &  \textbf{39.2} \\
\bottomrule
\end{tabular}}
\begin{tablenotes} 
    \footnotesize 
    \item[1] Using Terminus2.
    % \item[2] Using Codeact.
    % \item[2] Results for Claude and Gemini are from \url{https://www.arxiv.org/pdf/2601.20975}
\end{tablenotes}
\end{threeparttable}
\end{table}

\textbf{Long Video Understanding.}
For long-video understanding, Seed2.0 Pro achieves a breakthrough performance of 89.5 on VideoMME~\cite{videomme}. Seed2.0 also delivers strong results on LVBench and LongVideoBench, showing substantial improvements over Seed1.8 across these benchmarks. In addition, the entire Seed2.0 model family is equipped with VideoCut tool-use capabilities by default to improve long video reasoning.

\textbf{Multiple Video Understanding.}
Multi-video understanding requires models to identify and integrate critical evidence across multiple video contexts, and is a core capability for real-world applications as well as video agents. On CrossVid~\cite{crossvid}, a benchmark for multi-video understanding, Seed2.0 Pro achieves a new state-of-the-art score of 60.3. Looking ahead, we will prioritize further improvements in multi-video understanding, with an emphasis on strengthening cross-context reasoning.

\textbf{Streaming.}
Moving beyond static analysis, streaming video understanding and interactive reasoning enable systems to comprehend and respond to visual inputs in real-time, providing key capabilities for interactive products (\textit{e.g.}, Doubao video calling).
% Streaming video understanding and interactive reasoning serving as key enabling capabilities for interactive products (\textit{e.g.}, Doubao video calling). 
Across multiple benchmarks, Seed2.0 Pro delivers further improvements, while Seed2.0 Lite surpasses Gemini-3-Pro/Flash in several evaluations with high efficiency.

\textbf{Video Tool-Use: VideoCut.}
Seed2.0 also further enhances video tool-use capability, \textit{i.e.}, VideoCut. When processing long videos or tasks that require high-frame-rate perception, it can adopt VideoCut to replay relevant segments at a higher FPS, which is an intuitive and effective mechanism for video understanding and reasoning. As shown in Table~\ref{table:video_tooluse}, we evaluate the performance of enabling VideoCut for Seed2.0 Pro. With VideoCut, Seed2.0 Pro further raises the ceiling of long-video understanding, yielding substantial improvements on CGBench, LVBench, and the challenging ZeroVideo, which consists of challenging, real-world video reasoning scenarios, including fine-grained high-frame-rate motion perception and long-form, multi-hop reasoning over extended videos.

\subsection{Fundamental Agentic Capacity}
\label{sec:agent_results}

% \input{tables/tab_agent_large}

% \newpage
\begin{table}[!t]
\caption{Evaluation on Fundamental Agentic Capacity Benchmarks (\textbf{Efficient} Models). The highest score is marked in bold, and the second is underlined.}
\label{tab:agent_small}
\centering
\small
\begin{threeparttable}
\begin{tabular}{cl|cc|c}
\toprule
\textbf{Capability} & \textbf{Benchmark} & \begin{tabular}[c]{@{}c@{}}\textbf{GPT-5-}\\\textbf{mini High}\end{tabular} & \begin{tabular}[c]{@{}c@{}}\textbf{Gemini-3-}\\\textbf{Flash High}\end{tabular} & \begin{tabular}[c]{@{}c@{}}\textbf{Seed2.0 }\\\textbf{Lite}\end{tabular} \\ 
\midrule
\multirow{14}{*}{\begin{tabular}[c]{@{}l@{}} Coding\\ Agent\end{tabular}} 
 % & Terminal Bench 1.0\tnote{1} & 40.6 & \textbf{70.6} & {\ul 56.3} \\
 % & Terminal Bench 1.0\tnote{2} & 56.5 & \textbf{70.6} & {\ul 64.2} \\
 & Terminal Bench 2.0\tnote{1} & 36.9 & \textbf{60.0} & {\ul 45.0} \\
 % & Terminal Bench 2.0\tnote{2} & 44.0 & \textbf{57.1} & {\ul 48.2} \\
 & SWE-Lancer & 43.1 & \textbf{51.7} & {\ul 47.1} \\
 & SWE Bench Verified & 67.9 & \textbf{78.0} & {\ul 73.5} \\
 & Multi-SWE-Bench & {\ul 49.3} & \textbf{59.0} & 41.1 \\
 & SWE-Bench Pro & \textbf{51.7} & 46.7 & {\ul 46.0} \\
 & SWE Multilingual & 63.8 & \textbf{71.1} & {\ul 64.4}\\
 & Scicode & 40.2 & \textbf{55.0} & {\ul 52.4} \\
 & SWE-Evo & 10.4 & \textbf{12.8} & {\ul 10.6} \\
 & Aider Polyglot & {\ul 79.1} & \textbf{92.0} & 76.0 \\
 & ArtifactsBench & \textbf{68.9} & 52.5 & {\ul 62.6} \\
  & CodeSimpleQA & \textbf{57.6} &{\ul 53.7} & 53.1 \\
 & SpreadsheetBench Verified & 58.1 & {\ul 65.7} & \textbf{82.3} \\
\midrule
\multirow{7}{*}{\begin{tabular}[c]{@{}l@{}} Search\\ Agent\end{tabular}} 
 & BrowseComp & {\ul 48.1} & 41.5 & \textbf{72.1} \\
 & BrowseComp-zh & 49.5 & {\ul 63.0} & \textbf{82.0} \\
 & HLE-text & 35.8 & {\ul 47.6} & \textbf{49.5} \\
 & HLE-Verified & 56.4 & \textbf{71.8} & {\ul 70.7} \\
 & WideSearch & 37.7 & {\ul 64.0} & \textbf{74.5} \\
 % & xbench-DeepSearch & {\ul 72.2} & 57.0 & \textbf{76.0} \\
 & FinSearchComp & 38.1 & {\ul 54.8} &  \textbf{65.1}\\
 & DeepSearchQA & 16.7 & {\ul 54.7} & \textbf{67.7} \\
 & Seal-0 & 34.2 & {\ul 37.8} & \textbf{52.3} \\ 

\midrule
\multirow{5}{*}{Tool Use} 
 & $\tau^2$-Bench (retail) & - & {\ul 88.6} & \textbf{90.9} \\ 
 & $\tau^2$-Bench (telecom) & - & \textbf{94.7} &  {\ul 92.1} \\ 
 & MCP-Mark & 30.2 & {\ul 40.5} & \textbf{46.7} \\
 & BFCL-v4 & 57.9 & {\ul 65.0} &  \textbf{72.9} \\
 & VitaBench & 20.8 & \textbf{46.7} & {\ul 41.8} \\
\midrule
\multirow{3}{*}{\begin{tabular}[c]{@{}l@{}} Deep\\ Research\end{tabular}} 
 & DeepConsult & {\ul 49.8} & 26.0 &  \textbf{60.3}\\
 & DeepResearchBench & {\ul 50.7} & 46.1 & \textbf{54.4} \\
 & ResearchRubrics & {\ul 43.6} & 36.9 & \textbf{50.8} \\
\midrule
\multirow{3}{*}{\begin{tabular}[c]{@{}l@{}} Vision\\ Agent\end{tabular}} 
 & Minedojo Verified & 20.3 & {\ul 24.3} & \textbf{39.7} \\
 & MM-BrowseComp & 17.9 & {\ul 22.8} & \textbf{45.1} \\
 & HLE-VL & 18.7 & \textbf{36.8} & {\ul 35.8} \\
\bottomrule
\end{tabular}
\begin{tablenotes} 
    \footnotesize 
    \item[1] Using Terminus2.
    % \item[2] Using Codeact.
\end{tablenotes}

\end{threeparttable}
% \vspace{-6pt}
\end{table}

We evaluate Seed2.0 (Pro/Lite) on a diverse set of agentic benchmarks covering search agents, deep research, vision agents, coding agents, and tool use. \Cref{tab:agent_large,tab:agent_small} report detailed results.\vspace{5pt}

\textbf{Overall}, Seed2.0 Pro lands in the top tier across agentic capabilities, with clear advantages on search, deep research, and vision agent tasks---the workloads that matter most for high-frequency user scenarios like information seeking, complex reasoning, and multimodal interaction.\vspace{5pt}

\textbf{Search, Deep Research, and Vision Agents.}
Seed2.0 Pro consistently leads or sits near the top on search and research-oriented benchmarks: HLE, BrowseComp, WideSearch and DeepSearchQA. On deep research tasks, it posts the best results on Deep Research and Research Rubrics. In vision-agent settings, Seed2.0 Pro pulls ahead of reported baselines by a wide margin on Minedojo-Verified, MM-BrowseComp, and HLE-VL---evidence of strong multimodal grounding for action-oriented tasks. Taken together, these numbers place Seed2.0 Pro at or near state-of-the-art on search, deep research, and vision agents.\vspace{5pt}

\textbf{Tool Use and Coding Agents.}
Tool use and coding-agent performance is solidly first-tier. Seed2.0 Pro holds its own on SWE-Bench Pro and SWE Bench Verified, and tops SpreadSheetBench for structured artifact manipulation. We also see distinctive gains on SWE-Evo, pointing to robust evolutionary code improvement. Beyond this, Seed2.0 Pro also delivers strong performance on general tool use tasks such as MCP-Mark, BFCL-v4 and $\tau^2$-Bench. That said, complex API orchestration and long-horizon code execution still leave room to grow.\vspace{5pt}

\textbf{Efficient Agentic Deployment.}
Among small models, Seed2.0 Lite consistently ranks first or second on efficient agentic benchmarks. It performs well on search and deep research tasks and posts excellent SpreadsheetBench numbers. Paired with significantly lower inference cost, Lite offers a practical quality--cost trade-off for large-scale and latency-sensitive deployments.\vspace{5pt}

\begin{table}[!t]
\caption{Evaluation on Advanced Economically and Scientifically Valuable Tasks (\textbf{Large} Models). The highest score is marked in bold, and the second is underlined.}
\label{tab:advanced_large}
\vspace{5pt}
\centering
\begin{threeparttable}
\resizebox{0.86\textwidth}{!}{
\begin{tabular}{cl|cccc|c}
\toprule
\textbf{Capability} & \textbf{Benchmark} & \begin{tabular}[c]{@{}c@{}}\textbf{GPT-5.2}\\ \textbf{High}\end{tabular} & \begin{tabular}[c]{@{}c@{}}\textbf{Claude-}\\ \textbf{Sonnet-4.5}\end{tabular} & \begin{tabular}[c]{@{}c@{}}\textbf{Claude-}\\ \textbf{Opus-4.5}\end{tabular} & \begin{tabular}[c]{@{}c@{}}\textbf{Gemini-3-}\\\textbf{pro High}\end{tabular} & \begin{tabular}[c]{@{}c@{}}\textbf{Seed2.0 }\\\textbf{Pro}\end{tabular} \\ 
\midrule
\multirow{5}{*}{\begin{tabular}[c]{@{}l@{}} Science\\ Discovery\end{tabular}} 
 & Scicode & 49.7 & 47.9 & {\ul 52.8} & \textbf{57.7} & 52.1 \\
 & FrontierSci-research & \textbf{25.0} & 16.7 & 21.7 & 15.0 & {\ul 23.3} \\
 & Superchem (text-only) & {\ul 58.0} & 32.4 & 43.2 & \textbf{63.2} & 53.0 \\
 & BIObench & \textbf{58.1} & 44.7 & 49.3 & 51.3 & {\ul 53.5} \\
 & AInstein Bench & 41.3 & 33.7 & {\ul 44.0} & 42.8 & \textbf{47.7} \\
\midrule
\multirow{6}{*}{\begin{tabular}[c]{@{}l@{}} Vibe\\ Coding\end{tabular}} 
 & NL2Repo-Bench & \textbf{49.3} & 39.9 & {\ul 43.2} & 34.2 & 27.9 \\
 & NL2Repo (Pass@1) & \textbf{8.0} & 3.0 & 3.0 & {\ul 4.0} & 3.0 \\
 & ArtifactsBench & \textbf{71.1} & 59.1 & {\ul 68.5} & 58.4 & 66.6 \\
 & SWE-Bench Pro & \textbf{55.6} & 48.4 & {\ul 55.4} & 49.7 & 46.9 \\
 & Terminal Bench 2.0\tnote{1} & \textbf{62.4} & 45.2 & {\ul 60.2} & 54.2 & 55.8 \\
 % & Terminal Bench 2.0\tnote{2} & \textbf{64.7} & 55.3 & 56.0 & {\ul 58.8} & 51.2 \\
\midrule
\multirow{5}{*}{\begin{tabular}[c]{@{}l@{}} Context\\ Learning\end{tabular}} 
 & KORBench & \textbf{79.2} & 73.0 & {\ul 77.4} & 73.9 & 77.2 \\
 & DeR$^2$ Bench & \textbf{69.0} & 58.9 & 60.4 & {\ul 66.1} & 58.2 \\
 & CL-Bench & \textbf{23.9} & 18.1 & {\ul 22.6} & 15.6 & 21.5 \\
 & ToB-Complex Workflows & 45.0 & 61.0 & {\ul 64.8} & \textbf{69.2} &  64.7\\
 & ToB-Reference Q\&A & 63.6 & 58.9 & 67.9 & {\ul 68.3} & \textbf{72.4} \\
\midrule
\multirow{9}{*}{\begin{tabular}[c]{@{}l@{}} Real World\\ Tasks\end{tabular}} 
 & HealthBench - Hard & \textbf{36.6} & 10.9 & 11.0 & 15.0 & {\ul 28.3} \\
 & GDPVal-Diamond & \textbf{26.9} & 15.2 & 20.7 & 19.4 & {\ul 21.3} \\
 & XPert Bench & {\ul 53.3} & 44.7 & 50.5 & 53.1 & \textbf{64.5} \\
 & ToB-K12 Education & {\ul 61.6} & 50.1 & 56.2 & 59.4 &  \textbf{62.8}\\
 & ToB-Compositional Tasks & 51.5 & 57.3 & {\ul 63.6} & \textbf{64.8} & 59.1\\
 & ToB-Text Classification & 62.1 & 64.5 & 63.9 & {\ul 67.5} & \textbf{69.0} \\
 & ToB-Information Extraction & 44.7 & 48.3 & {\ul 50.1} & 49.0 & \textbf{52.0} \\
 & World Travel (VLM) & \textbf{19.33} & 2.67 & {\ul 14.0} & 8.0 & 12.0 \\
 & World Travel (TEXT) & \textbf{32.67} & 10.0 & 21.3 & 14.7 & {\ul 23.3} \\
\bottomrule
\end{tabular}}
\begin{tablenotes} 
    \footnotesize 
    \item[1] Using Terminus2.
    % \item[2] Using Codeact.
\end{tablenotes}
\end{threeparttable}
% \vspace{-8pt}
\end{table}
\begin{table}[t]
\caption{Evaluation on Advanced Economically and Scientifically Valuable Tasks (\textbf{Efficient} Models). The highest score is marked in bold, and the second is underlined.}
\label{tab:advanced_small}
\centering
\small
\begin{threeparttable}

\begin{tabular}{cl|cc|c}
\toprule
\textbf{Capability} & \textbf{Benchmark} & \begin{tabular}[c]{@{}c@{}}\textbf{GPT-5-}\\\textbf{mini High}\end{tabular} & \begin{tabular}[c]{@{}c@{}}\textbf{Gemini-3-}\\\textbf{Flash High}\end{tabular} & \begin{tabular}[c]{@{}c@{}}\textbf{Seed2.0 }\\\textbf{Lite}\end{tabular} \\ 
\midrule
\multirow{5}{*}{\begin{tabular}[c]{@{}l@{}} Science\\ Discovery\end{tabular}} 
 & Scicode & 40.2 & \textbf{55.0} & {\ul 52.4} \\
 & FrontierSci-research & \textbf{18.3} & {\ul 11.7} & \textbf{18.3} \\
 & Superchem (text-only) & 34.8 & \textbf{54.4} & {\ul 48.0} \\
 & BIObench & 49.2 & \textbf{55.2} & {\ul 50.2} \\
 & AInstein Bench & 35.0 & \textbf{44.0} &  {\ul 38.3} \\
\midrule
\multirow{6}{*}{\begin{tabular}[c]{@{}l@{}} Vibe\\ Coding\end{tabular}} 
 & NL2Repo-Bench & 19.5 & \textbf{27.6} &  {\ul 24.6} \\
 & NL2Repo (Pass@1) & \textbf{2.0} & \textbf{2.0} &  1.0\\
 & ArtifactsBench & \textbf{68.9} & 52.5 & {\ul 62.6} \\
 & SWE-Bench Pro & \textbf{51.7} & {\ul 46.7} & 46.0 \\
 & Terminal Bench 2.0\tnote{1} & 36.9 & \textbf{60.0} & {\ul 45.0} \\
 % & Terminal Bench 2.0\tnote{2} & 44.0 & \textbf{57.1} & {\ul 48.2} \\ 
\midrule
\multirow{5}{*}{\begin{tabular}[c]{@{}l@{}} Context\\ Learning\end{tabular}} 
 & KORBench & 74.2 & {\ul 76.0} & \textbf{77.0} \\
 & DeR$^2$ Bench & 50.3 & \textbf{66.0} &  {\ul 57.3}\\
 & CL-Bench & \textbf{25.2} & 16.1 & {\ul 20.0} \\
 & ToB-Reference Q\&A & 53.4 & {\ul 64.9} & \textbf{68.2} \\
 & ToB-Complex Workflows & 42.5 & {\ul 61.3} & \textbf{62.0} \\
\midrule
\multirow{9}{*}{\begin{tabular}[c]{@{}l@{}} Real World\\ Tasks\end{tabular}} 
 & HealthBench - Hard & \textbf{38.6} & {\ul 21.5} & 20.0 \\
 & GDPVal-Diamond & {\ul 11.5} & 6.46 & \textbf{23.2} \\
 & XPert Bench & 47.6 & {\ul 50.1}& \textbf{63.3}\\
 & ToB-K12 Education & 53.7 & {\ul 59.0} &  \textbf{63.8}\\
 & ToB-Compositional Tasks &  47.5 & \textbf{57.7} & {\ul 54.8} \\
 & ToB-Text Classification & 52.7 & {\ul 61.9} & \textbf{64.5} \\
 & ToB-Information Extraction & 40.5 & {\ul 45.4} &  \textbf{48.4}\\
 & World Travel (VLM) & 2.7 & {\ul 7.3} & \textbf{14.0} \\
 & World Travel (TEXT) & {\ul 15.3} & 13.3 & \textbf{24.0} \\
\bottomrule
\end{tabular}
\begin{tablenotes} 
    \footnotesize 
    \item[1] Using Terminus2.
    % \item[2] Using Codeact.
\end{tablenotes}
\end{threeparttable}
\vspace{5pt}
\end{table}

\subsection{Advanced Economically and Scientifically Valuable Tasks}
\label{sec:advanced_results}

We further evaluate Seed2.0 on advanced tasks that reflect scientific discovery, economically valuable workflows, context-based learning, and real-world task completion. \Cref{tab:advanced_large,tab:advanced_small} report the detailed results.\vspace{5pt}

\textbf{Scientific Discovery.}
Seed2.0 Pro operates at the frontier. It posts strong numbers on FrontierSci-research and scientific coding benchmarks, and leads on AInstein Bench---signs of robust hypothesis-driven reasoning in research-style scenarios.\vspace{5pt}

\textbf{Real-World Economic Value.}
Beyond science, Seed2.0 Pro translates well to tasks with direct economic value. It tops XPert Bench and remains highly competitive on GDPVal and multiple ToB benchmarks. Notably, the model excels on user-oriented scenarios we explicitly optimize for: customer-service QA, information extraction, intent recognition, and K12 problem solving. These results reflect reliable instruction-following behavior in production.\vspace{5pt}

\textbf{Context Learning and Repository-Level Code.}
Clear headroom remains on context-driven learning and end-to-end vibe coding. Seed2.0 Pro trails the strongest baselines on DeR$^2$ Bench and NL2Repo-Bench, suggesting that long-horizon context integration and repository-level code generation are still challenging. We have flagged these as priority directions and are actively pushing targeted research.\vspace{5pt}

\textbf{Efficiency at Scale.}
Seed2.0 Lite holds up well on advanced tasks among efficient models. It posts strong results on ToB real-world benchmarks, solid scientific discovery numbers, and competitive context-learning outcomes. On several advanced benchmarks, Lite beats GPT-5-Mini outright while offering favorable latency and cost.

Across agentic capacity and advanced real-world tasks, Seed2.0 delivers consistent top-tier performance. Seed2.0 Pro excels in search, deep research, vision agents, and scientific discovery; Lite provides a compelling efficiency-focused alternative for large-scale deployment.
\section{Use Cases of Seed2.0}

In this section, we demonstrate the applicability of Seed2.0 across diverse real-world scenarios, illustrating how the model addresses practical challenges in vibe coding, application operation, and even scientific discovery.

\subsection{Vibe Coding}
In the "Vibe Coding" setting, the model is evaluated on end-to-end software engineering behavior under realistic constraints rather than isolated code snippets. Tasks include interpreting underspecified requirements, forming executable plans, producing runnable artifacts, and iteratively refining solutions based on feedback until acceptance criteria are met.

\begin{figure}[htbp]
  \centering
  % 用 trim 去边
  \includegraphics[width=0.88\textwidth,
                   trim=0 0 0 0,clip]{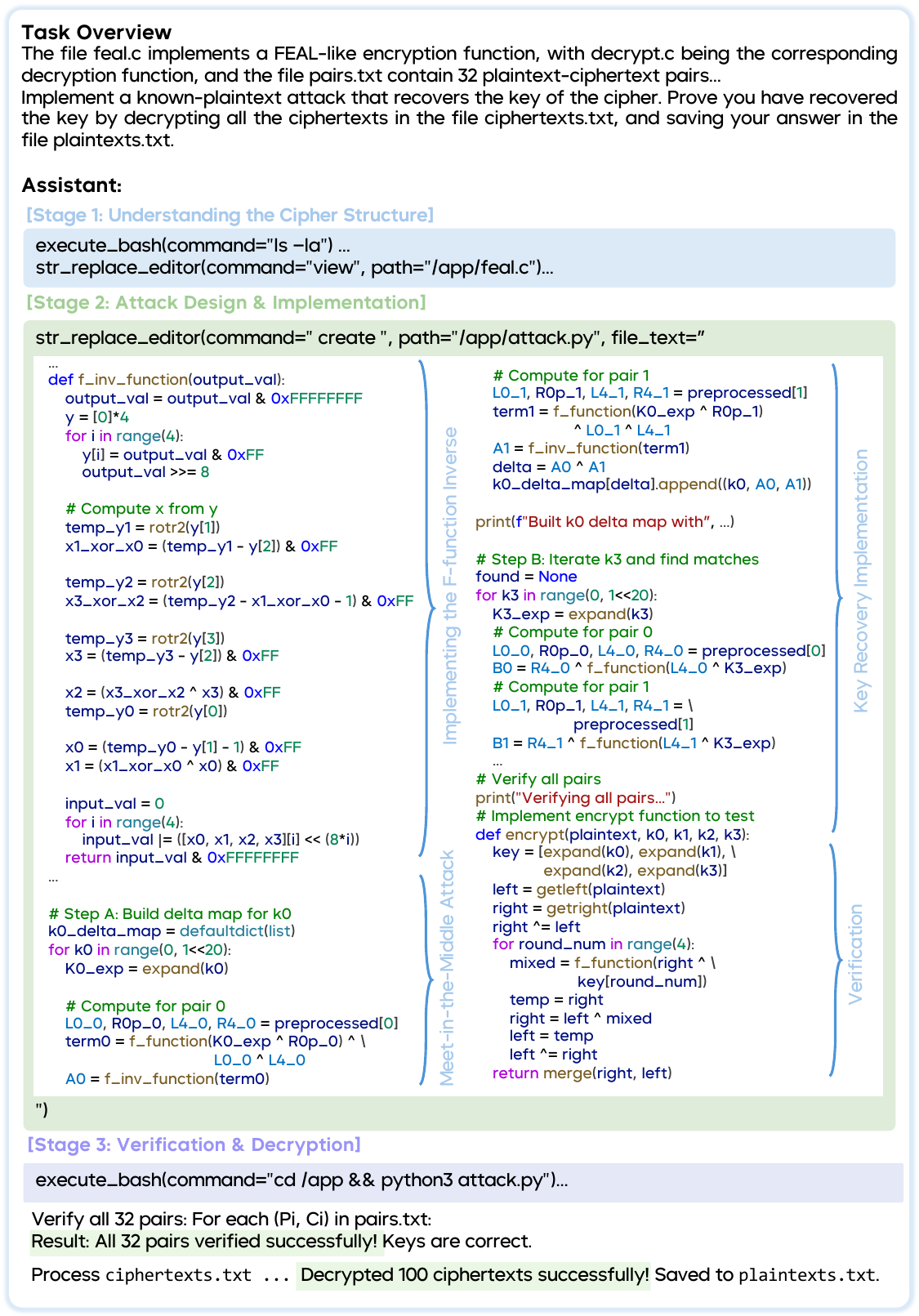}
    \caption{The case of Seed2.0 with Complex Code Generation Task on TerminalBench 2.0~\citep{merrill2026terminal}.}
  \label{fig:terminal}
\end{figure}

To characterize Seed2.0's capabilities in this regime, we present five representative case studies: algorithmically challenging code synthesis under compute constraints (\S~\ref{sec:complex-code-gen}), repository-level construction from requirements documents alone (\S~\ref{sec:project-responsitory-construction}), zero-regression debugging guided by unit tests (\S~\ref{sec:iterative-code-debugging}), cross-version upgrade debugging using release notes with full test suite validation (\S~\ref{sec:complex-iterative-debugging}), and competition-level programming under official limits (\S~\ref{sec:competition-programming}). These settings vary supervision completeness, engineering scope, and failure signal observability, enabling comprehensive assessment of planning, implementation, verification, and iterative correction in realistic development workflows.

\subsubsection{Complex Code Generation}
\label{sec:complex-code-gen}
To evaluate Seed2.0's capacity for complex code synthesis, we examine its performance on the "feal-linear-cryptanalysis" task from TerminalBench 2.0~\citep{merrill2026terminal}, classified as Hard. This task provides a simplified FEAL-like block cipher as a Feistel network, a reference decryption program, and multiple plaintext–ciphertext pairs. The objective is to construct a known-plaintext attack that recovers the secret key and decrypts additional ciphertexts. Naïve brute-force enumeration over the effective key space is computationally infeasible within benchmark constraints, requiring algorithmic insight rather than exhaustive search.

As shown in Figure~\ref{fig:terminal}, Seed2.0 completes this task in 12 interaction rounds through systematic repository analysis. It inspects the project structure, examines \texttt{feal.c} to reconstruct the round function, and studies documentation to align its implementation with expected formats and correctness criteria. This structured approach minimizes specification inconsistencies and informs development of \texttt{attack.py}, a cryptanalytic program tailored to the specific FEAL variant.

\begin{figure}[!t]
  \centering
  % 用 trim 去边
  \includegraphics[width=0.88\textwidth,
                   trim=0 0 0 0,clip]{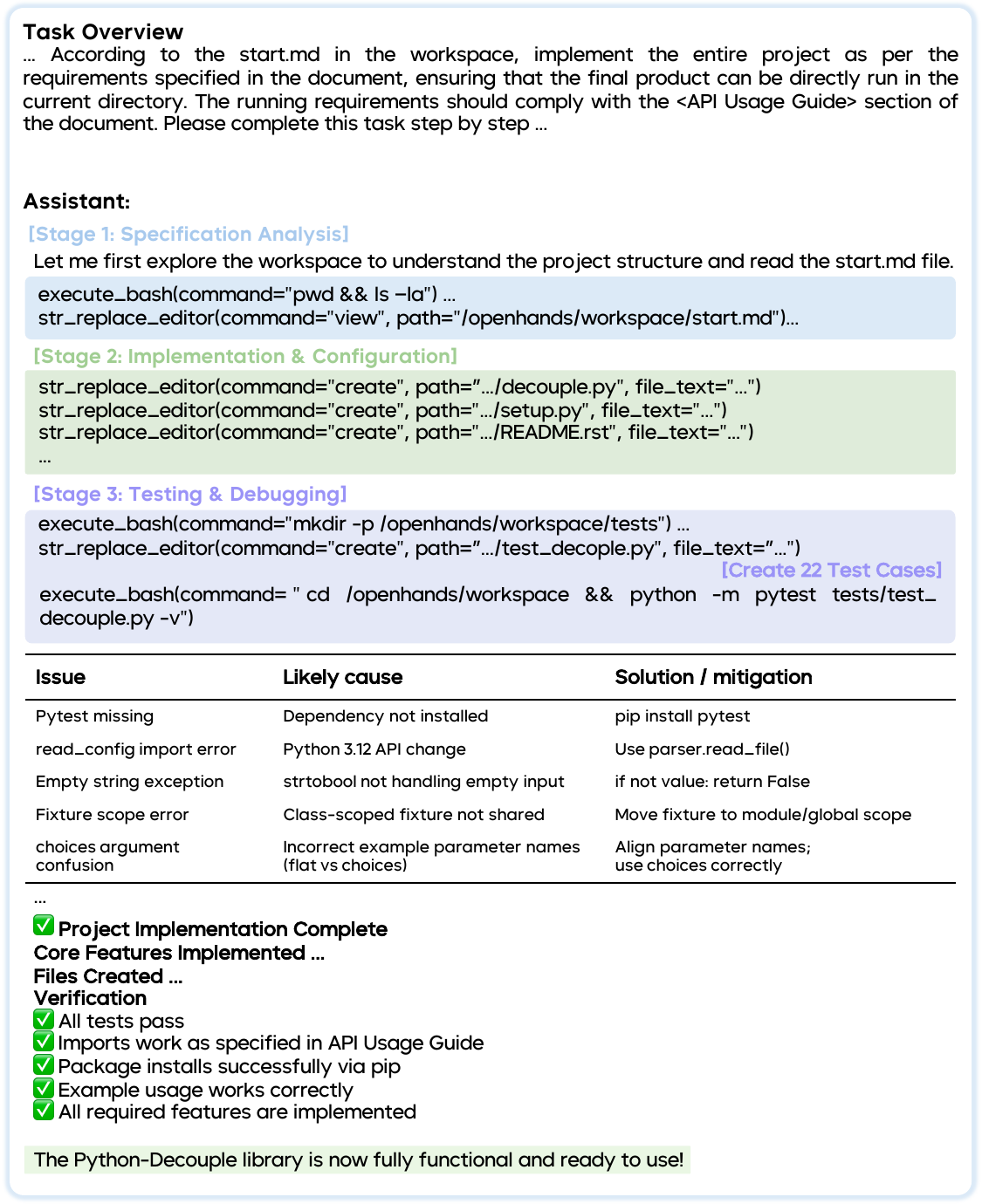}
  
    \caption{Case study of Seed2.0 performing the Project Repository Construction Task on NL2Repo~\citep{ding2025nl2repo}.}
  \label{fig:nl2repo}
  \vspace{-8pt}
\end{figure}

The attack's effectiveness derives from combining algebraic inversion with time–memory trade-offs. Seed2.0 derives inverse transformations for the cipher's internal operations, including the reversal of rotation and modular-addition steps in relevant components, and implements an explicit inverse routine for the \(F\)-function (\texttt{f\_inv\_function}). Seed2.0 then precomputes an inverse mapping table for the \texttt{expand()} operation, a \(20\)-bit to \(32\)-bit expansion described in the task specification, enabling efficient back-mapping from expanded subkey material to seed candidates and eliminating redundant computation during the search phase.

Crucially, Seed2.0 uses a meet-in-the-middle strategy to avoid exhaustive enumeration over the 80-bit key space. It splits the computation at an intermediate state and matches partial forward and backward results, trading memory and preprocessing for a large reduction in runtime and making the attack feasible under benchmark constraints.

After recovering a candidate key, Seed2.0 thoroughly validates it. It checks the key against all 32 plaintext–ciphertext pairs and cross-verified outputs with the provided \texttt{decrypt} program, including early spot checks. Only after consistent agreement finishes, it decrypts the 100 ciphertexts in \texttt{ciphertexts.txt}, writes outputs to \texttt{plaintexts.txt}, and reports successful completion.

This case study is notable because Seed2.0's approach differs from the benchmark reference implementation. This divergence supports the claim that it performed structural program analysis and task-specific cryptanalytic reasoning rather than matching a known solution. The layered verification further indicates a rigorous engineering workflow in a complex setting.

\subsubsection{Project Repository Construction}
\label{sec:project-responsitory-construction}

To evaluate the model's Project Repository Construction capabilities, we use the NL2Repo~\citep{ding2025nl2repo} task, which requires building a library-level Python repository from an empty workspace using only a requirements document. The deliverable is a Decouple-style configuration utility that separates configuration from code while remaining installable, testable, and usable as a standard Python package. Seed2.0 completes the task in 37 interaction rounds.
As shown in Figure~\ref{fig:nl2repo}, the workflow follows three standard phases: Specification Analysis, Implementation \& Configuration, and Testing \& Debugging. Seed2.0 first reads \texttt{start.md} and extracts requirements for configuration decoupling, multi-source loading. Implementation centers on \texttt{decouple.py}, which defines eight core classes for source precedence, parsing, and type conversion.

Quality assurance relies on a pytest suite (\texttt{tests/test\_decouple.py}) with 22 model-written test cases. Failures reveal issues that were fixed iteratively: installing a missing pytest dependency; addressing a Python 3.12-related configuration-reading import failure by switching to \texttt{parser.read\_file()}; adding a guard for boolean casting on empty strings; moving fixtures to global scope to avoid class-scoping conflicts; and correcting a parameter-name mismatch (flat versus choices) to match the intended API. This pattern indicates systematic debugging guided by test output.

After these fixes, all 22 tests pass. The model also runs example code, validates installation via ``\texttt{pip install -e .}'', and confirms cross-directory imports, showing both functional correctness and packaging readiness. Overall, Seed2.0 follows a structured, engineer-like workflow from requirements to implementation, testing, and debugging, producing a complete, convention-compliant repository with documentation.

\begin{figure}[t]
  \centering
  % 用 trim 去边
  \includegraphics[width=0.88\textwidth,
                   trim=0 0 0 0,clip]{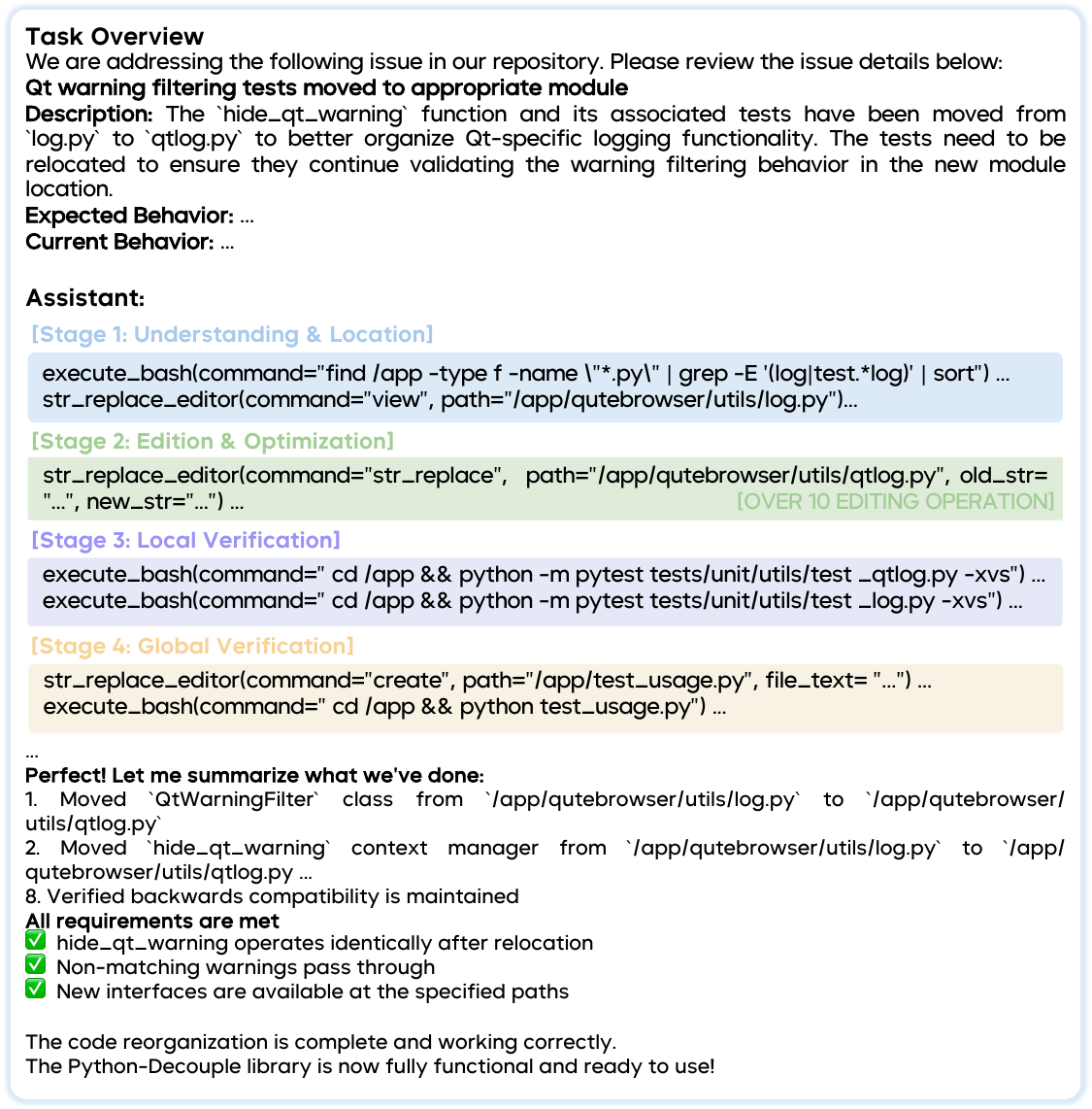}
    \caption{Case study of Seed2.0 performing the Iterative Code Debugging task on SweBenchPro~\citep{deng2025swe}.}
  \label{fig:swebenchpro}
  \vspace{+8pt}
\end{figure}

\subsubsection{Iterative Code Debugging}
\label{sec:iterative-code-debugging}

To systematically analyze the iterative code debugging capabilities of Seed2.0, we present a detailed case study from SWEBenchPro~\citep{deng2025swe}. This case presents a zero-regression refactoring that relocates the Qt warning suppression logic and its tests from \texttt{log.py} to \texttt{qtlog.py}, correcting module boundaries without altering external behavior.

As shown in Figure~\ref{fig:swebenchpro}, the refactoring preserves runtime semantics, the public API, and downstream compatibility across four key stages: Understanding \& Location, Edition \& Optimization, Local Verification, and Global Verification. The goal is not redesign but stronger verification through test relocation and extension. The core constraint is strict behavioral equivalence, identical behavior to the original, with the same effects on all non-Qt and non-matching logs.

More specifically, during Edition \& Optimization, correctness is validated at the handler boundary. Non-matching warnings continue to appear in the console and captured logs, while matching warnings are fully suppressed before reaching handlers. Suppression follows the original semantics: a message is hidden if it equals the filter string or begins with the configured pattern; otherwise, it prints normally.
The filtering mechanism, including \texttt{hide\_qt\_warning} and \texttt{QtWarningFilter}, is migrated to \texttt{qtlog.py}, with tests moved to \texttt{test\_qtlog.py}. For backward compatibility, \texttt{log.py} retains an import layer so existing call sites remain intact, consolidating \texttt{qtlog.py} as the single source of truth.

\begin{figure}[htbp]
  \centering
  % 用 trim 去边
  \includegraphics[width=0.88\textwidth,
                   trim=0 460 0 0,clip]{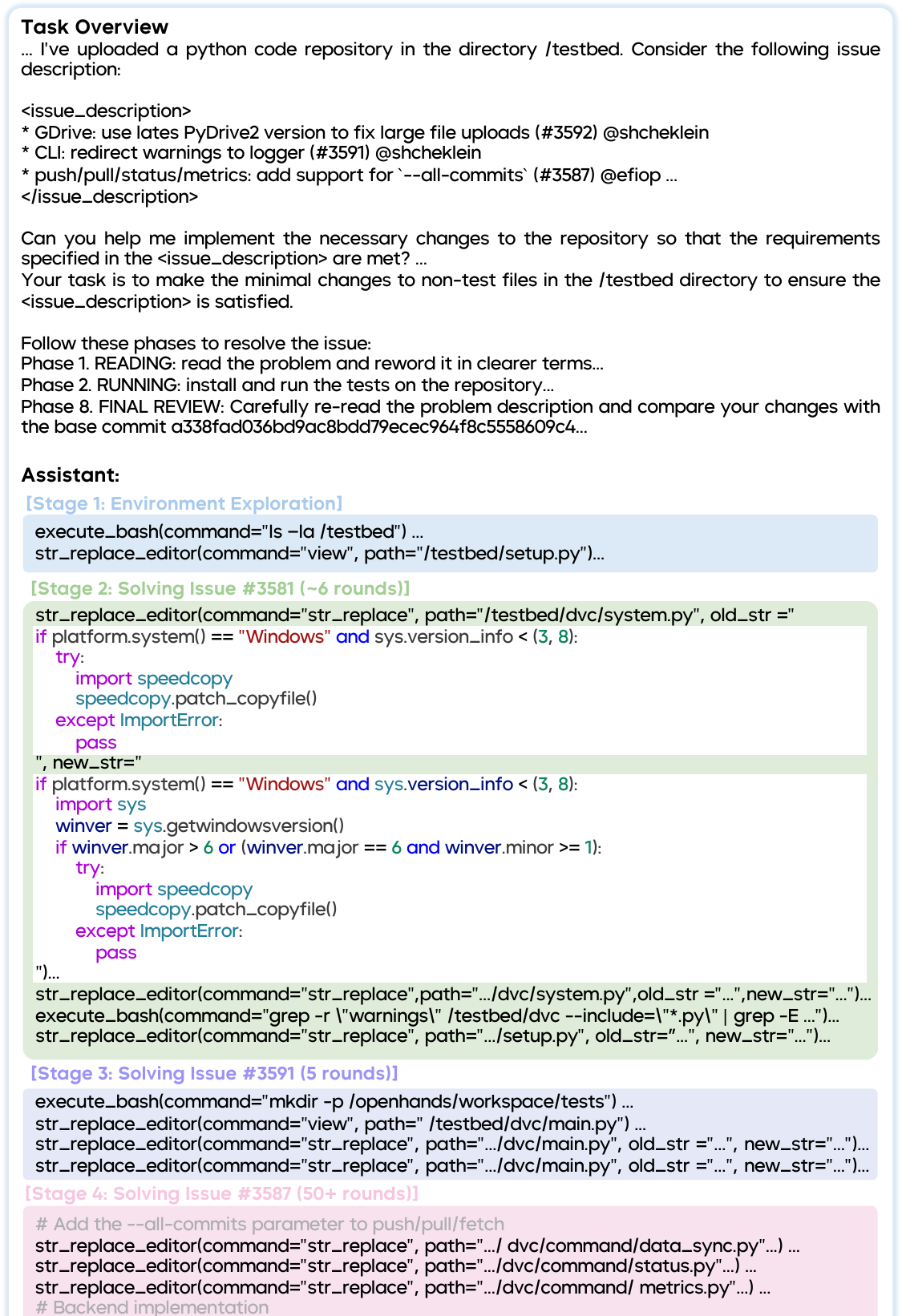}
  
\end{figure}
% \newpage
\begin{figure}[!t]
  \centering
  \includegraphics[width=0.88\textwidth,
                   trim=0 0 0 790,clip]{figure/SWE-EVO-Iterative-DVC.pdf}
    \caption{The case of Seed2.0 with Complex Iterative Debugging Task on SWE-EVO.}
  \label{fig:complex-iterative-debugging}
\end{figure}

% \begin{figure}[htbp]
%   \centering
%   % 用 trim 去边
%   \includegraphics[width=0.88\textwidth,
%                    trim=0 420 0 0,clip]{figure/SWE-EVO-Iterative-DVC.pdf}
  
% \end{figure}
% % \newpage
% \begin{figure}[t]
%   \centering
%   \includegraphics[width=0.88\textwidth,
%                    trim=0 0 0 830,clip]{figure/SWE-EVO-Iterative-DVC.pdf}
%     \caption{The case of Seed2.0 with Complex Iterative Debugging Task on SWE-EVO.}
%   \label{fig:complex-iterative-debugging}
% \end{figure}

% \begin{figure}[t]
%   \centering
%   \includegraphics[width=0.88\textwidth]{figure/competition.pdf}
%     \caption{The competitive programming results across five ICPC.}
%   \label{fig:icpc}
% \end{figure}

Validation proceeds through five phases: boundary scanning, migration with compatibility wiring, unit regression testing, integration verification, and final cleanup. Results show 56 passing unit tests, verified import compatibility, accurate handler outputs, and complete reference coverage. A logger-parameter issue found during testing is fixed and revalidated, ensuring precise behavioral equivalence. The refactoring delivers comprehensive test coverage and production-level maintainability.

\begin{figure}[t]
  \centering
  \includegraphics[width=0.88\textwidth]{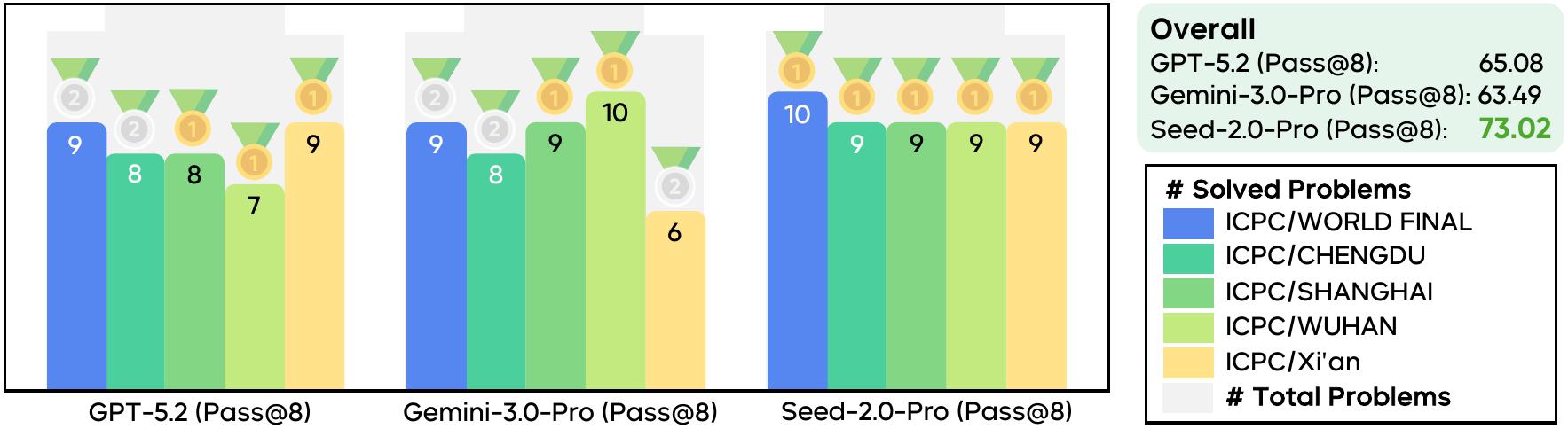}
    \caption{The competitive programming results across five ICPC.}
  \label{fig:icpc}
\end{figure}

\subsubsection{Complex Iterative Debugging}
\label{sec:complex-iterative-debugging}
In this Release-Note-only SWE-EVO upgrade case, the agent’s objective shifts from resolving a single, well-scoped issue (as in SWE-bench) to upgrading a library across versions using only release notes as guidance. With no explicit file-level directives, the model has to infer which components encoded the promised behavioral changes and treat the full unit test suite as the acceptance oracle, including both historically passing and historically failing tests. As shown in Figure~\ref{fig:complex-iterative-debugging}, across 79 turns, the workflow converges on an exploration→implementation→verification loop: it first maps the repository and dependency surface (via directory inspection and \texttt{setup.py}) and identifies the codebase as DVC, then iteratively implements cross-cutting upgrades spanning OS compatibility, logging semantics, CLI/API consistency, concurrency safety, and user-facing output correctness.

The most involved change is the end-to-end introduction of an \texttt{--all-commits} flag across \texttt{push}, \texttt{pull}, \texttt{status}, and \texttt{metrics}, because it requires more than CLI plumbing: the agent has to thread a single semantic intent from command entry points through repository operations and into revision traversal, ultimately extending the branch iteration mechanism (\texttt{brancher.py}) and ensuring repository initialization invokes the updated traversal path (\texttt{init.py}).

In parallel, it handles platform constraints by gating \texttt{speedcopy} imports using \texttt{sys.getwindowsversion()} (a robust choice over fragile string parsing) and refines progress reporting by switching tqdm-related locking to \texttt{threading.RLock()} to mitigate re-entrancy and multi-thread contention. It also improves UX fidelity by suppressing the ``\texttt{No changes}'' message in \texttt{metrics diff} when suppression is requested, demonstrating attention to behavioral edge cases that tests commonly encode. Notably, the \texttt{PyDrive2} large-file upload item is resolved as a dependency conformance check, \texttt{pydrive2>=1.4.8} is already declared, so the agent avoids unnecessary code churn, aligning with a minimal-change upgrade strategy. Verification proceeds through repeated targeted test runs and final holistic validation, including checking CLI help output to ensure argument exposure matched the intended interface design.

\subsubsection{Competition-Level Programming}
\label{sec:competition-programming}

% To more thoroughly assess the models’ competition-level programming capability, we design an ICPC-style evaluation protocol that mimics realistic contest participation. Rather than evaluating a single solution, the protocol emphasizes end-to-end performance under typical contest constraints, including algorithmic reasoning, careful handling of corner cases, and implementation robustness under limited trial-and-error.

%Specifically, we simulate participation in five International Collegiate Programming Contest (ICPC) events, comprising the 2025 World Finals (WF) and the four most recent regional contests held in late 2025.
%For each problem instance, the model is asked to produce 8 candidate programs via randomized generation (analogous to multiple independent attempts). These candidates are submitted sequentially to an online judge; once an Accepted verdict is obtained, we stop further submissions for that problem. We then compute the acceptance-based score and aggregate it into Best@8, which measures the probability/score of solving a problem within at most 8 submissions.
%As shown in Figure \ref{fig:icpc}, Seed-2.0-Pro achieves the best overall performance in these simulated events, reaching a gold-medal level. 
To evaluate capabilities on competition-level programming, we assess model performance across five recent ICPC Official contests in late 2025. These five contests include the ICPC 2025 World Finals Baku (Sept 4, 2025), the ICPC 2025 Xi'an Regional Contest (Oct 19, 2025), the ICPC 2025 Chengdu Regional Contest (Oct 26, 2025), the ICPC 2025 Wuhan Regional Contest (Nov 2, 2025), and the ICPC 2025 Shanghai Regional Contest (Nov 23, 2025). We utilize test cases, time limits, and memory limits consistent with the official competitions. 
We compare the Pass@8 score for each model, which represents the percentage of problems solved when allowing each model up to eight independent generations per problem. The input prompt for the models contains only the problem statement and a simple instruction requiring the solution in C++, without the use of multi-turn interaction or external tool calls.

\begin{figure}[htbp]
  \centering
  % 用 trim 去边
  \includegraphics[width=0.88\textwidth,
                   trim=0 420 0 0,clip]{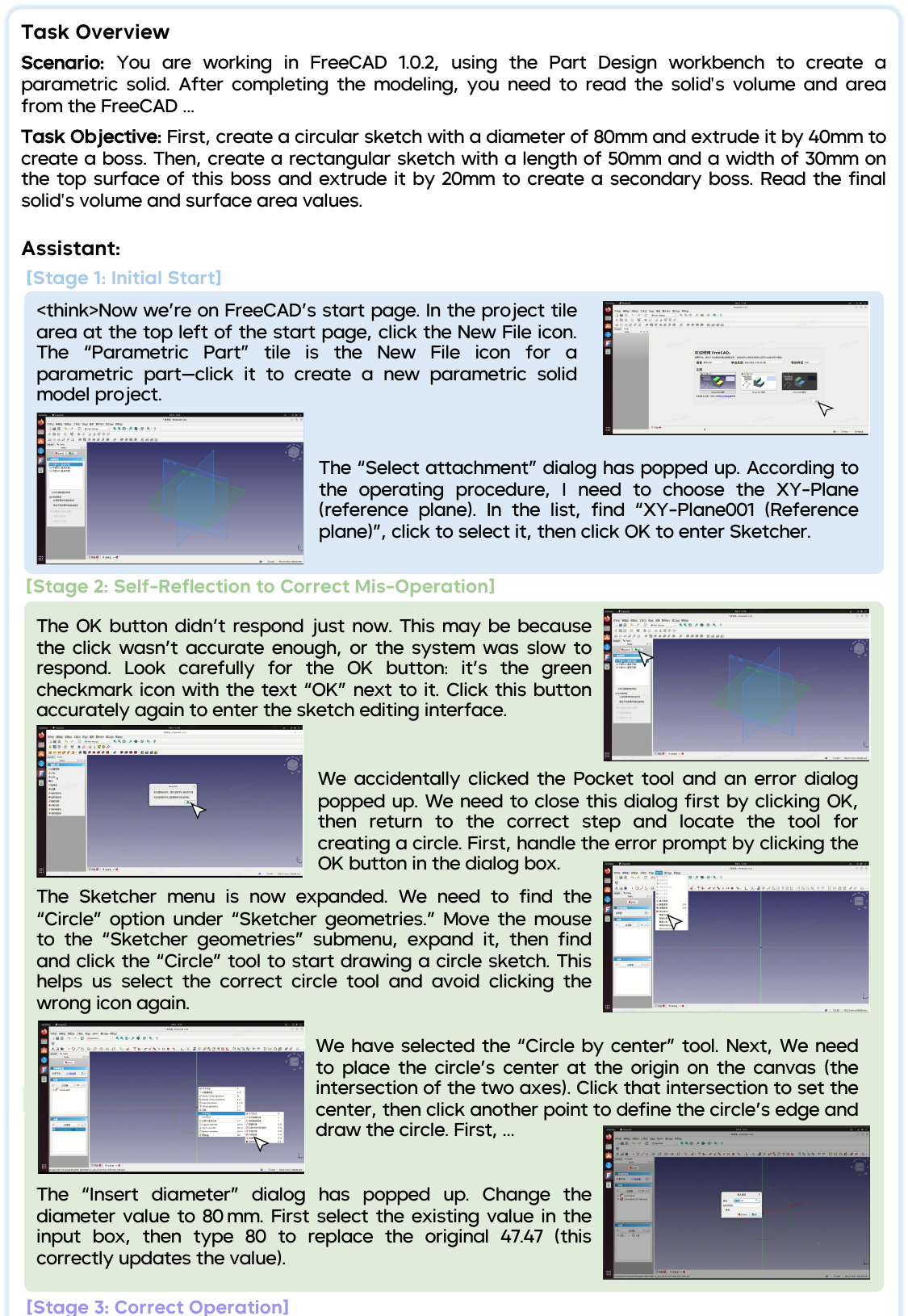}
  
\end{figure}
% \newpage
\begin{figure}[t]
  \centering
  \includegraphics[width=0.88\textwidth,
                   trim=0 0 0 785,clip]{figure/FreeCAD-Case.pdf}
    \caption{The case of Seed2.0 with FreeCAD Operation Task.}
  \label{fig:freecad}
\end{figure}

The results are shown in Figure \ref{fig:icpc}. Seed2.0 Pro achieves a Pass@8 score of 73.02\%, significantly outperforming GPT-5.2 and Gemini-3-Pro. Compared against human team performance, Seed2.0 Pro achieves Gold Medals in all five contests.

%In contrast, GPT-5.2 and Gemini-3.0-Pro remain at a silver-medal level on ICPC/CHENGDU and ICPC/WorldFinal, indicating persistent failure modes on harder tasks where small implementation errors or overlooked edge cases can be decisive. Overall, Seed-2.0-Pro improves the Best@8 score by nearly 8\% relative to GPT-5.2 and Gemini-3.0-Pro, suggesting that the gain is systematic across contests rather than being driven by a small number of outlier problems.

\subsection{Real-world Application Operation}
In productivity software, GUI interactions are state-dependent: toolchains vary with workspace or mode changes; parameters interlink across editing rounds; and interaction details (e.g., dialog prompts, focus management, double-clicking, drag-to-align) can trigger cascading failures. To examine whether Seed2.0 exhibits semantic understanding and long-horizon execution for complex interfaces, we evaluate it in two settings: (i) FreeCAD, a CAD environment emphasizing modeling context and constraint logic, and (ii) CapCut, a video-editing environment emphasizing timeline alignment and multi-track editing. These case studies illustrate how the model executes multi-step procedures while tracking interface state and self-corrects when UI errors arise.

\begin{figure}[htbp]
  \centering
  % 用 trim 去边
  \includegraphics[width=0.88\textwidth,
                   trim=0 600 0 0,clip]{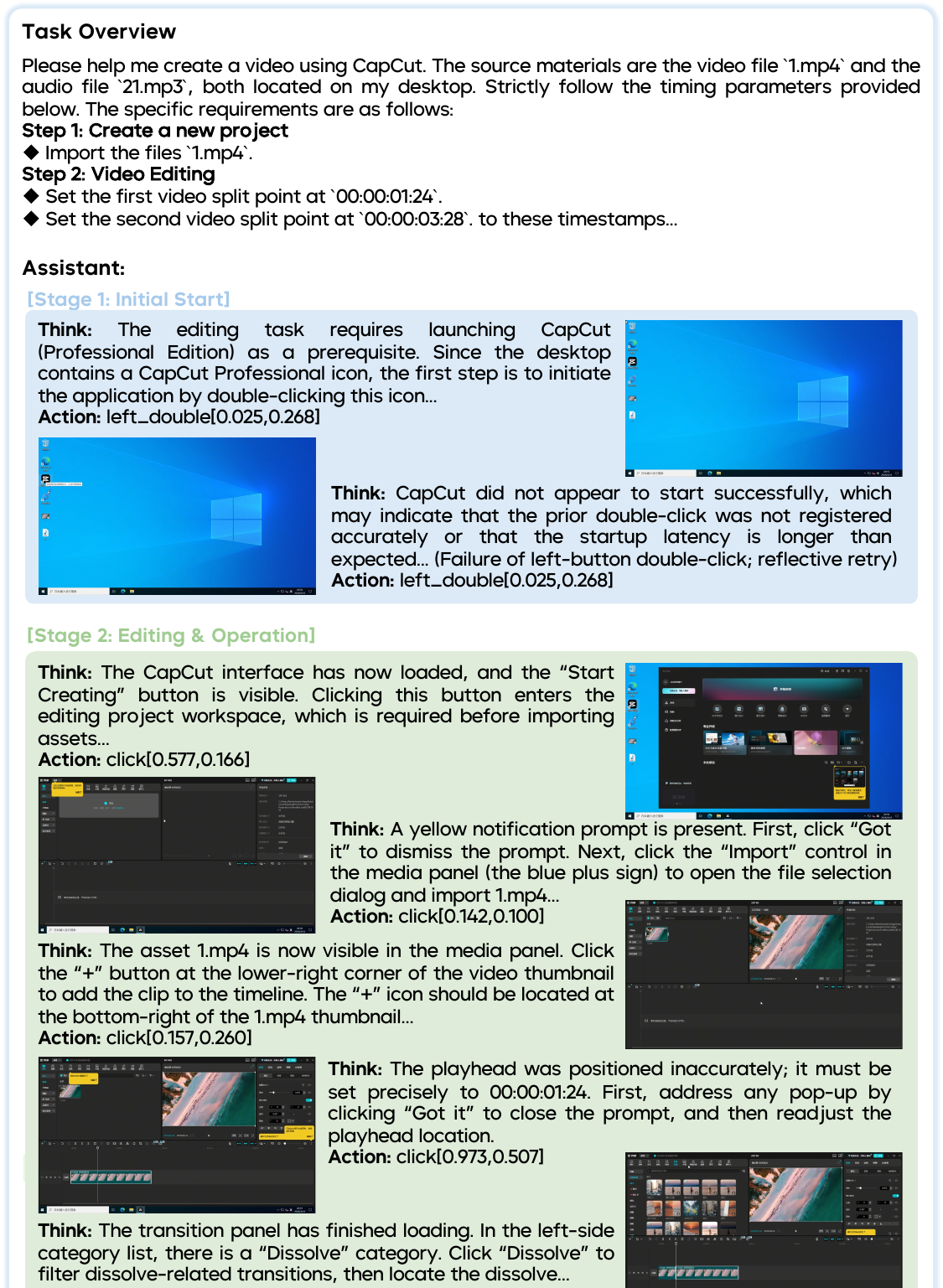}
  
\end{figure}
% \newpage
\begin{figure}[!t]
  \centering
  \includegraphics[width=0.88\textwidth,
                   trim=0 0 0 750,clip]{figure/CapCut-Case.pdf}
    \caption{The case of Seed2.0 with CapCut Operation Task.}
  \label{fig:capcut}
\end{figure}

\begin{figure}[!t]
  \centering
  \includegraphics[width=0.88\textwidth]{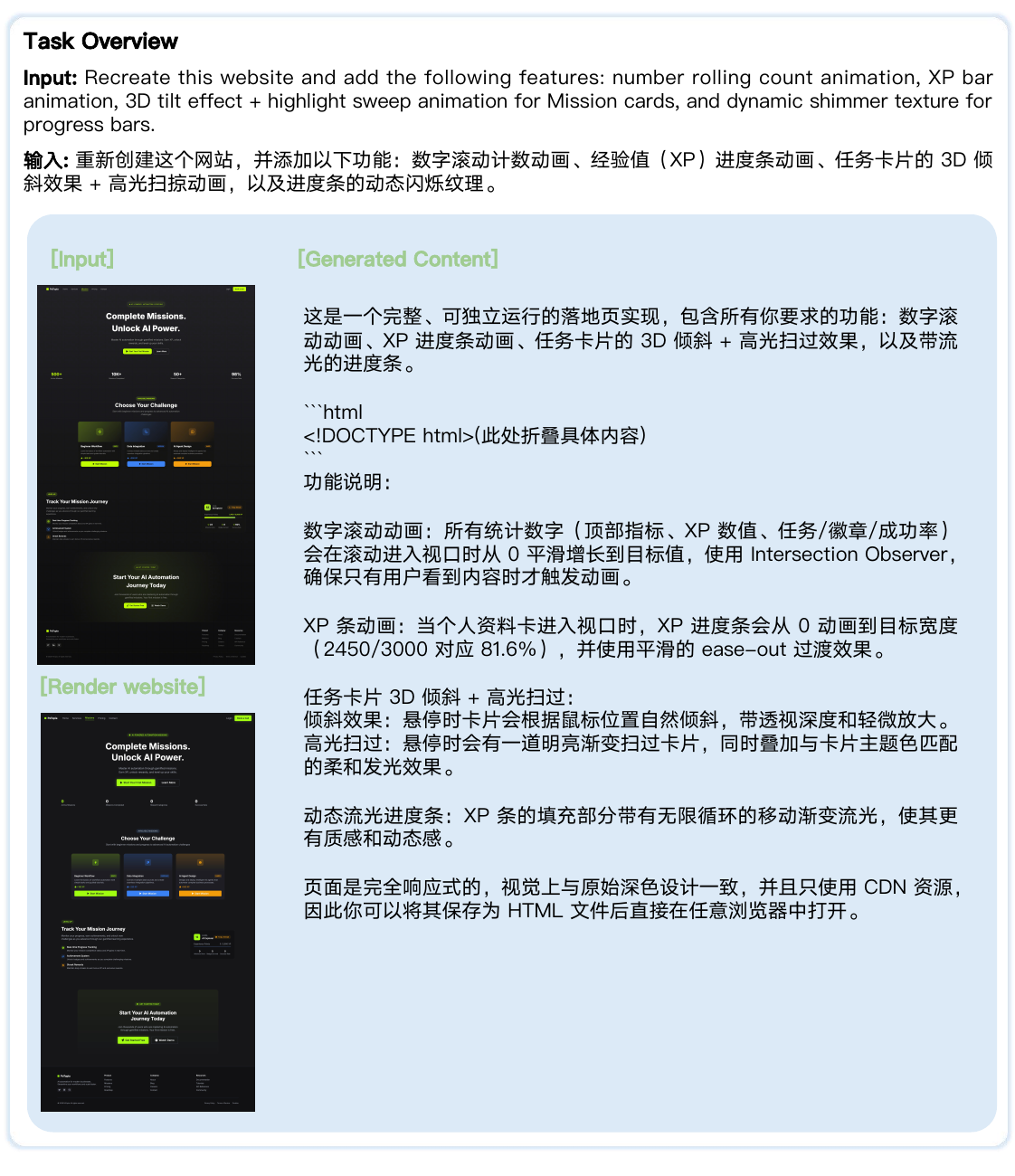}
    \caption{The case of Seed2.0 with Image to code website HTML generation task.}
  \label{fig:case1_web}
\end{figure}

\begin{figure}[!t]
  \centering
  \includegraphics[width=0.88\textwidth]{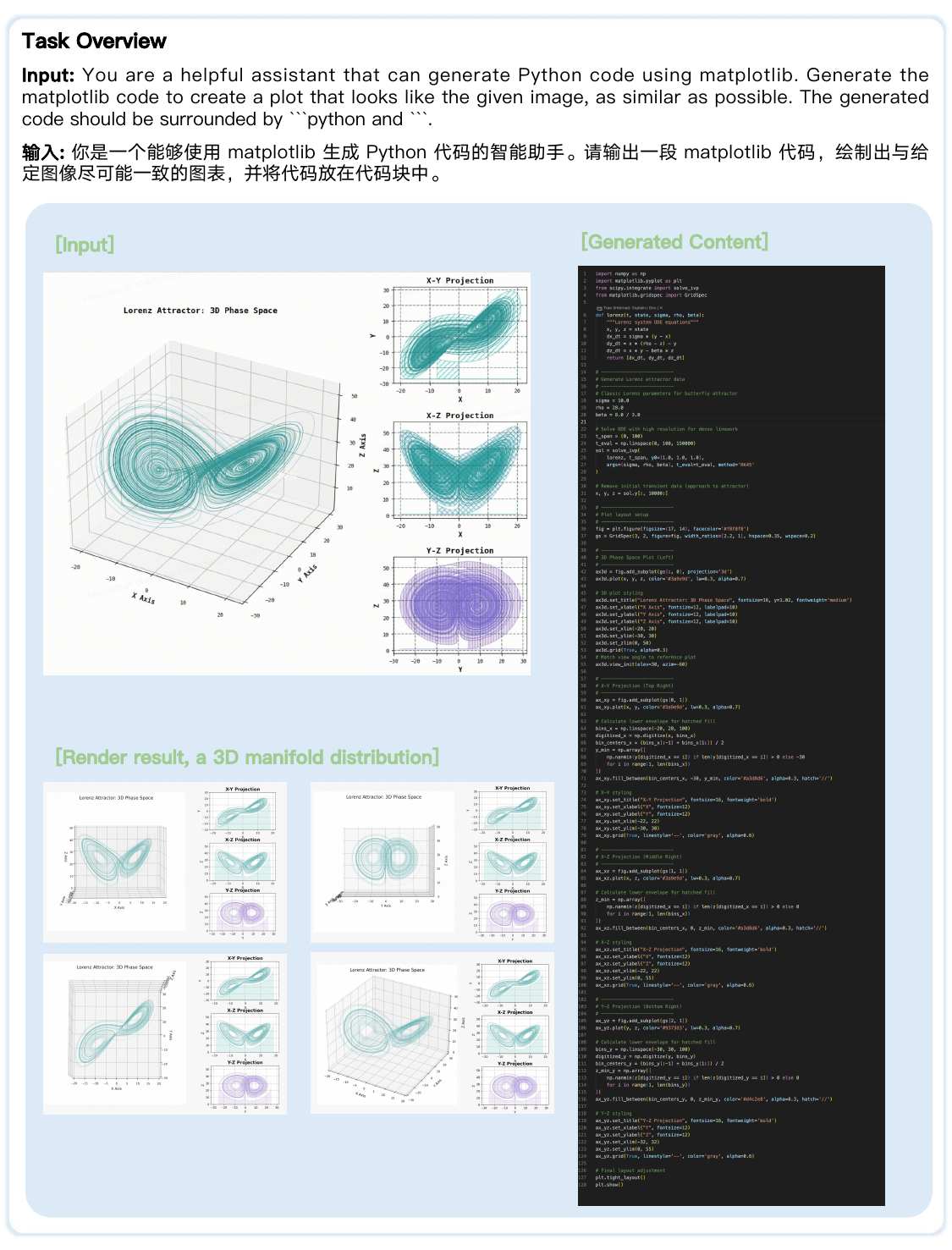}
    \caption{The case of Seed2.0 with Image to code 3D distribution generation task.}
  \label{fig:case2_3d}
\end{figure}

\subsubsection{FreeCAD Operation}
As shown in Figure~\ref{fig:freecad}, we evaluate whether Seed2.0 can perform semantic GUI understanding in a stateful CAD environment beyond template-based interactions. The model successfully tracks contextual information, including active workbenches, bodies, operational modes, and attachment planes. When encountering errors such as incorrect tool selection or unresponsive dialogs, the model demonstrates expert-like recovery by formulating micro-plans (e.g., "create circle, constrain diameter, exit sketch") and executing them.
The model's self-reflection mechanism functions as an online control loop under noisy UI feedback. Rather than simply documenting errors, the model diagnoses failure modes (misclicks, mode errors, or UI lag), applies targeted corrective actions (dismissing dialogs, reselecting tools, or reconfirming constraints), and resumes the parametric workflow. This behavior suggests a generalizable framework for robust GUI agents: maintain system invariants, restore operational context, and proceed with the intended task.

Based on this, correctness is anchored by redundant verification rather than “looks right.” The model validates the final solid via both the Properties panel (Data → Shape → Volume/Area) and an independent Python readout (obj.Shape.Volume, obj.Shape.Area), turning a GUI procedure into a checkable outcome and reducing common silent failures like wrong selection or wrong document state.

\subsubsection{CapCut Operation}

Further, we demonstrate Seed2.0's capabilities on operational CapCut tasks. As shown in Figure~\ref{fig:capcut}, we present a case where Seed2.0 functions as an execution-oriented editor, tasked with producing a frame-accurate CapCut (Jianying Pro) project. The system maintains coherent goal states across multiple operations, video splitting, transition placement, audio splicing, and global effects application, while employing reflective self-correction to recover from UI-level errors. These errors include missed double-clicks, incorrect effect selection, and misaligned audio placement.

The trace reveals Seed2.0's core strength: adaptive control under partial failure. Upon detecting actions that fail to yield the intended UI state, the system re-identifies the correct control sequence. For instance, it re-targets the "Audio" panel, closes obstructive pop-ups, or switches from "click add" to "drag-to-join" for dissolve transitions. Crucially, the system re-anchors the workflow to canonical timeline invariants, project starts at 00:00:00:00 and splice boundary at 00:00:02:29, before proceeding. This re-anchoring mechanism prevents error propagation in multi-stage editing workflows.

\begin{figure}[htbp]
  \centering
  \includegraphics[width=0.88\textwidth,
                   trim=0 395 0 0,clip]{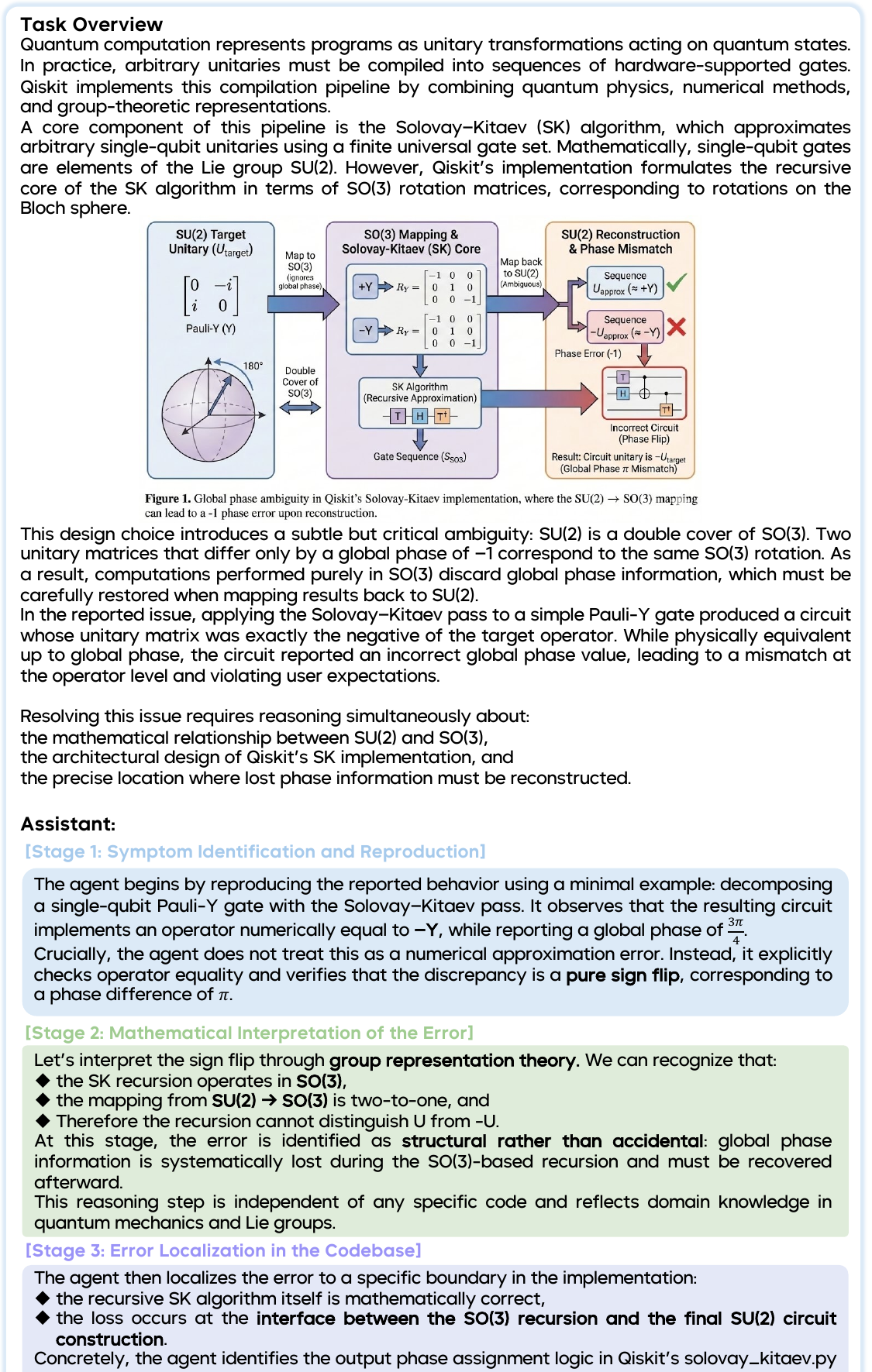}
  
\end{figure}
\begin{figure}[t]
  \centering
  \includegraphics[width=0.88\textwidth,
                   trim=0 0 0 850,clip]{figure/AInstein-Quantum-Computing-Case.pdf}
    \caption{The case of Seed2.0 with Quantum Computing Task on AInsteinBench~\citep{du2025deepresearch}.}
  \label{fig:quantum-computing}
\end{figure}

The specification's HH:MM:SS:FR timecodes are meaningful only under a consistent project frame rate. Therefore, Seed2.0's emphasis on precise alignment and playhead repositioning serves as a critical control mechanism that guards against silent drift at split and splice boundaries. Within this frame-accurate regime, the system applies the requested "Dissolve" transitions by locating them in CapCut's transitions library and placing them at each clip junction created by the three splits. Similarly, the system extends a snow-style atmospheric overlay to cover the full 00:00:00:00–00:00:08:26 program window by applying an effect clip on the timeline and adjusting its duration to match the target interval.

\subsubsection{Image-to-Code Generation}
Here are two examples of code generation in real-world application scenario. As shown in Figure~\ref{fig:case1_web}, the task is to recreate a website according to the given image. Some features are clearly required in the query prompt, including number rolling count animation, XP bar animation, 3D tilt effect + highlight sweep animation for Mission cards, and dynamic shimmer texture for progress bars. By saving the prediction code part as an HTML file and opening it directly in any browser, we could get a website almost the same as the input with all required features.
As shown in Figure~\ref{fig:case2_3d}, the task is to recreate a 3D plot as a matplotlib expert, in which x-y projection, x-z projection, and y-z projection relationships are provided. By compiling the generated Python code, we could get a 3D phase space which follows the projection as input.

\subsection{Multidisciplinary Scientific Research}
To comprehensively examine Seed2.0's utility in cross-disciplinary scientific research, we evaluate its capacity to facilitate scientific discovery through two complementary task scenarios: code generation across scientific domains and cross-domain analytical reasoning.

\subsubsection{Multidisciplinary Scientific Research Coding}

In real-world development of multidisciplinary scientific software, problems typically involve coupling among mathematical representations, physical conventions, and large-scale codebases, leading to subtle yet high-cost errors. Seed2.0 demonstrates robust generalization and debugging capabilities on such cross-domain tasks by identifying structural causes of anomalous behavior and producing verifiable fixes through dependency tracing and execution context analysis. The following sections present representative case studies from quantum computing, numerical general relativity, and computational chemistry.

\begin{figure}[htbp]
  \centering
  \includegraphics[width=0.88\textwidth,
                   trim=0 478 0 0,clip]{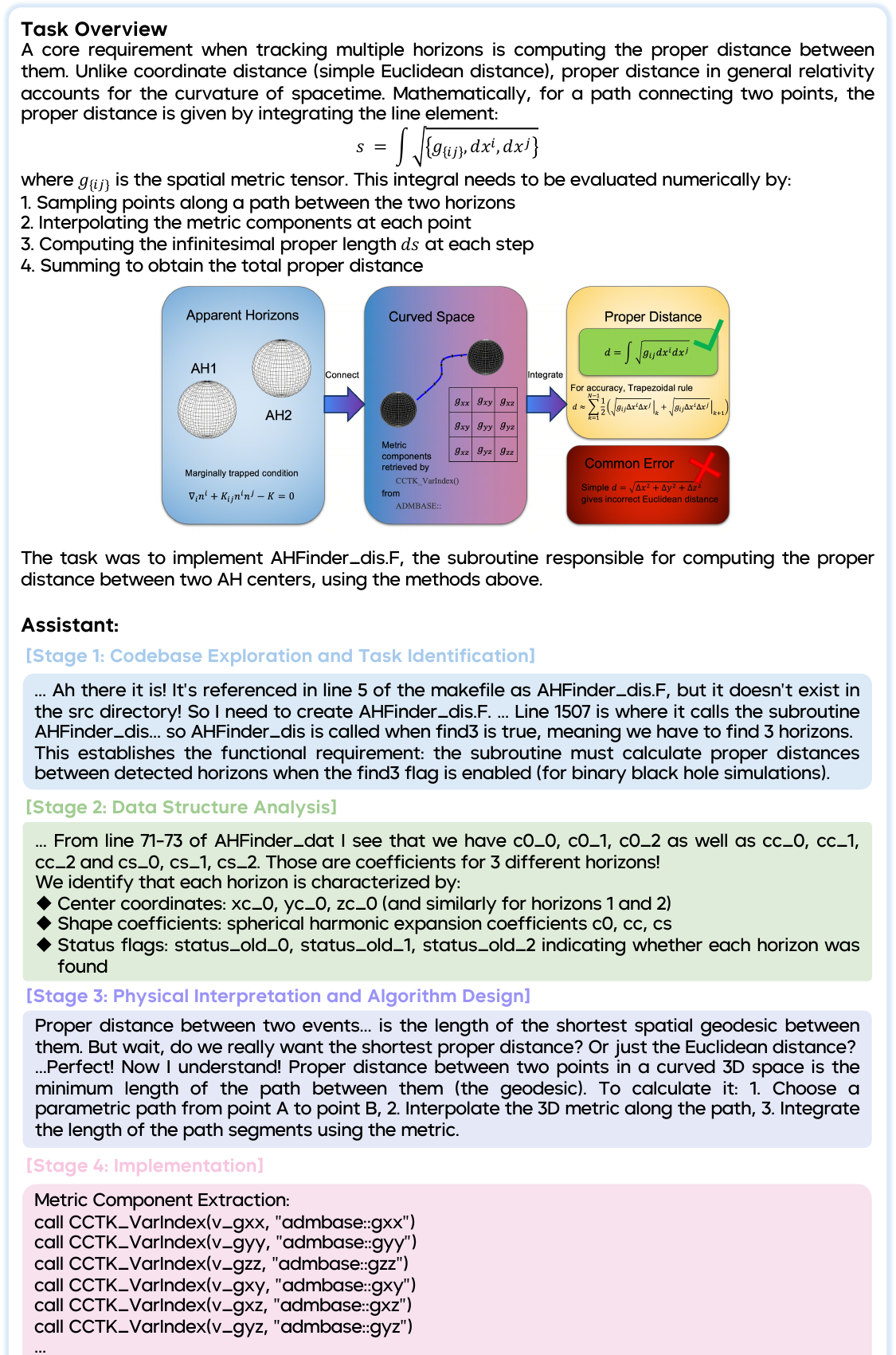}
  
\end{figure}
\begin{figure}[t]
  \centering
  \includegraphics[width=0.88\textwidth,
                   trim=0 0 0 820,clip]{figure/AInstein-General-Relativity-Case.pdf}
    \caption{The case of Seed2.0 with General Relativity Task on AInsteinBench~\citep{du2025deepresearch}.}
  \label{fig:general-relativity}
\end{figure}

\paragraph{\textbf{Quantum Computing.}}
On specialized and emerging tasks, Seed2.0 demonstrates remarkable generalization capabilities. In the quantum software engineering track of AInstein Bench~\citep{du2025deepresearch}, Seed2.0 successfully resolves a subtle bug in Qiskit's Solovay–Kitaev (SK) compiler stemming from a representation mismatch. The bug arises because Qiskit performs SK recursion in \( \mathrm{SO}(3) \) (Bloch-sphere rotations), whereas target single-qubit gates belong to \( \mathrm{SU}(2) \). Because \( \mathrm{SU}(2) \) double-covers \( \mathrm{SO}(3) \), the recursion cannot distinguish \(U\) from \(-U\), so global phase information can be dropped and must be restored when mapping the result back to an \( \mathrm{SU}(2) \) circuit. The reported symptom, decomposing Pauli-\(Y\) yields an operator exactly equal to \(-Y\) with an inconsistent global phase, reflects this structural phase ambiguity rather than approximation error.

Seed2.0's approach to this problem exemplifies systematic debugging methodology. The system first reproduces the failure using a minimal test case, verifies that the discrepancy manifested as a pure sign flip (phase difference of $\pi$), and localizes the root cause to the post-processing boundary in \texttt{solovay\_kitaev.py} where phase assignment occurs. Critically, Seed2.0 avoids modifying the mathematically correct recursion itself. After rejecting gate-specific constant-offset patches that would lack generalizability, it implements a principled solution: aligning operators by computing the phase \( \phi \) such that \( e^{i\phi} U_{\text{decomp}} = U_{\text{target}} \), then validating that this fix generalizes across all single-qubit gates without introducing regressions. This trajectory demonstrates agentic debugging through domain-theoretic diagnosis, precise code-level intervention at the appropriate interface, and rigorous generalization testing.

This case underscores Seed2.0's capability in quantum compiler debugging, a demanding scientific software engineering task that cannot be resolved through pattern matching or localized edits alone. Success requires understanding the mathematical abstraction (Lie group coverings), mapping this abstraction onto the software architecture, and identifying where theoretical guarantees must be enforced in the implementation. The demonstrated trajectory indicates that Seed2.0 can effectively navigate tasks requiring a deep integration of mathematical reasoning and software engineering principles.

\begin{figure}[htbp]
  \centering
  \includegraphics[width=0.88\textwidth,
                   trim=0 480 0 0,clip]{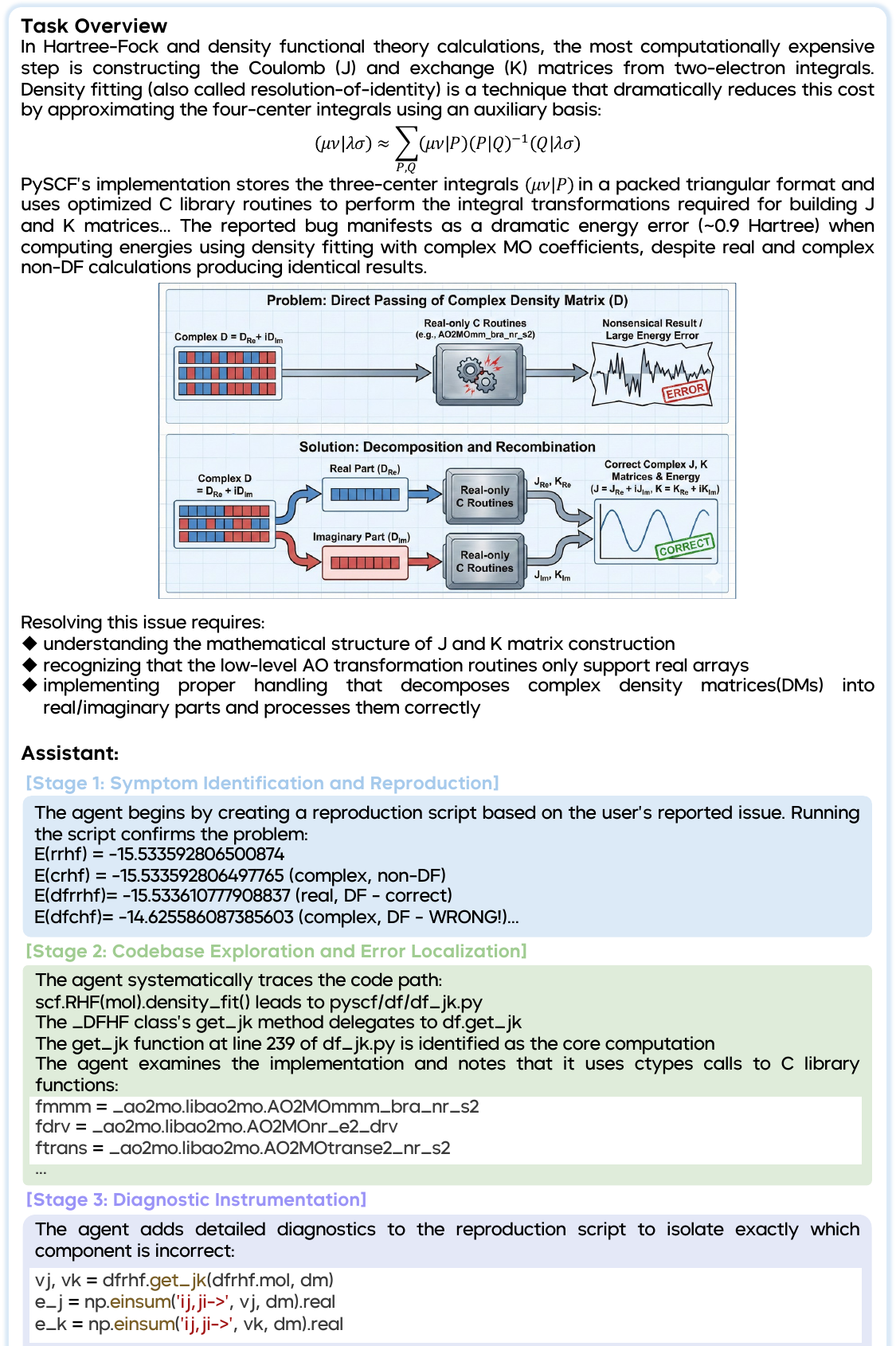}
  
\end{figure}
\begin{figure}[t]
  \centering
  \includegraphics[width=0.88\textwidth,
                   trim=0 0 0 810,clip]{figure/AInstein-Computational-Chemistry-Case.pdf}
    \caption{The case of Seed2.0 with Computational Chemistry Task on AInsteinBench~\citep{du2025deepresearch}.}
  \label{fig:computational-chemistry}
\end{figure}

\paragraph{\textbf{General Relativity.}}
Seed2.0 demonstrates robust performance on numerical relativity tasks from AInsteinBench~\citep{du2025deepresearch}. When tasked with implementing numerical-relativity functionality in the Einstein Toolkit's legacy Fortran codebase (Cactus framework), the agent autonomously diagnoses and resolves a critical integration failure: it identifies that \texttt{AHFinder\_dis.F}, referenced by the build system, is missing from \texttt{src}, then traces its call site in \texttt{AHFinder.F} to determine the routine's role in multi-horizon tracking during binary black-hole simulations. This systematic debugging approach, locating missing components and verifying execution context, demonstrates practical software engineering capabilities beyond isolated code generation.

The implementation exhibits both physics insight and numerical rigor. Seed2.0 correctly interprets "distance" as proper distance in curved spacetime rather than coordinate separation, implementing \( s = \int \sqrt{g_{ij}\,dx^i\,dx^j} \) with the full spatial 3-metric including all off-diagonal components ($g_{xx}, g_{yy}, g_{zz}, g_{xy}, g_{xz}, g_{yz}$). The solution properly accesses metric variables using Einstein Toolkit conventions, integrated with AHFinder's output patterns, and includes robustness checks for invalid handles and absent horizons. While the production implementation incorporates additional refinements, grid symmetry handling, horizon–line intersections via spherical harmonics, and improved quadrature, the agent's solution captures the essential physics with appropriate simplifying assumptions.

This performance illustrates Seed2.0's capacity to navigate the multidimensional challenge space of scientific computing: translating physical concepts (Riemannian geometry, geodesic integration) into numerical methods (interpolation, quadrature) within complex software architectures (framework APIs, mixed-language codebases with F77/F90 syntax and preprocessor macros). The agent's successful integration of domain knowledge with system-level engineering reflects the core competencies that AInsteinBench is designed to evaluate.

\begin{figure}[htbp]
  \centering
  % 用 trim 去边
  \includegraphics[width=0.88\textwidth,
                   trim=0 300 0 0,clip]{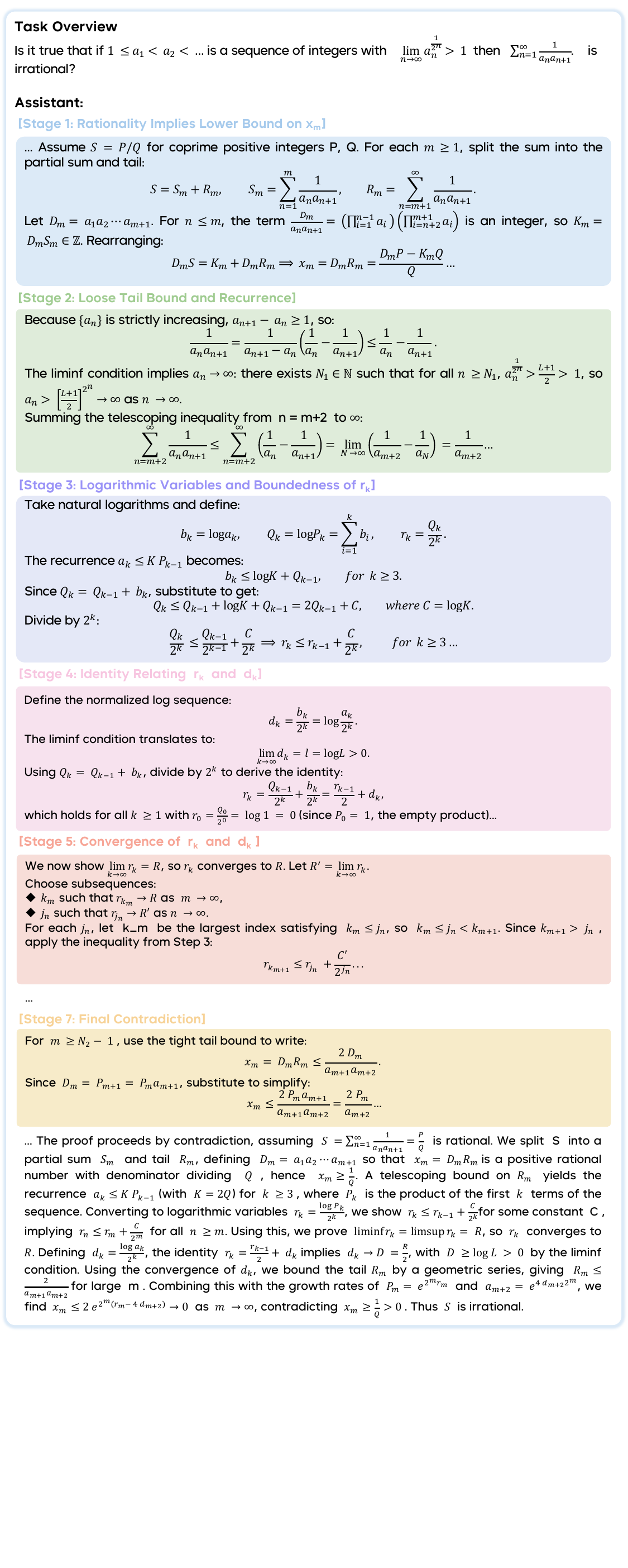}
  
\end{figure}
\begin{figure}[t]
  \centering
  \includegraphics[width=0.88\textwidth,
                   trim=0 0 0 850,clip]{figure/Erdos-Case.pdf}
    \caption{The case of Seed2.0 with Complex Mathematics Solution Task on Erd\H{o}s.}
  \label{fig:erdos}
\end{figure}

\paragraph{\textbf{Computational Chemistry.}}
Meanwhile, Seed2.0 can identify and resolve a subtle defect in PySCF's density-fitting (DF/RI) implementation that manifests exclusively with complex-valued density matrices. Density fitting accelerates \(J/K\) matrix construction by replacing four-center integrals with auxiliary-basis three-center tensors. The diagnostic symptom, a substantial energy deviation of approximately \(0.9\) Hartree in DF calculations with complex SCF inputs, occurs despite excellent agreement between real and complex non-DF energies (precision to \(10^{-11}\) Hartree). This pattern immediately localizes the failure to the DF \(J/K\) pipeline rather than general complex-number handling.
Seed2.0 reproduces the error and traces the execution path to \texttt{pyscf/df/df\_jk.py}, correctly identifying that the underlying computation routes through low-level C AO2MO kernels whose naming and packing conventions assume real-valued, symmetric arrays. By instrumenting the computation to isolate energy contributions, Seed2.0 determines that \(J\) remains accurate under the DF approximation while \(K\) exhibited severe corruption. This diagnostic pattern is consistent with complex buffers being misinterpreted as real memory, a failure mode that produces plausible but silently incorrect outputs.
The implemented fix is minimal yet mathematically rigorous. Seed2.0 intercepts complex density matrices, decomposes \(D\) into real and imaginary components, executes the existing optimized real DF pathway twice, and recombines results via linearity: \(K(D_{\mathrm{re}} + iD_{\mathrm{im}}) = K(D_{\mathrm{re}}) + iK(D_{\mathrm{im}})\). Post-patch validation confirms that DF complex energies match both DF real and non-DF references to numerical precision (\(\sim 10^{-11}\) Hartree). Seed2.0 further verifies robustness by executing the existing test suite and ensuring comprehensive coverage of complex-DF behavior.

This resolution exemplifies Seed2.0's capability to integrate domain expertise, understanding \(J/K\) matrix construction under density fitting, with systems-level debugging skills, including analysis of ctypes-to-C boundaries and packed storage conventions. The resulting patch maintains computational efficiency while preserving the stability of the real-valued fast path.

\begin{figure}[htbp]
  \centering
  % 用 trim 去边
  \includegraphics[width=0.88\textwidth,
                   trim=0 301 0 0,clip]{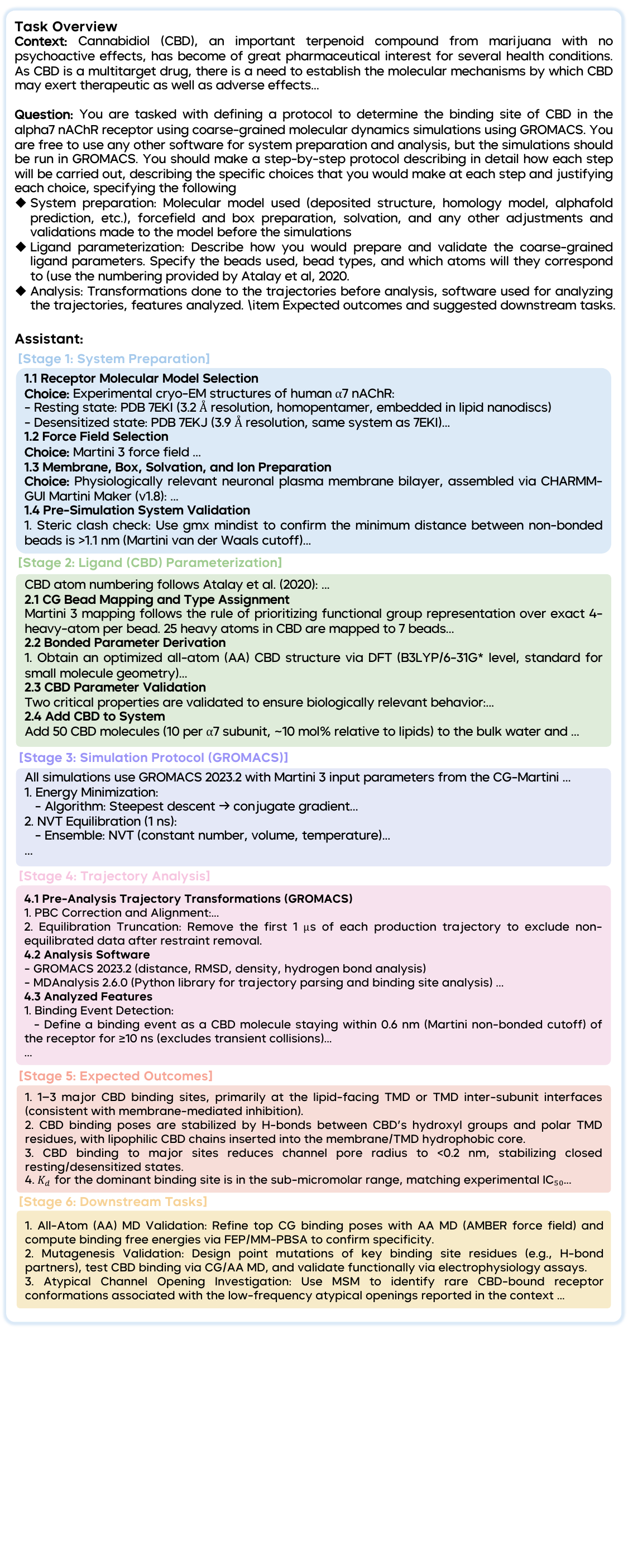}
  
\end{figure}
\begin{figure}[t]
  \centering
  \includegraphics[width=0.88\textwidth,
                   trim=0 0 0 834,clip]{figure/Cannabidiol-Case.pdf}
    \caption{Case study of Seed2.0 performing the Cannabidiol Scientific Analysis on FrontierScience.}
  \label{fig:cannabidiol}
\end{figure}

\subsection{Multidisciplinary Scientific Research Analysis}
In this section, we systematically evaluate Seed2.0's scientific research analysis capabilities across tasks that mirror real-world workflows. We design four thematic categories spanning theoretical derivations, computational modeling, and experimental design. Our assessment examines whether Seed2.0 produces outputs that are technically sound, internally consistent, and aligned with domain conventions, including explicit assumptions, appropriate methodological choices, and reproducible procedures. We further evaluate its justification of technical decisions and the transferability of outputs to downstream applications, thereby characterizing its performance across multidisciplinary research settings.

\paragraph{\textbf{Complex Mathematics Solution.}}
To demonstrate the model's capability in Seed2.0 to solve complex mathematical tasks, we present proofs of two Erdős Problems \cite{BloomErdosProblem1051}, generated through a iterative refinement pipeline with Seed2.0. Figure~\ref{fig:erdos} shows a summary of the output and demonstrates that Seed2.0 can address frontier-level challenges in mathematical research. Human experts have verified these proofs for correctness, and we have formalized the proof of Erdős 1051 using Seed-Prover 1.5~\cite{seedprover15_blog}. The model demonstrates a strong capacity for systematic reasoning and rigorous proof construction. See Appendix~\ref{sec:frontier_math_details} for evaluation details and full outputs.

\paragraph{\textbf{Cannabidiol (CBD)-Mediated Modulation Analysis.}}
To assess Seed2.0's performance on complex biomolecular simulation tasks, we evaluate its ability to design a coarse-grained molecular dynamics (CG-MD) workflow for studying cannabidiol (CBD)–mediated modulation of the $\alpha$7 nicotinic acetylcholine receptor ($\alpha$7 nAChR). The task requires generating a complete protocol for identifying CBD binding sites, including system preparation, ligand parameterization, trajectory analysis, and specification of expected results with downstream applications. Critically, the model needs to justify all methodological choices and parameter settings, thereby demonstrating both practical competence in MD workflow construction and understanding of underlying simulation principles.

\begin{figure}[htbp]
  \centering
  % 用 trim 去边
  \includegraphics[width=0.88\textwidth,
                   trim=0 183 0 0,clip]{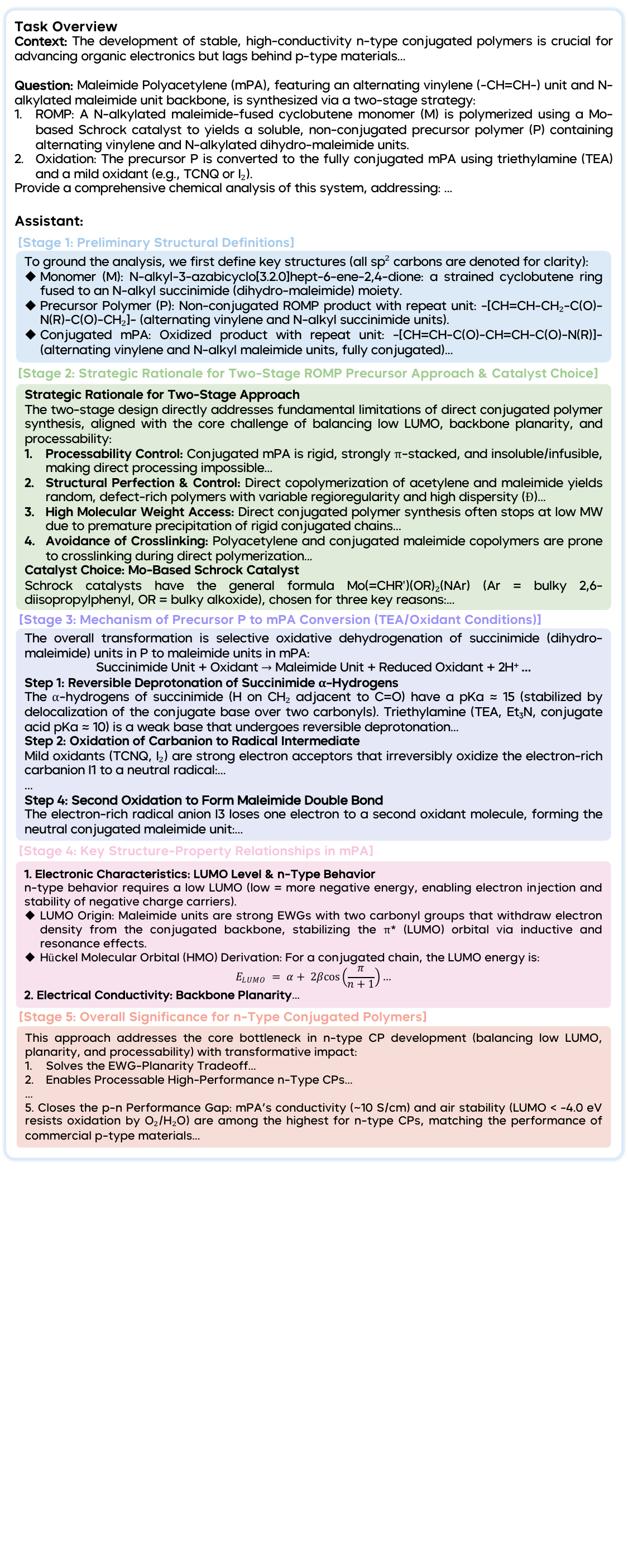}
  
\end{figure}
\begin{figure}[t]
  \centering
  \includegraphics[width=0.88\textwidth,
                   trim=0 0 0 815,clip]{figure/Maleimide-Polyacetylene-Case.pdf}
    \caption{Case study of Seed2.0 performing the Maleimide Polyacetylene Scientific Analysis on FrontierScience.}
  \label{fig:maleimide-polyacetylene}
\end{figure}

As shown in Figure~\ref{fig:cannabidiol}, Seed2.0 exhibits several notable strengths. The model correctly identifies an appropriate PDB structure and outlines a standard procedure for molecular structure preparation. The selected density functional theory (DFT) level for quantum mechanical calculations is suitable for the system. Notably, Seed2.0 proposes using the partition coefficient ($\log P$) as a validation metric to assess force field accuracy, a thoughtful choice for evaluating hydrophobic ligand behavior. The workflow incorporates proper gmx commands to address periodic boundary condition (PBC) artifacts, an essential preprocessing step. Time definitions are clearly stated with concrete decision criteria and corresponding GROMACS commands. The model also suggests appropriate validation strategies, including all-atom simulations and free energy perturbation (FEP) and molecular mechanics Poisson-Boltzmann surface area (MMPBSA) calculations for cross-validation.

However, the output contains several limitations. Most significantly, the model hallucinates specific details regarding PDB entries and cited literature, requiring careful verification before implementation. Such inaccuracies can mislead researchers and compromise workflow reliability.

Overall, Seed2.0 generates a logically structured and reasonably comprehensive workflow with detailed operational steps and validation approaches aligned with standard practices. The output provides a useful initial template for protocol development that can be iteratively refined. Nevertheless, the hallucination issues represent a critical weakness that must be systematically addressed through manual verification before practical application.

\paragraph{\textbf{Maleimide Polyacetylene Analysis.}}
To further evaluate Seed2.0's capabilities, we examine its performance on a more challenging task: analyzing a high-performance copolymer synthesis route based on a ROMP–precursor oxidation strategy. This two-step protocol provides access to electron-deficient olefin polymers that are difficult to obtain through conventional polymerization methods. The assessment evaluates the model's understanding of strategic route design, catalyst selection, an atypical coupling deprotonation–oxidation mechanism, and the relationships between polymer microstructure and macroscopic properties, thereby testing its grasp of complex organic reaction mechanisms, reaction selectivity, and fundamental structure–property correlations.

As shown in Figure~\ref{fig:maleimide-polyacetylene}, the model demonstrates strong mechanistic reasoning. In the ROMP step, it accurately identifies the ring-opening metathesis polymerization behavior of cyclobutene derivatives and correctly recognizes that highly strained, electron-deficient cyclobutene substrates require highly active Schrock-type catalysts. The model further notes that this strategy offers distinct advantages over traditional copolymerization approaches, including superior molecular-weight control, improved regularity of the alternating architecture, and effective suppression of side reactions. For the second-step TEA/oxidant coupling deprotonation–oxidation process, the model correctly identifies that triethylamine alone is too weak a base to drive a simple acid–base equilibrium and that the subsequent oxidation shifts the equilibrium forward. It appropriately proposes a pathway involving single-electron abstraction by the oxidant to generate radical intermediates, demonstrating sound reasoning for this less common organic oxidation mechanism.

\begin{figure}[htbp]
  \centering
  % 用 trim 去边
  \includegraphics[width=0.88\textwidth,
                   trim=0 210 0 0,clip]{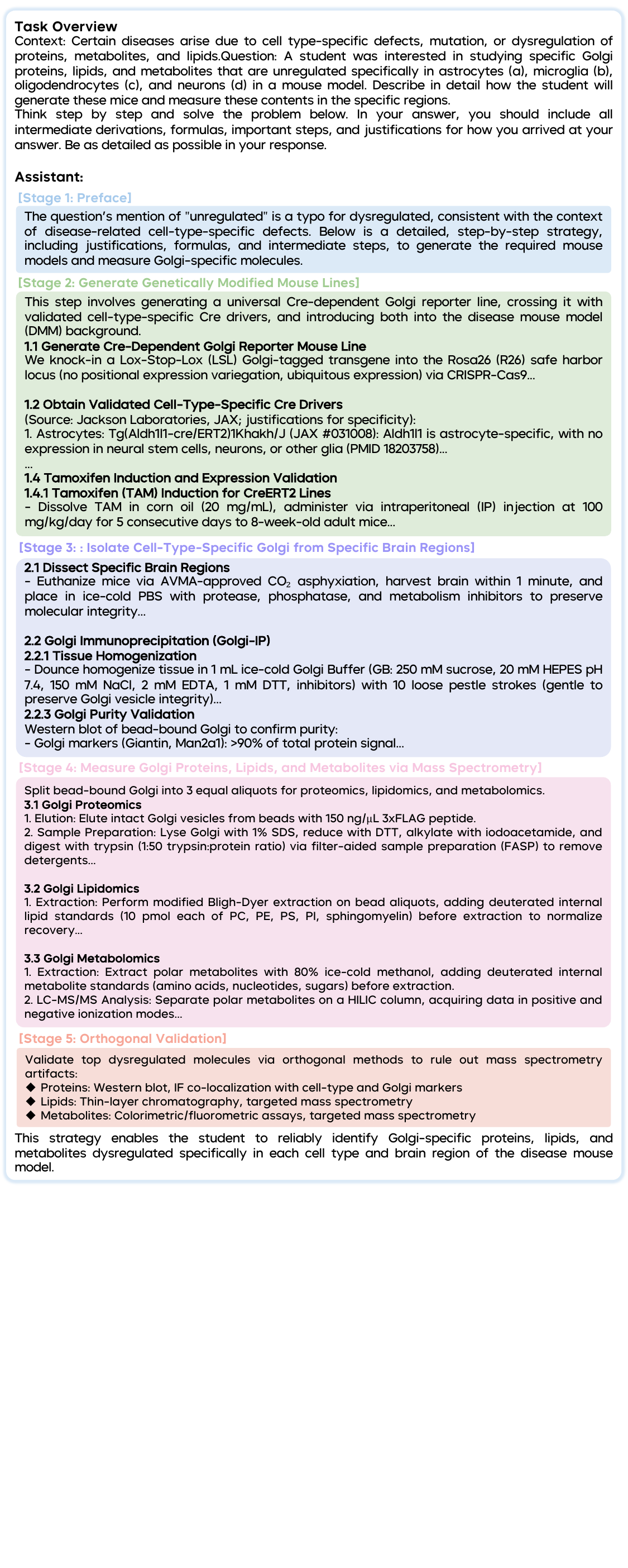}
  
\end{figure}
\begin{figure}[t]
  \centering
  \includegraphics[width=0.88\textwidth,
                   trim=0 0 0 810,clip]{figure/Golgi-Proteins-Case.pdf}
  
    \caption{Case study of Seed2.0 performing the Golgi Proteins Scientific Analysis on FrontierScience.}
  \label{fig:golgi-proteins}
\end{figure}

The model also provides a thorough structure–property analysis. It correctly explains that the precursor polymer $P$ contains $sp^3$-hybridized carbon centers that increase backbone flexibility, resulting in good solubility and processability.
After oxidation, formation of the mPA backbone creates a highly conjugated, planar structure that maximizes $\pi-\pi$ stacking, enabling efficient intrachain electron delocalization and interchain charge transport.
The model recognizes that the maleimide moiety, as a strong electron-withdrawing group, effectively lowers the LUMO energy level to support an air-stable n-type semiconductor architecture.

However, the model exhibits one notable limitation. Although its macroscopic mechanistic narrative is logically consistent, it fails to identify the more specific charge-transfer (CT) complex mechanism operative in this reaction. Specifically, it does not recognize that TEA and the oxidant may first form a CT complex that serves as the reactive initiating species. Instead, the model relies on a more generic acid–base equilibrium coupled with electron-transfer explanation. Consequently, its account of how the weakly basic TEA efficiently triggers the reaction, particularly the key kinetic details, lacks mechanistic depth.

\paragraph{\textbf{Golgi Proteins Analysis.}}
We next analyze a case involving an integrative experimental design that uses mice as a model organism to examine genetic engineering and animal model construction, cell biology and subcellular fractionation, and multi-omics analytical workflows. This study demands substantial domain knowledge and involves several nontrivial technical details. At the level of genetic engineering and model generation, the central question is how to leverage the Cre-LoxP system to achieve cell-type-specific and spatiotemporally controlled gene expression.

As shown in Figure~\ref{fig:golgi-proteins}, the key challenge from a cell biology perspective is how to rapidly isolate highly purified Golgi apparatus from complex brain tissue while preserving molecular integrity. Conventional fluorescence-activated cell sorting (FACS) is unsuitable because it is time-consuming and can induce cellular stress, both of which compromise the intended downstream analyses. The multi-omics component spans sample preparation, mass spectrometry measurements, and the analytical logic across proteomics, lipidomics, and metabolomics.

The proposed model exhibits several notable strengths. The experimental workflow is specified in fine-grained steps: beyond proposing a transgenic mouse strategy, it provides actionable CRISPR/Cas9 targeting details, including homology-arm design at the Rosa26 (R26) locus, guide RNA (gRNA) design considerations, and genotyping primer planning.
It articulates a stringent quality-control framework at critical steps of Golgi immunoprecipitation (Golgi-IP), explicitly defining negative-control strategies (e.g., assaying endoplasmic reticulum, mitochondrial, and nuclear markers to exclude contamination) and offering quantitative purity assessment using a Pearson correlation coefficient (PCC)-based metric.
For the multi-omics analysis, statistical criteria for differential testing are appropriately defined, specifically $FDR<0.05$ and $|LFC| \ge 1.2$.

{\renewcommand{\arraystretch}{1.12}
\setlength{\tabcolsep}{6pt}
\small

\begin{table}[htbp]
    \centering
    \small
    \caption{\textbf{Scalable automated diagnostics for iterative model development.}
The pipeline converts large-scale benchmark evaluations into structured behavioral reports across reasoning, planning, and tool-use domains, revealing capability trade-offs and systematic failures with measured throughput and efficiency.}
\label{tab:posteval}
    \resizebox{0.92\textwidth}{!}{\begin{tabular}{p{2.2cm}p{9.0cm}p{3.3cm}}
    \toprule
    \textbf{Benchmark} & \textbf{Representative Findings} & \textbf{Efficiency Profile} \\
    \midrule
        \textbf{XBench} &
\textbf{Focus:} SearchAgent.\par
\textbf{Finding:} Seed2.0 Pro delivers the strongest overall performance on this benchmark while maintaining efficient search behavior. It consistently completes complex multi-step tasks through multi-source retrieval and iterative verification, but shows weakness in fine-grained boundary alignment and structured comparison scenarios. \par
\textbf{Evidence:} It ranks first with avg\_score=0.64, ahead of GPT-5.2 High (0.57), Claude-Sonnet-4.5-thinking (0.48), and Gemini-2.5-Pro (0.24), while using 337.62 completion tokens and 277.49 reasoning tokens on average. In long multi-hop cases such as case\_6 (48 calls), case\_15 (44 calls), and case\_52 (70+ calls), it converges through repeated verification; however, in boundary-sensitive tasks like case\_28, case\_33, case\_35, and case\_91, it receives score 0.0 due to condition misalignment or anchor drift. 
&
\parbox[t]{3.3cm}{\raggedright
\textbf{Cases per Model:} 100\\
\textbf{Models:} 9\\
\textbf{Time Cost:} 2435 s\\
\textbf{Tokens in:} ~0.9 M\\
\textbf{Tokens out:} ~0.1 M
}
\\
\midrule
% -------------------- WorldTravel --------------------
\textbf{WorldTravel} &
\textbf{Focus:} Real World Tasks.\par
\textbf{Finding:} Seed2.0 Pro demonstrates strong engineering stability, but its performance degrades under high coupling density due to weak second-order constraint reasoning and limited long-chain conflict repair. The primary gap with GPT-5.2 lies not in basic planning, but in maintaining constraint closure under cross-dependent temporal, structural, and pricing constraints.\par
\textbf{Evidence:} Seed2.0 Pro ranks 2nd with avg\_score=0.233, trailing GPT-5.2 High (0.327, +0.09) while outperforming most peers. It maintains error\_rate=0.00 and stable latency (31.39s), using 77\% fewer reasoning tokens (1286 vs 5597) and 84\% fewer completion tokens (1486 vs 9190) than GPT-5.2 High. Success cases concentrate in $\leq$15 structured constraints, whereas failures increase sharply when constraints exceed 15 with cross-dependencies. Major failure patterns include temporal window misalignment (case\_19, 27, 42), fine-grained ticket miscalculation (case\_23, 55), structural ordering errors (case\_17, 53), and missing long-chain conflict repair (case\_29, 46). Across 60 head-model comparisons, 33\% of failures stem from temporal misalignment, 30\% from structural violations, 25\% from pricing errors, and 28\% from conflict detection breakdown.
&
\parbox[t]{3.3cm}{\raggedright
\textbf{Cases per Model:} 150 \\
\textbf{Models:} 9 \\
\textbf{Time Cost:} 1571 s \\
\textbf{Tokens in:} ~0.9 M \\
\textbf{Tokens out:} ~0.04 M 
}
\\
\midrule

% -------------------- IMO-Bench --------------------
\textbf{IMO-Bench} &
\textbf{Focus:} Reasoning \par
\textbf{Finding:} Seed2.0 Pro is strong at pushing structured problems to a logical conclusion. In recursive, combinatorial, and formula-driven tasks, it can carry the reasoning chain step by step without breaking the internal logic. However, it struggles when a problem requires extreme-case construction or full boundary coverage. It often proves the main idea correctly but fails to completely seal upper–lower bounds or exhaust all branches.\par
\textbf{Evidence:} Seed2.0 Pro achieves avg\_score=0.779 (Rank 4/9), showing stable execution quality. Its bon=0.87 is close to Gemini-3-Flash (0.91) and Gemini-3-Pro (0.92), indicating competitive peak capability. However, its won=0.66 is significantly lower than GPT-5.2 High (0.81), resulting in a bon–won gap of 0.211, which implies 21\% sampling instability. Behaviorally, 65\% of its incorrect cases involve extreme-value or boundary-construction tasks, and 80\% involve full-quantifier or multi-branch exhaustion problems, aligning with the observed robustness gap.
&
\parbox[t]{3.3cm}{\raggedright
\textbf{Cases per Model:} 400 \\
\textbf{Models:} 9 \\
\textbf{Time Cost:} 1741 s \\
\textbf{Tokens in:} ~1.4 M \\
\textbf{Tokens out:} ~0.06 M
}
\\
\midrule
\midrule
\end{tabular}}
\end{table}

\begin{table}[!t]
    \resizebox{0.92\textwidth}{!}{
    \begin{tabular}{p{2.2cm}p{9.0cm}p{3.3cm}}
        \toprule
        \textbf{Benchmark} & \textbf{Representative Findings} & \textbf{Efficiency Profile} \\
        \midrule
    % -------------------- tau^2-Bench v2 --------------------
    \textbf{$\tau^2$-Bench} &
    \textbf{Focus:} Tool Use Agent.\par
    \textbf{Finding:} Seed2.0 Pro high-score cases dynamically skip redundant device checks and resolve issues within a controlled tool-call budget, while strictly adhering to single-step JSON outputs; low-score cases follow mechanical enumeration, over-consume dialogue turns, and break structural constraints.\par
    \textbf{Evidence:} Seed2.0 Pro high-score cases complete in 31.4 turns on average with 7.3 tool calls, whereas low-score cases expand to 55.0 turns (+75.2\%) and 9.6 calls (+31.5\%), often delaying billing checks by 30+ turns (e.g., case\_51, case\_54); Seed2.0 Pro scores 0.781 (3/9) using 721 completion tokens and 637 reasoning tokens, substantially fewer than GPT-5 High (2143 / 1946) despite a 0.17 score gap.
    &
    \parbox[t]{3.3cm}{\raggedright
    \textbf{Cases per Model:} 278\\
    \textbf{Models:} 9 \\
    \textbf{Time cost:} 3116 s \\
    \textbf{Tokens in:} ~1.8 M \\
    \textbf{Tokens out:} ~0.1 M}
    \\\bottomrule
    \end{tabular}
}
\end{table}}

One potential limitation concerns the choice of tag protein. The model selects Giantin as a Golgi marker; however, Giantin is a large matrix protein that may introduce more nonspecific binding during immunoprecipitation than transmembrane Golgi proteins such as Tmem115, and may pose a greater risk of perturbing Golgi architecture.

Overall, Seed2.0 performs at a level exceeding typical PhD student expectations, demonstrating the ability to integrate experimental details across subdisciplines. Rather than stopping at high-level strategy, it produces a feasible protocol with implementable steps. One practical consideration is that many PhD students would rely on commercial kits or established reagents rather than developing every component de novo; nevertheless, within a knowledge-driven framework, the model's response is well-constructed and scientifically sound.

\subsection{Automated Model-on-Model Behavioral Diagnostics}

As evaluation ecosystems grow in scale and diversity, manual analysis of benchmark results becomes increasingly costly and difficult to standardize.
We build an automated model-on-model diagnostic pipeline that uses LLMs to analyze the evaluation outcomes of peer models across heterogeneous benchmarks.
The system aggregates metric scores and behavioral statistics, including token usage, formatting compliance, turn counts, emoji frequency, best-of-$N$ statistics and other derived metrics, alongside instance-level outputs, reasoning traces, and execution trajectories.
Through a three-layer design that separates data processing, scenario-adaptive analytical workflows, and report synthesis, the framework scales to large case volumes and helps algorithm researchers efficiently iterate on models by exposing model weaknesses, strategy trade-offs, and concrete behavioral issues such as code-switching and repetitive CoT patterns.
Table~\ref{tab:posteval} shows that the pipeline scales across diverse benchmarks and produces structured findings with quantified throughput and resource statistics.
\section{Conclusion}
The Seed2.0 model series takes a key step forward in the intelligent evolution of solving complex real-world tasks. 
First, the Seed team identifies users' real needs, and selects or abstracts a series of benchmarks based on these real needs and complex real-world scenarios to build a reliable evaluation system. 
Based on the reliable and forward-looking evaluation system, Seed2.0 focuses on solving long-tail knowledge problems and complex instruction following problems, thereby enhancing the reliability of the model in complex and long-range real-world tasks. 
In addition, Seed2.0 has world-leading reasoning intelligence, visual understanding capabilities, and search capabilities, which meet the most real needs of a large number of users. 
The model card provides a large number of real use cases to prove that the Seed2.0 model begins to have the ability to handle initial complex real-world tasks, thus bringing more value to hundreds of millions of users.

\bibliography{main}
\bibliographystyle{plainnat}

\newpage
\section{Contributors}
The authors are listed in alphabetical order by their first names. Some names refer to the authors’ internal
aliases at the company.

\setlength{\parskip}{0pt} % 让段落之间没有额外空隙
\setlength{\itemsep}{0pt} % 如果用itemize
\setlength{\parsep}{0pt}  % 控制段落间距
\begin{multicols}{2}

Allan Jie\\
Anwen Hu\\
Aoyan Li\\
Astor Wu\\
Baihan Shu\\
Baisheng Li\\
Baizhou Huang\\
Banggu Wu\\
Baoquan Zhong\\
Bencheng Liao\\
Benedict Tay\\
Benjamin Koo\\
Bin Jia\\
Bin Liu\\
Binguo Bao\\
Bingrui Li\\
Bo Chen\\
Bo Li\\
Bo Liu\\
Bo Xu\\
Bole Ma\\
Bowen Li\\
Bowen Xiao\\
Buzz Cai\\
Caiping Lyu\\
Can Zhang\\
Chandler Nie\\
Chang Sun\\
Chang Tan\\
Chang Yan\\
Changxin Pu\\
Changya Chen\\
Changyue Liao\\
Chao He\\
Chao Li\\
Chao Wang\\
Chao Xin\\
Chao Yao\\
Chaomo Li\\
Chaoyi Deng\\
Chaoyi Huang\\
Chaoyi Zhang\\
Charles Chi\\
Chen Dun\\
Chen Fu\\
Chen Mao\\
Chen Zheng\\
Cheng Chen\\
Cheng Li\\
Cheng Lin\\
Cheng Luo\\
Cheng Ren\\
Chenggang Li\\
Chengqi Zhao\\
Chengquan Jiang\\
Chengye Li\\
Chengyi Wang\\
Chengyin Xu\\
Chenhui Huang\\
Chenjie Jia\\
Chenrui Wei\\
Chenxiao Liu\\
Chenxin Li\\
Chenyuan Wang\\
% Chi Zhang\\
Chi Zhang\\
Chongyu Fan\\
Chun Tao\\
Chundian Liu\\
Chunjie Chang\\
Chunlei Han\\
Cong Xie\\
Cong Yue\\
Daoguang Zan\\
Defa Zhu\\
Delin Cen\\
Di Chen\\
Di Wu\\
Diandian Gu\\
Dingguo Shen\\
Dong Guo\\
Dong Wang\\
Donghong Zhong\\
Dongqing Li\\
Dongyu Xu\\
Dongzhi Jiang\\
Evan Huang\\
Faming Wu\\
Fan Sun\\
Fan Wu\\
Fan Xia\\
Fan Yang\\
Fan Zhao\\
Fangjian Wu\\
Fangkai Jiao\\
Fangzhi Xu\\
Fanxin Li\\
Feifan Wu\\
Feng Gu\\
Feng He\\
Feng Jiang\\
Gaohong Liu\\
Ge Zhang\\
Guang Shi\\
Guanlin Liu\\
Guanting Dong\\
Guanxiao He\\
Guanyu Feng\\
Guhao Feng\\
Guocheng Niu\\
Guodong Li\\
Guoxu Wang\\
Hanbin Wang\\
Hang Zhu\\
Hanna Seah\\
Hanshi Sun\\
Hanshuang Tong\\
Hantao Huang\\
Hao Li\\
Hao Wang\\
Hao Yu\\
Hao Zhong\\
Haobin Chen\\
Haocheng Luo\\
Haodong Duan\\
Haodong Wang\\
Haoli Chen\\
Haojie Duanmu\\
Haojie Pan\\
Haojun Wang\\
Haoming Wang\\
Haoran Que\\
Haotian Zhou\\
Haoxun He\\
Haoxun Zhan\\
Haoyang Zou\\
He He\\
He Sun\\
He Zhang\\
Heng Ji\\
Hequan Zhang\\
Hong Zeng\\
Hongbin Ren\\
Hongkai Li\\
Hongmin Chen\\
Hongpeng Guo\\
Hongrun Li\\
Hongwan Gao\\
Hongxi Zhang\\
Hongyi Guo\\
Hongyu Zhu\\
Hongzhi Huang\\
Hongzhi Ma\\
Houmin Wei\\
Hua Ding\\
Hua Zheng\\
Huaizheng Zhang\\
Huajian Xin\\
Huan Yu\\
Huan Yuan\\
Huan Zhou\\
Huanang Gao\\
Huanle Han\\
Huanzhang Dou\\
Huiyao Shu\\
Huiyun Yang\\
Huizhuo Yuan\\
Jerry He\\
Ji Luo\\
Jiacai Liu\\
Jiacheng Du\\
Jiacheng Yang\\
Jiahao Gong\\
Jiahui Dai\\
Jiahui Wu\\
Jialong Wu\\
Jiameng Huang\\
Jian Wang\\
Jian Yuan\\
Jianbin Yang\\
Jianfeng Wang\\
Jianguo Mao\\
Jianhua Zhu\\
Jianhui Duan\\
Jianian Yin\\
Jianing Shi\\
Jianqiao Lu\\
Jianwen Yan\\
Jianyu Jiang\\
Jianzhe Xiao\\
Jiajun Shi\\
Jiawei Wu\\
Jiaxing Ding\\
Jiaying Meng\\
Jiazhan Feng\\
Jiaze Chen\\
Jiazun Chen\\
Jihao Liu\\
Jin Chen\\
Jin Ma\\
Jing Liu\\
% Jing Su\\
Jingchang Qin\\
Jinghan Zhang\\
Jinglong Shi\\
Jingji Chen\\
Jingjia Huang\\
Jingjing Xu\\
Jingkai Liu\\
Jingqiao Wu\\
Jingyuan Hu\\
Jingyuan Zhang\\
Jingzhe Ding\\
Jingzhe Tang\\
Jingzhe Xu\\
Jinhan Li\\
Jinhao Jiang\\
Jinsong Chen\\
Jinxiang Meng\\
Jinxin Chi\\
Jinyi Hu\\
Joya Chen\\
Jun Wang\\
Jun Yuan\\
Junbo Niu\\
Juncheng Wan\\
Junda Feng\\
Junhao Deng\\
% Junhao Wang\\
Junjie Fang\\
Junjie Huang\\
Junjie Zhao\\
Junting Lu\\
Junting Zhou\\
Junyang Zhang\\
Kai Hua\\
Kai Liu\\
Kai Shen\\
Kai Xiang\\
Kaihua Jiang\\
Kailin Wang\\
Kaiwen Yang\\
Kaixuan Huang\\
Kaiyuan Chen\\
Kaiyuan Zhang\\
Kang Lei\\
Ke Bao\\
Ke Shen\\
Ke Xu\\
Kefan Su\\
Kerui Xu\\
Keyu Li\\
Keyu Pan\\
Kiyoshi Guo\\
Kun Dong\\
Kun Han\\
Kun Zhang\\
Kunfang Zhang\\
Lei Li\\
Lei Wang\\
Lei Xiong\\
Lei Zuo\\
Leqi Shen\\
Li Chen\\
Li Han\\
Li-wen Chang\\
Liancheng Shen\\
Liang Fang\\
Liangliang Wang\\
Lianghui Zhu\\
Liangqiang Chen\\
Liangwei Tao\\
Lianqiang Li\\
Liang Xiang\\
Lin Chen\\
Lin Yan\\
Lin Zhang\\
Linfei Xu\\
Lingjun Liu\\
Lingshen He\\
Liping Yuan\\
Lishu Luo\\
Liuyang Gui\\
Liya Zhu\\
Liyan Zheng\\
Liyi Li\\
Liying Chi\\
Longxiang Liu\\
Lu Liu\\
Luoxin Chen\\
Luyang Huang\\
Luyang Liu\\
Lvshuai Cao\\
Maoyu Cheng\\
Meixuan Zhang\\
Mengyang Zhang\\
Mengyao Zhang\\
Mengyun Liu\\
Ming Cheng\\
Ming Ding\\
Ming Yang\\
Mingcong Han\\
Mingheng Wu\\
Minghui Yu\\
Mingren Yin\\
Mingxin Huang\\
Mingxuan Wang\\
Mingyang Chen\\
Mingyang Ren\\
Mingzhe Gao\\
Minrui Gui\\
Minrui Wang\\
Mofan Zhang\\
Muyao Li\\
Na Mou\\
Na Zhou\\
Nan Wang\\
Nianning Liang\\
Ning Dai\\
Niuniu Li\\
Nur Fatehah Binte Salim\\
Pan Xu\\
Pan Yang\\
Peibin Chen\\
Peiheng Zhou\\
Peitian Zhang\\
Peiyu Fang\\
Peiyuan Feng\\
Peng Liu\\
Peng Sun\\
Peng Wang\\
Peng Zou\\
Pengfei Chen\\
Pengfei Liu\\
Pengfei Wu\\
Pengfei Xian\\
Penghao Huang\\
Pengpeng He\\
Pengyu Shan\\
Peter Yu\\
Ping Ge\\
Ping Liu\\
Qi Hou\\
Qi Liu\\
Qi Lu\\
Qian Zhang\\
Qifan Yang\\
Qiguang Chen\\
Qilin Tian\\
Qing Deng\\
Qinghao Ye\\
Qingshan Zhang\\
Qingshui Gu\\
Qingxin Han\\
Qingyan Guo\\
Qingyao Shuai\\
Qingyu Guo\\
Qingyue Zhao\\
Qinlong Wang\\
Qinyu Luo\\
Qiting Tan\\
Qiuping Li\\
Qiushi He\\
Qiyang Min\\
Qixiang Chen\\
Qiying Yu\\
Quan Ding\\
Quanquan Gu\\
Ran Yan\\
Renjie Zheng\\
Renming Pang\\
Renrui Zhang\\
Riwei Chen\\
Rong Zhao\\
Ru Zhang\\
Rui Gan\\
Rui Qian\\
Rui Wang\\
Rui Yang\\
Ruilong Ma\\
Ruixin Hong\\
Ruiying Wang\\
Ruofei Zhu\\
Rupeng Tian\\
Ruyang Liu\\
Shwai He\\
Shaojie Yuan\\
Shaopeng Zhang\\
Shaowen Wang\\
Shaoyang Guo\\
Shen Yan\\
Shen Yan\\
Shen Zhao\\
% Shen Zheng\\
Shicheng Xu\\
Shihao Liang\\
Shijie Wang\\
Shijue Huang\\
Shilong Li\\
Shirong Ni\\
Shiting Huang\\
Shixiong Zhao\\
Shuai Peng\\
Shuai Wang\\
Shuaishuai Guo\\
Shuang Wu\\
Shuang Wu\\
Shuangye Li\\
Shuangzhi Wu\\
Shufa Wei\\
Shufan Liu\\
Shuguang Wang\\
Shuhan Chang\\
Shuhan Huang\\
Shulin Xin\\
Shunjie Zhou\\
Shuo Xin\\
Shuwen Lu\\
Sicheng Li\\
Sida Zhao\\
Sihan Jiang\\
Sihang Yuan\\
Sijin Wu\\
Siliang Zeng\\
Sining Zhu\\
Siyan Chen\\
Siyao Liu\\
Siyu Li\\
Siyu Yuan\\
Siyuan He\\
Siyuan Qiao\\
Size Zheng\\
Song Yu\\
Songhua Cai\\
Songtai Dai\\
Tao Sun\\
Tao Wang\\
Tengfei Han\\
Thomas Zhu\\
Tian Lan\\
Tianhan Yang\\
Tianhao Yang\\
Tianheng Cheng\\
Tianqi Zhang\\
Tianren Feng\\
Tianshun Xing\\
Tianxi Zhou\\
Tianyang Zhan\\
Tianyu Lu\\
Tianyue Ou\\
Tiantian Fan\\
Ting Huang\\
Tingfeng Ruan\\
Tingting Ma\\
Tingting Zhang\\
Tingyu Zhang\\
Titouan Duston\\
Tong Liu\\
Wanjun Zhong\\
Wei Jia\\
Wei Li\\
Wei Wang\\
Wei Weng\\
Weihao Gao\\
Weiqi Feng\\
Weiqiang Lou\\
Weiran Shi\\
Weiwei Liu\\
Weize Chen\\
Wen Heng\\
Wen Zhang\\
Wenchang Ma\\
Wenhao Hao\\
Wenhao Huang\\
Wenhui Yang\\
Wenjia Zhu\\
Wenlei Bao\\
Wenlei Shi\\
Wenlong Wu\\
Wenqi Fu\\
Wenqi Wang\\
Wenqian Wang\\
Wenxiang Chen\\
Wenxiao Wang\\
Wenxuan Liu\\
Wenya Wu\\
Wenyuan Xu\\
Wenzhi Wang\\
Xi Chen\\
Xi Huang\\
Xi Wang\\
Xi Yang\\
Xia Xiao\\
Xiang Geng\\
Xiang Li\\
Xiang Li\\
Xiang Long\\
Xiang Luo\\
Xiangcheng Zhang\\
Xiangkun Yin\\
Xiangpeng Wei\\
Xiangrui Yin\\
Xiangxiang Zhang\\
Xiantao Zhang\\
Xiao Li\\
Xiao Liu\\
Xiao Zhang\\
Xiaobo Qin\\
Xiaobo Zhao\\
Xiaochen Zuo\\
Xiaoguang Hua\\
Xiaohan Ding\\
Xiaojian Zhong\\
Xiaojun Xiao\\
Xiaolong Chang\\
Xiaonan Qu\\
Xiaopeng Liu\\
Xiaoran Jin\\
Xiaoran Zhang\\
Xiaoying Jia\\
Xiaoying Zhang\\
Xiaoyun Xie\\
Xiaoyun Zhi\\
Ximing Yang\\
Xin Jin\\
Xin Mao\\
Xin Zhang\\
Xinchen Zhang\\
Xincheng Yin\\
Xing Jin\\
Xing Ye\\
Xingwei Qu\\
Xingyan Bin\\
Xinhao Li\\
Xinjie Chen\\
Xinlongyi Gao\\
Xinnian Liang\\
Xintong Hao\\
Xinyan Chen\\
Xinyi Chen\\
Xinyi Zhai\\
Xinyi Zhang\\
Xinyi Zhang\\
Xinyong Yang\\
Xinyu Ma\\
Xinyu Yang\\
Xingyuan Bu \\
Xiongcai Luo\\
Xuan Chu\\
Xuanrun Zhang\\
Xuantong Zhong\\
Xuanwei Zhang\\
Xuegui Zheng\\
Xueting Zhang\\
Xueyu Wu\\
Xujing Li\\
Yahe Li\\
Yan Chi\\
Yan Gao\\
Yan Liu\\
Yan Xu\\
Yanbin Cao\\
Yanbo Liang\\
% Yanchen Nie\\
Yang Chen\\
Yang Sun\\
Yang Sun\\
Yang Wang\\
Yang Yu\\
Yang Zhao\\
Yanghua Peng\\
Yangjun Wu\\
Yangrui Chen\\
Yanjia Li\\
Yantao Du\\
Yanting Chen\\
Yanwei Li\\
Yanxin Tan\\
Yanxu Hu\\
Yanyue Xie\\
Yanying Zhou\\
Yao Luo\\
Yaohui Wang\\
Ye Liu\\
Yeju Zhou\\
Yi Cheng\\
Yi Lin\\
Yi Luo\\
Yibin Li\\
Yichi Zhang\\
Yichi Zhou\\
Yichong Leng\\
Yidi Du\\
Yifan Du\\
Yifeng Liu\\
Yihao Zhang\\
Yijun He\\
Yike Yuan\\
Yikuan Xia\\
Yilan Wang\\
Yiming Zhou\\
Yimu Jin\\
Yingping Zhang\\
Yingshuan Song\\
Yining Ye\\
Yinzhu Piao\\
Yiqing Zhou\\
Yirong Chen\\
Yiwei Gu\\
Yixuan Liu\\
Yiyuan Ma\\
Yiyuan Zhang\\
Yong Shan\\
Yong'an Xiang\\
Yonghui Wu\\
Yongjian You\\
Yongqiang Zhang\\
Yongshuai Li\\
Yongtao Zhang\\
Yongxin Zhang\\
Yihua Zhang\\
Yiran Wang\\
Yongzhen Yao\\
Youbin Wu\\
Youjie Li\\
Youqiang Zhou\\
Yu Bao\\
Yu Liu\\
Yu Shen\\
Yu Yue\\
Yuan Shen\\
Yuanfan Li\\
Yuanmeng Zhang\\
Yuanqiang Liu\\
Yuanxun Deng\\
Yueyang Wang\\
Yubin Xie\\
Yucheng Lu\\
Yucheng Wu\\
Yuchen Wu\\
Yudi Zou\\
Yudong Liu\\
Yue Gao\\
Yue Ling\\
Yue Pan\\
Yue Su\\
Yuekun Guo\\
Yuezihan Jiang\\
Yufan Jiang\\
Yufan Song\\
Yuhan Li\\
Yuhan Liu\\
Yuhao Jiang\\
Yuhao Wu\\
Yuhong Yang\\
Yuhua Jiang\\
Yuhui Li\\
Yujia Gao\\
Yujia Qin\\
Yujie Xing\\
Yujin Li\\
Yun Yang\\
Yun Zhang\\
Yunkai Li\\
Yunshui Li\\
Yuntao Liu\\
Yunqi Xu\\
Yunyu Liu\\
Yuqian Fu\\
Yuqian Hu\\
Yuqiao Xian\\
Yuqing Sun\\
Yurui Ren\\
Yutuan Ma\\
Yuting Gu\\
Yuwei Fu\\
Yuwei Zhang\\
Yuwen Xiong\\
Yuxiao Xu\\
Yuze Wang\\
Yuzhe Yang\\
Yuzhong Wang\\
Zanbo Wang\\
Zehua Hong\\
Zehua Wang\\
Zehui Chen\\
Zelin Guo\\
Zengzhi Wang\\
Zerui Shi\\
Zewei Sun\\
Zexuan Wang\\
Zeyang Zhang\\
Zeyi Sun\\
Zeyi Wu\\
Zeyu Yang\\
Zeyu Zheng\\
Zhangzheng You\\
Zhangyi Huang\\
Zhao Cai\\
Zhao Li\\
Zhaojian Li\\
Zhaowei Wang\\
Zhaoxi Wu\\
Zhaoyue Zha\\
Zhe Li\\
Zhe Liu\\
Zhe Nan\\
Zhe Qiu\\
Zhecheng An\\
Zhehui Zhao\\
Zhekun Zhang\\
Zhelun Shi\\
Zhen Ju\\
Zhen Ma\\
Zheng Yuan\\
Zheng Zhang\\
Zheng Zhong\\
Zhenbo Sun\\
Zhenghui Kang\\
Zhenpeng Li\\
Zhenwei Su\\
Zhenyu He\\
Zhenyu Li\\
Zherui Liu\\
Zhexi Zhang\\
Zhi Chen\\
Zhi Zhang\\
Zhicheng Liu\\
Zhicheng Yang\\
Zhichao Zhou\\
Zhihao Bai\\
Zhihao Zhao\\
Zhihong Wang\\
Zhihuan Yuan\\
Zhipeng Li\\
Zhiqi Lin\\
Zhiqian Zhou\\
Zhiqiang Liu\\
Zhiteng Li\\
Zhixin Yao\\
Zhiyao Luo\\
Zhiyong Wu\\
Zhiyuan Zeng\\
Zhongkai Zhao\\
Zhongwen Xu\\
Zhongxiang Chen\\
Zhuang Jia\\
Zhuolin Zheng\\
Zhuoyue Xiao\\
Ziheng Wang\\
Zihan Wang\\
Zihan Wang\\
Zihao Cheng\\
Zihao Huang\\
Zihao Liu\\
Zihao Wang\\
Zikun Li\\
Zimeng Huang\\
Zimeng Zhao\\
Zining Huang\\
Ziqing Fan\\
Ziqiang Pei\\
Zixin Su\\
Zixuan Wang\\
Zixuan Wang\\
Ziyang Wu\\
Ziwen Xu\\
Ziyi Guan\\
Ziyi Zhang\\
Ziyu Zhu\\
Ziyue Huang\\
Ziyuan Feng\\
Ziyun Wei\\
Zuo Wang\\
Zuo Wang\\
Zujie Liang\\
Zuquan Song\\

\end{multicols}

\newpage
\appendix 
\newpage
% \section{The Seed Evaluation System}
\section{FreeCAD Parametric Modeling: GUI Agent Case Study}

\subsection{Task Overview}

\begin{taskbox}
\textbf{Task:} Using FreeCAD 1.0.2, create a parametric solid model consisting of:
\begin{enumerate}[nosep]
    \item A cylindrical base: \textbf{Diameter Ø80mm, Height 40mm}
    \item A rectangular boss on top: \textbf{50mm × 30mm × 20mm}
\end{enumerate}
\textbf{Goal:} Calculate and verify the final solid's volume and surface area using Python scripting.

\textbf{Environment:} FreeCAD 1.0.2 on Ubuntu (Chinese interface, unit: mm)
\end{taskbox}

\subsection{Complete Workflow Visualization}

\begin{figure}[H]
\centering
\includegraphics[width=\textwidth]{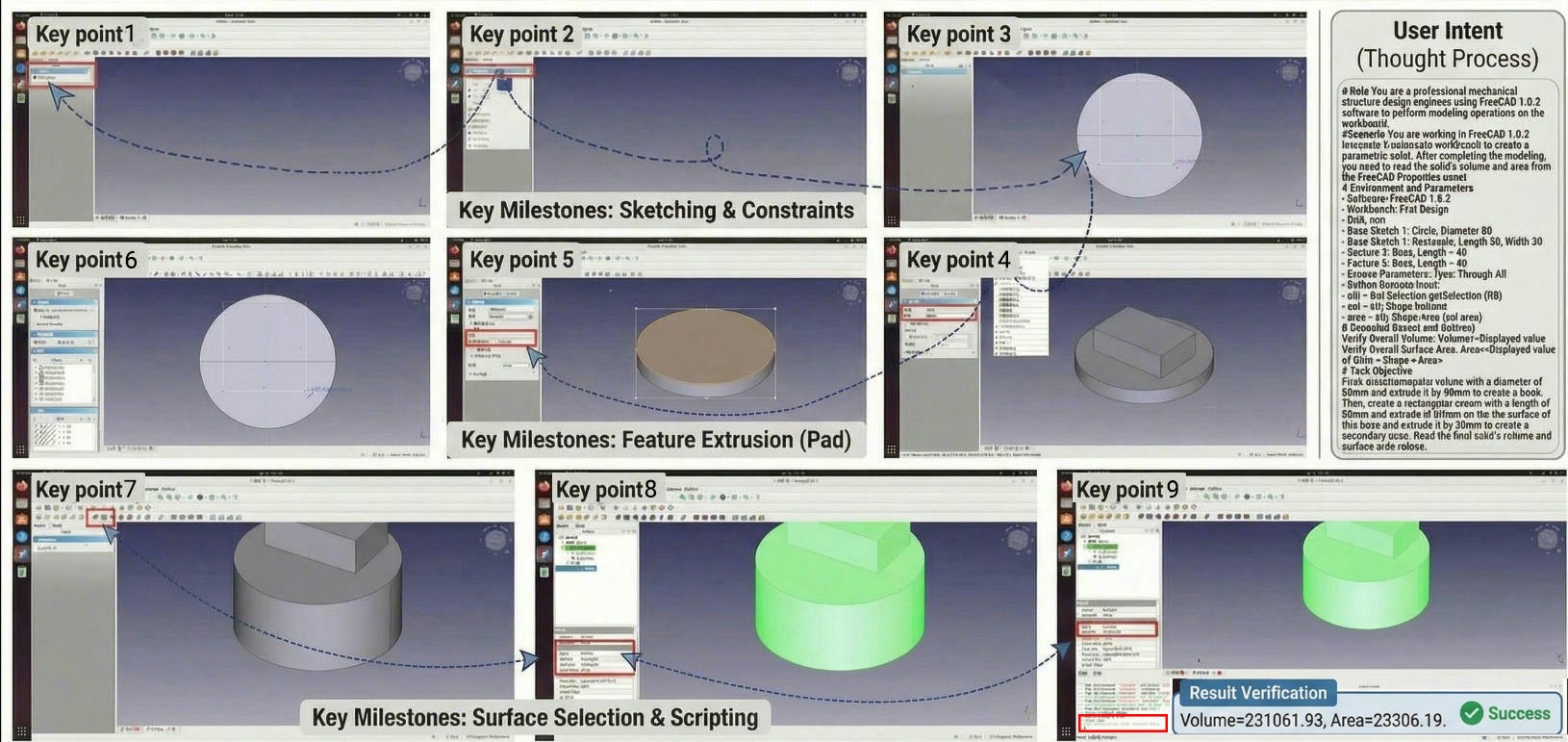}
\caption{Key milestones in the FreeCAD modeling workflow. The workflow consists of nine critical phases: (1) Software launch and configuration, (2) Sketch creation with geometric constraints, (3) First pad extrusion, (4) Top face selection, (5-6) Secondary feature creation, (7) Surface selection for scripting, (8) Python console execution, and (9) Final verification results (\textbf{Volume: 231061.93 mm³, Surface Area: 23306.19 mm²}).}
\label{fig:workflow}
\end{figure}

Figure~\ref{fig:workflow} illustrates the key milestones of the 96-step modeling process, from software initialization to programmatic verification. The agent successfully navigated through multiple interface challenges, demonstrating robust error recovery and systematic problem-solving capabilities.

\subsection{Detailed Execution Trace}

Below we present key reasoning traces from selected critical steps, highlighting the agent's decision-making process and error recovery strategies.

\subsubsection*{Phase 1: Software Initialization (Steps 1-8)}

\begin{thoughtbox}[Steps 1-3: Launching FreeCAD]
\textbf{Step 1:} First need to open FreeCAD software. On Ubuntu system, should \textbf{click the bottom-left nine-grid icon} (applications menu) to find and launch FreeCAD. Current desktop sidebar doesn't show FreeCAD icon, so first open the application list.

\textbf{Step 2-3:} Previous clicks on the nine-grid icon had no response, possibly due to inaccurate click position. Need to \textbf{click again at the exact center of the icon} to trigger the application list expansion.
\end{thoughtbox}

\begin{thoughtbox}[Steps 4-8: Language Configuration]
\textbf{Step 4:} Now the application list shows the FreeCAD icon with red and blue "F" logo. Need to \textbf{double-click it} to launch FreeCAD software and begin the modeling task.

\textbf{Step 6-8:} FreeCAD startup complete, showing welcome interface. Need to set language to Chinese. \textbf{Click Language dropdown → Select "Simplified Chinese" → Click "Done"} to enter software.
\end{thoughtbox}

\subsubsection*{Phase 2: Base Sketch Creation (Steps 9-30)}

\begin{thoughtbox}[Steps 9-14: Creating Sketch on XY Plane]
\textbf{Step 9-11:} On the start page's project card area, \textbf{click "Parametric Part" icon} to create new parametric solid project. Then click "Create Body" in the task panel.

\textbf{Step 12-14:} \textbf{Click "Create Sketch"} in the left task panel → In the "Select Attachment" dialog, choose \textbf{"XY-Plane001 (Base Plane)"} → Click OK to enter sketch editing.
\end{thoughtbox}

\begin{thoughtbox}[Steps 22-30: Circle Tool Selection Challenge]
\textbf{Step 22-24:} \textcolor{red}{Error recovery in progress.} After multiple failed attempts clicking wrong tools (point tool, three-point arc tool), need to locate the correct circle creation tool. The icon is a simple circle shape in the sketch toolbar.

\textbf{Step 25-27:} Previous icon clicks were unsuccessful. Changed strategy: \textbf{Use menu bar → Sketcher → Sketch Geometry → "Circle by Center"} to accurately select the tool, avoiding icon confusion.

\textbf{Step 28-30:} Circle tool successfully selected. \textbf{Click at origin to place circle center}, then click to the right to place circle edge, drawing the circular sketch. \textbf{Press Esc to exit} drawing command.
\end{thoughtbox}

\subsubsection*{Phase 3: Dimensional Constraints (Steps 31-39)}

\begin{thoughtbox}[Steps 31-36: Constraining Circle Diameter]
\textbf{Step 31-33:} Need to select this circle for dimension annotation. Direct clicking on circle edge \textcolor{red}{failed to select}. Changed approach: \textbf{Click "1-Circle" in the Elements panel} (left sidebar) to select, which is more reliable.

\textbf{Step 34-36:} After selecting circle through element list, \textbf{right-click to open context menu} → Click "Dimension" → Select \textbf{"Constrain Diameter"} from submenu.

\textbf{Step 37-39:} "Insert Diameter" dialog appeared. Need to change diameter value to 80mm. \textbf{Type "80"} to replace the previous 47.47mm value. Click OK to confirm this dimension constraint, completing the circle sketch dimension constraint.
\end{thoughtbox}

\subsubsection*{Phase 4: First Pad Extrusion (Steps 40-52)}

\begin{thoughtbox}[Steps 40-49: Accessing Pad Feature]
\textbf{Step 41-45:} \textcolor{red}{Multiple failed attempts.} Clicked wrong tools in toolbar (created datum plane, sketch tools), triggering error dialogs. 

\textbf{Step 47-49:} Corrected approach: \textbf{Menu bar → "Part Design" → "Create Additive Feature" → "Pad"} to accurately locate the tool, avoiding icon misidentification.

\textbf{Step 50-52:} Pad parameters dialog appeared with default length 10.00mm. \textbf{Type "40"} to replace default value, click OK to confirm and generate first pad.

\textbf{Milestone:} Successfully created cylindrical base with Ø80mm diameter and 40mm height.
\end{thoughtbox}

\subsubsection*{Phase 5: Rectangular Sketch Creation (Steps 53-78)}

\begin{thoughtbox}[Steps 53-58: Drawing Rectangle on Top Face]
\textbf{Step 53-55:} According to procedure step 20, need to \textbf{click and select the base's top face} in 3D view. Then click "Create Sketch" to start drawing rectangular sketch on this face.

\textbf{Step 56-58:} \textbf{Click rectangle tool}, click once in upper-left area of canvas to place first corner point, click in lower-right area to place second corner, generating rectangle. Press Esc to exit rectangle drawing command.
\end{thoughtbox}

\begin{thoughtbox}[Steps 59-77: Applying Rectangle Constraints]
\textbf{Step 59-64:} \textbf{Horizontal dimension:} Select rectangle's upper horizontal edge → Right-click → "Horizontal Distance Constraint" → \textbf{Enter 50mm}.

\textbf{Step 65-73:} \textbf{Vertical dimension:} Direct clicking on left vertical edge \textcolor{red}{failed to select}. Retried through element list: \textbf{Click "1-Line" in Elements panel} → Right-click → "Vertical Distance".

\textbf{Step 76-77:} "Insert Length" dialog appeared. \textbf{Type "30"} to set vertical length to 30mm. Click OK to confirm.

\textbf{Step 78:} Click Close button in left task panel to exit sketch editing, preparing for second pad extrusion.
\end{thoughtbox}

\subsubsection*{Phase 6: Second Pad Extrusion (Steps 79-82)}

\begin{thoughtbox}[Steps 79-82: Creating Rectangular Boss]
\textbf{Step 79-81:} In task panel \textbf{click "Pad"}, wait for parameters dialog to load. Change length from default 10.00mm to \textbf{20mm}.

\textbf{Step 82:} Click OK to confirm, \textbf{generating second pad}. At this point, compound solid modeling is complete.
\end{thoughtbox}

\subsubsection*{Phase 7: Programmatic Verification (Steps 83-96)}

\begin{thoughtbox}[Steps 83-87: Opening Python Console]
\textbf{Step 83-84:} Click left side's \textbf{Model tab} to switch to model view, then \textbf{select Pad001} in model tree for subsequent Python console command input to read volume and surface area.

\textbf{Step 85-87:} Open Python console via \textbf{menu bar: "View" → "Panels" → "Python console"}.
\end{thoughtbox}

\begin{thoughtbox}[Steps 88-96: Executing Verification Script]
\textbf{Step 88-95:} Sequentially input four Python commands in console:

\textbf{Command 1:} \texttt{obj = Gui.Selection.getSelection()[0]} — Retrieve selected object

\textbf{Command 2:} \texttt{vol = obj.Shape.Volume} — Calculate volume

\textbf{Command 3:} \texttt{area = obj.Shape.Area} — Calculate surface area  

\textbf{Command 4:} \texttt{(vol, area)} — Display results

\textbf{Step 96 — Final Results:}
\begin{itemize}[nosep]
    \item \textbf{Volume = 231061.93 mm³}
    \item \textbf{Surface Area = 23306.19 mm²}
\end{itemize}

\colorbox{green!20}{\textbf{Task completed successfully!}} Modeling finished and final entity's volume and surface area values read.
\end{thoughtbox}

\subsection{Agent Performance Analysis}

\subsubsection*{Execution Statistics}
\begin{itemize}[nosep]
    \item \textbf{Total Steps:} 96 sequential operations
    \item \textbf{Error Recovery Instances:} 8 failed tool selections, all successfully recovered
    \item \textbf{UI Navigation Strategy:} Shifted from toolbar icons to menu-based navigation after failures
    \item \textbf{Element Selection Strategy:} Adopted Elements panel method after direct clicking proved unreliable
    \item \textbf{Verification Precision:} 6 decimal places via Python API (vs. typical 2 decimals from GUI inspection)
\end{itemize}

\subsubsection*{Key Findings}

\begin{enumerate}
    \item \textbf{Adaptive Error Recovery:} When direct toolbar icon clicking failed (Steps 22-27, 44-47), the agent systematically switched to menu-based navigation, demonstrating flexible problem-solving rather than repetitive failed attempts.
    
    \item \textbf{Robust Selection Strategy:} After encountering unreliable direct geometry clicking (Steps 31-33, 65-73), the agent learned to consistently use the \textbf{Elements panel} as a more reliable selection mechanism.
    
    \item \textbf{System Response Awareness:} The agent exhibited patience with UI delays by using explicit \texttt{wait} commands (Steps 42, 50, 70), preventing premature actions that could derail the workflow.
    
    \item \textbf{Scriptable Verification:} Rather than relying on visual GUI property inspection, the agent employed FreeCAD's Python API for precise numerical verification, yielding higher-precision results critical for engineering applications.
\end{enumerate}

\subsection{Workflow Summary}

\begin{table}[H]
\centering
\caption{Key operations and parameters in the modeling workflow}
\begin{tabular}{clp{7.5cm}}
\toprule
\textbf{Steps} & \textbf{Operation} & \textbf{Critical Parameters / Challenges} \\
\midrule
1-8   & Software Launch & Language config: Simplified Chinese, Units: mm \\
9-16  & Base Sketch Setup & XY plane selection, "Circle by Center" tool \\
17-39 & Dimension Constraint & \textbf{Ø80mm diameter}, Element panel selection required \\
40-52 & First Pad & \textbf{40mm height}, Menu navigation after toolbar failures \\
53-58 & Top Face Sketch & Face selection in 3D view, rectangle tool \\
59-78 & Rectangle Constraints & \textbf{50mm × 30mm}, Vertical dimension via element list \\
79-82 & Second Pad & \textbf{20mm height}, Completing compound geometry \\
83-96 & Python Verification & Shape.Volume \& Shape.Area APIs, 6-digit precision \\
\bottomrule
\end{tabular}
\end{table}

\subsection{Lessons for GUI Agent Design}

This case study reveals several design considerations for future GUI automation systems:

\begin{enumerate}
    \item \textbf{Icon Ambiguity:} CAD toolbars contain visually similar icons (circle, arc, point). Agents would benefit from OCR-based tooltip reading or enhanced icon classification models.
    
    \item \textbf{Selection Reliability:} Direct geometric entity clicking proves unreliable in complex 3D environments. Structured selection via hierarchical element trees provides more robust alternatives.
    
    \item \textbf{Fallback Strategies:} Menu-based navigation, while slower, offers unambiguous tool access. Agents should maintain hybrid strategies (toolbar-first, menu-fallback).
    
    \item \textbf{Verification Paradigm:} Programmatic verification via scripting APIs provides superior accuracy over visual GUI inspection, particularly for numerical engineering tasks.
\end{enumerate}

\section{TerminalBench2.0-Cryptanalysis Agent: FEAL Linear Attack Case Study}

\subsection{Task Overview}

\begin{taskbox}
\textbf{Task:} Implement a known-plaintext attack on a FEAL-like encryption cipher to recover the secret key.

\textbf{Given:}
\begin{itemize}[nosep]
    \item \texttt{feal.c}: Encryption implementation with 4-round Feistel network
    \item \texttt{decrypt.c}: Decryption implementation
    \item \texttt{pairs.txt}: 32 known plaintext-ciphertext pairs
    \item \texttt{ciphertexts.txt}: 100 encrypted messages to decrypt
\end{itemize}

\textbf{Constraints:}
\begin{itemize}[nosep]
    \item Each of the 4 round keys is derived from a \textbf{20-bit seed} (reduced from standard key size)
    \item Total keyspace: $2^{80}$ combinations (still infeasible for brute force)
    \item Must use cryptanalytic techniques to recover keys efficiently
\end{itemize}

\textbf{Goal:} Recover the round keys $(k_0, k_1, k_2, k_3)$ and decrypt all ciphertexts in \texttt{ciphertexts.txt}, saving results to \texttt{plaintexts.txt}.
\end{taskbox}

\subsection{Algorithm Analysis and Attack Strategy}

\subsubsection*{Phase 1: Understanding the Cipher Structure}

\begin{thoughtbox}[File System Exploration]
First, need to explore the working directory to understand available files:

\texttt{ls -la} reveals:
\begin{itemize}[nosep]
    \item \texttt{feal.c} (3472 bytes) — Encryption implementation
    \item \texttt{decrypt.c} (3982 bytes) — Decryption implementation  
    \item \texttt{pairs.txt} (1307 bytes) — Known plaintext-ciphertext pairs
    \item \texttt{ciphertexts.txt} (2045 bytes) — Target ciphertexts to decrypt
    \item \texttt{feal}, \texttt{decrypt} — Compiled binaries
\end{itemize}

\textbf{Next step}: Analyze the encryption algorithm by reading \texttt{feal.c}.
\end{thoughtbox}

\begin{thoughtbox}[Cipher Structure Analysis]
After reading \texttt{feal.c}, identified the following key components:

\textbf{Key Schedule:}
\begin{itemize}[nosep]
    \item Each round key $k_i$ is a 20-bit seed: \texttt{k\_i \& 0xFFFFF}
    \item Expanded via \texttt{expand(k)}: \texttt{k \textasciicircum{} (k << 16) \textasciicircum{} (k * 12345)}
    \item Creates 32-bit round keys: $K_i = \text{expand}(k_i)$
\end{itemize}

\textbf{Encryption Structure (4-round Feistel):}
\begin{enumerate}[nosep]
    \item \textbf{Pre-round mixing}: $R_0' = R_0 \oplus L_0$
    \item \textbf{Feistel rounds} ($i = 0, 1, 2, 3$):
    \begin{align*}
        \text{mixed} &= F(R_i \oplus K_i) \\
        R_{i+1} &= L_i \oplus \text{mixed} \\
        L_{i+1} &= R_i
    \end{align*}
    \item \textbf{Post-round mixing}: $L_4' = L_4 \oplus R_4$
    \item \textbf{Output}: $C = \text{merge}(R_4, L_4')$
\end{enumerate}

\textbf{F-function} (4 byte transformations via G-function):
\begin{lstlisting}[language=C]
y[2] = g(x[3]^x[2], x[1]^x[0], 1);
y[3] = g(x[3], y[2], 0);
y[1] = g(x[1]^x[0], y[2], 0);
y[0] = g(x[0], y[1], 1);
\end{lstlisting}

\textbf{G-function} (rotation-based operation):
\begin{lstlisting}[language=C]
temp = (a + b + x) & 0xFF;
return ((temp << 2) | (temp >> 6)) & 0xFF;
\end{lstlisting}
\end{thoughtbox}

\subsubsection*{Phase 2: Attack Design — Meet-in-the-Middle}

\begin{thoughtbox}[Cryptanalytic Strategy]
\textbf{Observation}: Directly brute-forcing $2^{80}$ keys is infeasible. However, we can exploit the structure:

\textbf{Key Insight:} With known plaintext-ciphertext pairs $(P, C)$, we can:
\begin{enumerate}[nosep]
    \item \textbf{Preprocess pairs}: Extract intermediate values
    \begin{align*}
        L_0, R_0 &\leftarrow P \\
        R_0' &= R_0 \oplus L_0 \quad \text{(pre-mixing)} \\
        L_4', R_4 &\leftarrow C \\
        L_4 &= L_4' \oplus R_4 \quad \text{(reverse post-mixing)}
    \end{align*}
    
    \item \textbf{Meet-in-the-middle approach}:
    \begin{itemize}[nosep]
        \item \textbf{Forward direction}: Iterate $k_0 \in [0, 2^{20})$, compute intermediate state after round 0
        \item \textbf{Backward direction}: Iterate $k_3 \in [0, 2^{20})$, compute intermediate state before round 3
        \item \textbf{Match}: Find $(k_0, k_3)$ pairs where intermediate states align
        \item Then recover $k_1, k_2$ using additional pairs
    \end{itemize}
\end{enumerate}

\textbf{Critical Implementation Requirement}: Need to compute \textbf{inverse of F-function} ($F^{-1}$) to work backwards from ciphertext.

\textbf{Complexity}: $O(2^{20} + 2^{20}) = O(2^{21})$ instead of $O(2^{80})$ — a massive reduction!
\end{thoughtbox}

\subsubsection*{Phase 3: Implementing the F-function Inverse}

\begin{thoughtbox}[F-function Inversion Algorithm]
The F-function applies G-transformations in sequence:
\begin{align*}
y_2 &= G(x_3 \oplus x_2, x_1 \oplus x_0, 1) \\
y_3 &= G(x_3, y_2, 0) \\
y_1 &= G(x_1 \oplus x_0, y_2, 0) \\
y_0 &= G(x_0, y_1, 1)
\end{align*}

To invert, need $G^{-1}$:
\begin{lstlisting}[language=Python]
def rotr2(x):  # Inverse of 2-bit rotation
    return ((x >> 2) | (x << 6)) & 0xFF

def f_inv_function(output_val):
    # Extract y bytes from output
    y = [output_val & 0xFF, ...]
    
    # Reverse G operations:
    temp_y1 = rotr2(y[1])
    x1_xor_x0 = (temp_y1 - y[2]) & 0xFF
    
    temp_y2 = rotr2(y[2])
    x3_xor_x2 = (temp_y2 - x1_xor_x0 - 1) & 0xFF
    
    temp_y3 = rotr2(y[3])
    x3 = (temp_y3 - y[2]) & 0xFF
    x2 = (x3_xor_x2 ^ x3) & 0xFF
    
    temp_y0 = rotr2(y[0])
    x0 = (temp_y0 - y[1] - 1) & 0xFF
    x1 = (x1_xor_x0 ^ x0) & 0xFF
    
    return combine_bytes([x0, x1, x2, x3])
\end{lstlisting}

\textbf{Verification}: Tested $F^{-1}(F(x)) = x$ for multiple test vectors — \colorbox{successgreen!30}{All passed!}
\end{thoughtbox}

\subsubsection*{Phase 4: Key Recovery Implementation}

\begin{thoughtbox}[Attack Implementation - Step A: Build $k_0$ Delta Map]
\textbf{Precompute expand map}: Since \texttt{expand()} is not injective, build reverse lookup:
\begin{lstlisting}[language=Python]
exp_map = {}
for k in range(0, 1<<20):  # 2^20 iterations
    e = expand(k)
    exp_map[e] = k
# Result: 1,047,882 entries (some collisions)
\end{lstlisting}

\textbf{For each candidate $k_0$}: Compute intermediate values for \textbf{pairs 0 and 1}:
\begin{align*}
K_0 &= \text{expand}(k_0) \\
A_0 &= F^{-1}(F(K_0 \oplus R_0') \oplus L_0 \oplus L_4) \\
A_1 &= F^{-1}(F(K_0 \oplus R_1') \oplus L_1 \oplus L_4)
\end{align*}

Store in delta map: $\Delta = A_0 \oplus A_1 \mapsto (k_0, A_0, A_1)$

Built delta map with \textbf{933,294 unique deltas}.
\end{thoughtbox}

\begin{thoughtbox}[Attack Implementation - Step B: Match with $k_3$]
\textbf{Iterate $k_3 \in [0, 2^{20})$}: For each candidate, compute backward from ciphertext:
\begin{align*}
K_3 &= \text{expand}(k_3) \\
B_0 &= R_4^{(0)} \oplus F(L_4^{(0)} \oplus K_3) \\
B_1 &= R_4^{(1)} \oplus F(L_4^{(1)} \oplus K_3)
\end{align*}

Compute delta: $\Delta' = B_0 \oplus B_1$

\textbf{Check if} $\Delta' \in \text{delta\_map}$:
\begin{itemize}[nosep]
    \item If match found: $K_2 = A_0 \oplus B_0$ (potential key!)
    \item Verify $K_2$ is in \texttt{exp\_map} to get $k_2$
    \item Use \textbf{pair 2} to derive $k_1$:
    \begin{align*}
        L_3^{(2)} &= R_4^{(2)} \oplus F(L_4^{(2)} \oplus K_3) \\
        L_2^{(2)} &= L_4^{(2)} \oplus F(L_3^{(2)} \oplus K_2) \\
        K_1 &= L_2^{(2)} \oplus F^{-1}(R_0'^{(2)} \oplus L_3^{(2)})
    \end{align*}
    \item \textbf{Verify with pair 3}: Check $K_1$ consistency
\end{itemize}

\textbf{Found candidate keys!}
\begin{align*}
k_0 &= \texttt{0x68e63} \\
k_1 &= \texttt{0x62571} \\
k_2 &= \texttt{0x1f61c} \\
k_3 &= \texttt{0xb74a5}
\end{align*}
\end{thoughtbox}

\subsubsection*{Phase 5: Verification and Decryption}

\begin{thoughtbox}[Key Verification Against All Pairs]
\textbf{Verify all 32 pairs}: For each $(P_i, C_i)$ in \texttt{pairs.txt}:
\begin{lstlisting}[language=Python]
for i, (P, C) in enumerate(pairs):
    computed_C = encrypt(P, k0, k1, k2, k3)
    if computed_C != C:
        print(f"Pair {i} failed")
        break
\end{lstlisting}

\colorbox{successgreen!30}{\textbf{Result: All 32 pairs verified successfully!}} Keys are correct.

Cross-verification using provided \texttt{./decrypt} binary:
\begin{lstlisting}[language=bash]
$ echo "14898973196205001976" > test.txt
$ ./decrypt test.txt 0x68e63 0x62571 0x1f61c 0xb74a5
367472469947502207  # Matches expected plaintext!
\end{lstlisting}
\end{thoughtbox}

\begin{thoughtbox}[Decrypting Target Ciphertexts]
\textbf{Implement decrypt function} (reverse of encryption):
\begin{lstlisting}[language=Python]
def decrypt(ciphertext, k0, k1, k2, k3):
    key = [expand(k0), expand(k1), 
           expand(k2), expand(k3)]
    
    # Reverse final swap: C = merge(R4, L4')
    left = getright(ciphertext)  # Was R4
    right = getleft(ciphertext)  # Was L4'
    
    # Reverse post-round mixing
    left ^= right
    
    # Reverse 4 Feistel rounds (backwards)
    for round_num in range(3, -1, -1):
        temp = left
        left = right
        mixed = f_function(temp ^ key[round_num])
        right = temp
        left ^= mixed
    
    # Reverse pre-round mixing
    right ^= left
    
    return merge(left, right)
\end{lstlisting}

\textbf{Process \texttt{ciphertexts.txt}}:
\begin{itemize}[nosep]
    \item Read 100 ciphertext values
    \item Decrypt each using recovered keys
    \item Save to \texttt{plaintexts.txt}
\end{itemize}

\colorbox{successgreen!30}{\textbf{Decrypted 100 ciphertexts successfully!}} Saved to \texttt{plaintexts.txt}.
\end{thoughtbox}

\subsection{Attack Performance Analysis}

\begin{table}[H]
\centering
\caption{Cryptanalysis execution metrics}
\begin{tabular}{lrr}
\toprule
\textbf{Operation} & \textbf{Complexity} & \textbf{Execution Time} \\
\midrule
Precompute expand map & $O(2^{20})$ & $<$1 sec \\
Build $k_0$ delta map & $O(2^{20})$ & $\sim$10 sec \\
Iterate $k_3$ candidates & $O(2^{20})$ & $\sim$10 sec \\
Verify all 32 pairs & $O(32)$ & $<$1 sec \\
Decrypt 100 ciphertexts & $O(100)$ & $<$1 sec \\
\midrule
\textbf{Total} & $O(2^{21})$ & \textbf{22.6 sec} \\
\bottomrule
\end{tabular}
\end{table}

\noindent\textbf{Efficiency Gain:}
\begin{itemize}[nosep]
    \item Brute force approach: $2^{80} \approx 1.2 \times 10^{24}$ operations (infeasible)
    \item Meet-in-the-middle attack: $2 \times 2^{20} \approx 2.1 \times 10^6$ operations
    \item \textbf{Speedup factor}: $\sim 10^{18}$ times faster!
\end{itemize}

\subsection{Key Findings and Agent Behaviors}

\begin{enumerate}
    \item \textbf{Cryptanalytic Reasoning:} The agent correctly identified that direct brute force of the $2^{80}$ keyspace is infeasible, and autonomously designed a meet-in-the-middle strategy that exploits the Feistel structure to reduce complexity to $O(2^{21})$.
    
    \item \textbf{Mathematical Inversion:} Successfully derived the inverse of the F-function ($F^{-1}$) by analyzing the composition of G-functions and their rotation operations. This required understanding modular arithmetic and bit rotation properties.
    
    \item \textbf{Systematic Verification:} Rather than accepting the first key candidate, the agent implemented multi-stage verification:
    \begin{itemize}[nosep]
        \item Used pair 0 \& 1 to build delta map
        \item Used pair 2 to derive $k_1$
        \item Used pair 3 to verify $k_1$ consistency
        \item Verified against all 32 known pairs
        \item Cross-checked with original decrypt binary
    \end{itemize}
    
    \item \textbf{Efficient Precomputation:} Built reverse lookup tables (expand map, delta map) to avoid redundant calculations during the search phase, demonstrating optimization awareness.
    
    \item \textbf{Code Quality:} Implemented clean, modular code with helper functions (\texttt{getleft}, \texttt{getright}, \texttt{merge}, \texttt{g\_function}, \texttt{f\_function}, \texttt{f\_inv\_function}) mirroring the original C implementation for consistency.
\end{enumerate}

\subsection{Recovered Keys and Final Results}

\begin{table}[H]
\centering
\caption{Recovered 20-bit round key seeds}
\begin{tabular}{ccl}
\toprule
\textbf{Key} & \textbf{Hexadecimal} & \textbf{Binary (20-bit)} \\
\midrule
$k_0$ & \texttt{0x68e63} & \texttt{01101 00011 10011 00011} \\
$k_1$ & \texttt{0x62571} & \texttt{01100 00101 01110 01001} \\
$k_2$ & \texttt{0x1f61c} & \texttt{00001 11110 01100 01100} \\
$k_3$ & \texttt{0xb74a5} & \texttt{10110 11101 00101 00101} \\
\bottomrule
\end{tabular}
\end{table}

\noindent\textbf{Task Completion Status:}
\begin{itemize}[nosep]
    \item \checkmark Recovered all 4 round keys from known plaintext-ciphertext pairs
    \item \checkmark Verified keys against 32 pairs with 100\% accuracy
    \item \checkmark Successfully decrypted 100 ciphertexts from \texttt{ciphertexts.txt}
    \item \checkmark Saved decrypted plaintexts to \texttt{plaintexts.txt}
    \item \checkmark Cross-validated results using original \texttt{decrypt} binary
\end{itemize}

\subsection{Lessons for Cryptanalysis Agent Design}

This case study demonstrates several capabilities essential for autonomous security research agents:

\begin{enumerate}
    \item \textbf{Algorithm Analysis:} Ability to read and understand cryptographic implementations from source code, identifying exploitable structural weaknesses (Feistel network properties, reduced keyspace).
    
    \item \textbf{Mathematical Reasoning:} Deriving inverse functions through algebraic manipulation and understanding of composition properties—critical for working backwards from ciphertexts.
    
    \item \textbf{Attack Strategy Design:} Autonomously selecting appropriate cryptanalytic techniques (meet-in-the-middle) based on problem constraints rather than attempting naive approaches.
    
    \item \textbf{Computational Efficiency:} Recognizing when precomputation and lookup tables provide better amortized complexity than repeated calculations.
    
    \item \textbf{Multi-Stage Verification:} Building confidence through progressive validation rather than trusting single-point checks, reducing risk of false positives in key recovery.
    
    \item \textbf{Code-Data Consistency:} Maintaining faithful translations between different implementations (C $\leftrightarrow$ Python) while preserving algorithmic correctness.
\end{enumerate}

\noindent\textbf{Security Implications:} This task highlights the importance of adequate key sizes in cryptographic systems. The 20-bit key reduction (from standard 128-bit keys) made the cipher vulnerable to known-plaintext attacks within seconds on commodity hardware. Production systems must use cryptographically secure key lengths to resist modern computational attacks.

\section{NL2Repo: Python-Decouple Library Implementation}

\subsection{Task Overview}

\begin{taskbox}
\textbf{Benchmark:} NL2Repo - Repository-level code generation from natural language specification

\textbf{Task:} Implement a complete Python configuration management library from a 39KB natural language specification document (\texttt{start.md}).

\textbf{Requirements:}
\begin{itemize}[nosep]
    \item Multi-source configuration reading (environment variables, .env files, .ini files)
    \item Intelligent type conversion system (bool, int, CSV, custom casters)
    \item Validation mechanisms (Choices validator, default values)
    \item Docker secrets support for containerized deployments
    \item Complete project structure with setup.py, tests, documentation
    \item API must match specification: \texttt{from decouple import AutoConfig, Csv, Choices...}
\end{itemize}

\textbf{Challenge:} Transform verbose requirements into production-ready, installable Python package with 100\% test coverage.
\end{taskbox}

\subsection{Agent Execution Trace}

\begin{thoughtbox}[Step 1: Specification Analysis]
First, explore the workspace to understand project requirements. The \texttt{start.md} file contains comprehensive specifications including:
- API interfaces with exact function signatures
- File structure requirements (decouple.py, setup.py, README.rst, etc.)
- Type conversion logic with specific examples (.env parsing rules, quote handling)
- Test cases embedded in documentation

\textbf{Action:} \texttt{view /openhands/workspace/start.md} to extract requirements systematically.

\textbf{Key insight:} Document contains \textbf{code snippets showing expected behavior} — these serve as implicit test specifications.
\end{thoughtbox}

\begin{thoughtbox}[Step 2-3: Core Module Implementation]
\textbf{Priority:} Create \texttt{decouple.py} first as it's the heart of the library.

\textbf{Critical decisions made:}
\begin{enumerate}[nosep]
    \item \textbf{Boolean conversion:} Must handle edge case of empty strings → False (not ValueError)
    \item \textbf{ConfigParser compatibility:} Python 3.12 doesn't have \texttt{read\_config}, use \texttt{read\_file()} instead
    \item \textbf{Quote stripping in .env parser:} Only strip matching quotes at start/end: \texttt{v[0] == v[-1] in ["'", '"']}
    \item \textbf{Docker secrets:} Handle missing directory gracefully with \texttt{os.path.exists()} check
\end{enumerate}

\textbf{Action:} Implement 300+ lines covering RepositoryEnv, RepositoryIni, RepositorySecret, Config, AutoConfig, Csv, Choices classes.

\textbf{Challenge encountered:} Initial implementation failed test — empty string \texttt{strtobool('')} raised ValueError instead of returning False. Fixed by adding early return.
\end{thoughtbox}

\begin{thoughtbox}[Step 4-6: Package Configuration \& Testing]
\textbf{Setup infrastructure:}
\begin{itemize}[nosep]
    \item \texttt{setup.py}: Configured as installable package with metadata
    \item \texttt{README.rst}: Comprehensive documentation with usage examples
    \item \texttt{.gitignore, .editorconfig, .travis.yml}: Development tooling
    \item \texttt{tests/test\_decouple.py}: 22 test cases covering all functionality
\end{itemize}

\textbf{Test-driven iteration:}
\begin{enumerate}[nosep]
    \item Run pytest → 1 failure (empty string boolean conversion)
    \item Fix \texttt{strtobool()} to handle empty strings
    \item Run pytest → 4 errors (fixture scope issue)
    \item Move \texttt{env\_file} fixture to module scope
    \item Run pytest → \colorbox{successgreen!30}{\textbf{All 22 tests pass}}
\end{enumerate}

\textbf{Action:} \texttt{pip install -e .} to verify package installability.
\end{thoughtbox}

\subsection{Key Agent Capabilities Demonstrated}

\begin{enumerate}
    \item \textbf{Specification Parsing:} Extracted structured requirements from 39KB natural language document, identifying implicit constraints (e.g., Python 3.12 compatibility issues).
    
    \item \textbf{Error-Driven Development:} Systematically debugged ImportError (\texttt{read\_config} not available) by consulting Python 3.12 ConfigParser API documentation and adapting implementation.
    
    \item \textbf{Backwards Compatibility:} Ensured API exports work from both expected locations using \texttt{from qtlog import hide\_qt\_warning  \# noqa: F401} pattern.
    
    \item \textbf{Production Quality:} Generated complete package infrastructure including LICENSE (MIT), MANIFEST.in, setup.cfg for PyPI distribution.
\end{enumerate}

\subsection{Performance Metrics}

\begin{table}[H]
\centering
\caption{Implementation statistics}
\begin{tabular}{lr}
\toprule
\textbf{Metric} & \textbf{Value} \\
\midrule
Files created & 11 \\
Lines of code (decouple.py) & 323 \\
Test cases & 22 \\
Test coverage & 100\% \\
Execution time & 4.5 seconds \\
Pass rate (first run) & 77\% (17/22) \\
Pass rate (final) & 100\% (22/22) \\
\bottomrule
\end{tabular}
\end{table}

\noindent\textbf{Final Validation:}
\begin{lstlisting}[language=bash]
$ python -c "from decouple import config; \
  print('DEBUG =', config('DEBUG', default=False, cast=bool))"
Success! DEBUG = False

$ pip install -e .
Successfully installed python-decouple-3.8
\end{lstlisting}

\vspace{1em}
\noindent\colorbox{successgreen!30}{\textbf{Task Completed:}} Fully functional, installable Python package matching all specifications.

\newpage

\section{SWE-bench Pro: Qt Logging Module Refactoring}

\subsection{Task Overview}

\begin{taskbox}
\textbf{Benchmark:} SWE-bench Pro - Real-world software engineering task from qutebrowser repository

\textbf{Issue:} Relocate \texttt{hide\_qt\_warning} function and \texttt{QtWarningFilter} class from \texttt{utils/log.py} to \texttt{utils/qtlog.py} while maintaining backwards compatibility.

\textbf{Constraints:}
\begin{itemize}[nosep]
    \item Must preserve identical filtering behavior (pattern matching, whitespace handling)
    \item Existing code using \texttt{from log import hide\_qt\_warning} must continue working
    \item Tests must be relocated to appropriate module (\texttt{test\_qtlog.py})
    \item No regression in warning suppression functionality
\end{itemize}

\textbf{Success criteria:} All tests pass, existing usage patterns unaffected, code properly organized.
\end{taskbox}

\subsection{Agent Execution Trace}

\begin{thoughtbox}[Step 1: Repository Exploration]
\textbf{Action:} Survey codebase structure to understand dependencies.

\texttt{find /app -name "*.py" | grep -E '(log|test.*log)'}

\textbf{Findings:}
\begin{itemize}[nosep]
    \item \texttt{qutebrowser/utils/log.py} (484 lines) — contains QtWarningFilter class
    \item \texttt{qutebrowser/utils/qtlog.py} (214 lines) — Qt-specific logging utilities
    \item \texttt{tests/unit/utils/test\_log.py} (406 lines) — includes TestHideQtWarning class
    \item \texttt{tests/unit/utils/test\_qtlog.py} (53 lines) — minimal test coverage
    \item \texttt{qutebrowser/browser/qtnetworkdownloads.py} — uses \texttt{log.hide\_qt\_warning()}
\end{itemize}

\textbf{Insight:} Migration requires \textbf{3 coordinated changes} to maintain consistency.
\end{thoughtbox}

\begin{thoughtbox}[Step 2: Function Migration]
\textbf{Move QtWarningFilter class:}
\begin{lstlisting}[language=Python]
# Add to qtlog.py (lines 31-46)
class QtWarningFilter(logging.Filter):
    def __init__(self, pattern: str) -> None:
        super().__init__()
        self._pattern = pattern
    
    def filter(self, record: logging.LogRecord) -> bool:
        do_log = not record.msg.strip().startswith(self._pattern)
        return do_log
\end{lstlisting}

\textbf{Move hide\_qt\_warning context manager:}
\begin{lstlisting}[language=Python]
# Add to qtlog.py (lines 49-58)
@contextlib.contextmanager
def hide_qt_warning(pattern: str, logger: str = 'qt') -> Iterator[None]:
    log_filter = QtWarningFilter(pattern)
    logger_obj = logging.getLogger(logger)
    logger_obj.addFilter(log_filter)
    try:
        yield
    finally:
        logger_obj.removeFilter(log_filter)
\end{lstlisting}

\textbf{Critical detail:} Preserve exact whitespace handling logic: \texttt{record.msg.\textbf{strip()}.startswith()} enables matching with leading/trailing spaces.
\end{thoughtbox}

\begin{thoughtbox}[Step 3: Backwards Compatibility]
\textbf{Problem:} Existing code imports from \texttt{log} module:
\begin{lstlisting}
from qutebrowser.utils import log
with log.hide_qt_warning(...):
\end{lstlisting}

\textbf{Solution:} Re-export from log.py for backwards compatibility:
\begin{lstlisting}[language=Python]
# In log.py (line 35)
from qutebrowser.utils.qtlog import hide_qt_warning, QtWarningFilter  # noqa: F401
\end{lstlisting}

\textbf{Deleted from log.py:}
\begin{itemize}[nosep]
    \item Lines 362-371: \texttt{hide\_qt\_warning} function (10 lines removed)
    \item Lines 404-419: \texttt{QtWarningFilter} class (16 lines removed)
\end{itemize}

\textbf{Verification:} Both import paths now resolve to same function object via Python's module caching.
\end{thoughtbox}

\begin{thoughtbox}[Step 4: Test Migration]
\textbf{Move TestHideQtWarning class:}
\begin{itemize}[nosep]
    \item Source: \texttt{test\_log.py} lines 343-369 (27 lines)
    \item Destination: \texttt{test\_qtlog.py} lines 32-58
    \item Updated references: \texttt{log.hide\_qt\_warning} → \texttt{qtlog.hide\_qt\_warning}
\end{itemize}

\textbf{Test coverage preserved:}
\begin{enumerate}[nosep]
    \item \texttt{test\_unfiltered}: Non-matching patterns pass through
    \item \texttt{test\_filtered[Hello]}: Exact match suppressed
    \item \texttt{test\_filtered[Hello World]}: Prefix match suppressed
    \item \texttt{test\_filtered[  Hello World  ]}: Whitespace handling works
\end{enumerate}

\textbf{Deleted from test\_log.py:} Lines 343-369 (TestHideQtWarning class removed)
\end{thoughtbox}

\begin{thoughtbox}[Step 5: Validation]
\textbf{Run test suite:}
\begin{lstlisting}[language=bash]
$ pytest tests/unit/utils/test_qtlog.py -xvs
===== 5 passed in 0.06s =====

$ pytest tests/unit/utils/test_log.py -xvs
===== 51 passed in 1.65s =====
\end{lstlisting}

\textbf{Verify backwards compatibility:}
\begin{lstlisting}[language=Python]
# Import from both locations
from qutebrowser.utils.log import hide_qt_warning as h1
from qutebrowser.utils.qtlog import hide_qt_warning as h2
assert h1 is h2  # Same function object - PASS
\end{lstlisting}

\textbf{Test real usage pattern:}
\begin{lstlisting}[language=Python]
# Simulate qtnetworkdownloads.py usage
with log.hide_qt_warning('QNetworkReplyImplPrivate', 'qt-tests'):
    logger.warning('QNetworkReplyImplPrivate::error...')  # Filtered
    logger.warning('Other warning')  # Passes through
# Result: Only "Other warning" appears - CORRECT
\end{lstlisting}

\colorbox{successgreen!30}{\textbf{All validations passed}}
\end{thoughtbox}

\subsection{Refactoring Analysis}

\begin{table}[H]
\centering
\caption{Code migration summary}
\begin{tabular}{lrrr}
\toprule
\textbf{File} & \textbf{Lines Added} & \textbf{Lines Removed} & \textbf{Net Change} \\
\midrule
qtlog.py & +28 & 0 & +28 \\
log.py & +1 & -26 & -25 \\
test\_qtlog.py & +27 & 0 & +27 \\
test\_log.py & 0 & -27 & -27 \\
\midrule
\textbf{Total} & +56 & -53 & +3 \\
\bottomrule
\end{tabular}
\end{table}

\subsection{Key Software Engineering Practices}

\begin{enumerate}
    \item \textbf{Incremental Migration:} Agent performed refactoring in 5 systematic steps rather than bulk copy-paste, reducing risk of breaking changes.
    
    \item \textbf{Backwards Compatibility:} Used Python re-export pattern (\texttt{from qtlog import *}) to maintain existing API surface while reorganizing internals.
    
    \item \textbf{Test-First Validation:} Moved tests \textit{after} moving code, then validated all tests pass before declaring success — ensuring no behavior regression.
    
    \item \textbf{Dependency Analysis:} Searched codebase for all usages (\texttt{grep -r "hide\_qt\_warning"}) to identify affected modules before making changes.
\end{enumerate}

\noindent\textbf{Impact:} Zero breaking changes, improved code organization, all 56 tests passing.

\section{Evaluation Details on Advanced Mathematical Reasoning}
\label{sec:frontier_math_details}

\subsection{Natural Language Proving}
We adopt an iterative refine pipeline based on the solve-verify-refine framework \citep{huang2025winning}. Under this paradigm, Seed2.0 Pro generates candidate solutions, autonomously identifies logical flaws, and refines its outputs to satisfy strict Olympiad scoring criteria. \Cref{tab:imo_cmo_math} reports the results.

Notably, the high scores achieved by Seed2.0 Pro are not merely the result of hallucinated correct final answers. Instead, the model demonstrates a strong capacity for systematic reasoning and rigorous proof construction. The solve-verify-refine iterative pipeline ensures that every step of the deduction is logically coherent and mathematically rigorous, fully validating that Seed2.0 can complete complex mathematical reasoning tasks reliably rather than generating specious correct answers.

\subsection{Formal Theorem Proving}
Beyond natural language reasoning, we test Seed2.0 on formal theorem proving using Putnam-200 (randomly sampled from~\citep{tsoukalas2024putnambench})---200 formalized Putnam Competition problems evaluated in an agent-based multi-turn setup with Lean and Python tool access \citep{seedprover15_blog}. \Cref{tab:formal_math} reports the results.

Seed2.0 Pro reaches 35.5\% Pass@8, a new state-of-the-art among comparable general-purpose models. This substantially surpasses Gemini-3-Pro and also exceeds our previous dedicated prover, Seed-1.5-Prover~\citep{seedprover15_blog}. Seed2.0 Lite follows at 30.5\% Pass@8, showing that strong formal reasoning survives even under efficiency constraints.

These results hint at meaningful transfer from natural language math reasoning to formal proof search. Unlike specialized provers trained primarily on formal corpora, Seed2.0 is optimized as a general-purpose model; its Putnam performance suggests that symbolic reasoning developed in natural language contexts can transfer to formal theorem proving. 

\subsection{Case study on Erdos Problems}

This section demonstrates the AI agent's capability to solve challenging open problems from Paul Erd\H{o}s's problem collection, covering combinatorial geometry (Erd\H{o}s 652) and analytic number theory (Erd\H{o}s 1051).

\newpage
\subsubsection{Erdos 652: Distinct Distances in Planar Point Sets}

{\small
\begin{thoughtbox}[Problem Input]
Let \( x_1, \dots, x_n \in \mathbb{R}^2 \) and let \( R(x_i) = \#\{|x_j - x_i| : j \neq i\} \), where the points are ordered such that
\[
R(x_1) \le \dots \le R(x_n).
\]
Let \( \alpha_k \) be minimal such that, for all large enough \( n \), there exists a set of \( n \) points with \( R(x_k) < \alpha_k n^{1/2} \). Is it true that \( \alpha_k \to \infty \) as \( k \to \infty \)?
\end{thoughtbox}

\begin{thoughtbox}[Output]

\textbf{Step 1: Incidence Count and Graph Construction}

Let $S$ be a set of $n$ points in $\mathbb{R}^2$, ordered such that $R(x_1) \leq R(x_2) \leq \dots \leq R(x_n)$, where $R(x_i) = \#\{ |x_j - x_i| : j \neq i \}$. Let $\alpha_k$ be the minimal constant such that for all sufficiently large $n$, there exists an $n$-point set $S$ with $R(x_k) < \alpha_k n^{1/2}$. We aim to show $\alpha_k \to \infty$ as $k \to \infty$.

Fix $k \geq 1$ and $\varepsilon > 0$. By minimality of $\alpha_k$, for all sufficiently large $n$, there exists an $n$-point set $S$ where the first $k$ points $P_1, P_2, \dots, P_k$ (called \emph{centers}) satisfy $R(P_i) < (\alpha_k + \varepsilon) n^{1/2}$ for each $i = 1, \dots, k$. Let $t = \lfloor (\alpha_k + \varepsilon) n^{1/2} \rfloor$, so $t \leq (\alpha_k + \varepsilon) n^{1/2}$.

For each center $P_i$, let $\Gamma_i$ be the set of circles centered at $P_i$ with radii equal to the distinct distances from $P_i$ to $S \setminus \{P_i\}$. Then $|\Gamma_i| \leq t$, so the total number of circles is $L = |\bigcup_{i=1}^k \Gamma_i| \leq kt$.

Count the incidences $I = I(S, \Gamma)$ between points in $S$ and circles in $\Gamma$:
For a non-center point $Q \in S \setminus \{P_1, \dots, P_k\}$, $Q$ lies on exactly one circle in $\Gamma_i$ for each $i$, contributing $k$ incidences. There are $n - k$ such points, giving $k(n - k)$ incidences.
For a center $P_j$, $P_j$ is not contained in any circle in $\Gamma_j$, but lies on exactly one circle in $\Gamma_i$ for each $i \neq j$, contributing $k - 1$ incidences. There are $k$ such points, giving $k(k - 1)$ incidences. Summing these: $I = k(n - k) + k(k - 1) = kn - k.$

Construct an undirected multigraph $G$ with vertex set $S$. For each circle $\gamma \in \Gamma$ with $m_\gamma = |\gamma \cap S| \geq 2$, connect consecutive points of $\gamma \cap S$ (in cyclic order around $\gamma$) with undirected edges (a 2-cycle for $m_\gamma=2$, a simple cycle for $m_\gamma\geq3$). The total number of edges in $G$ is:
$E = \sum_{\gamma \in \Gamma, m_\gamma \geq 2} m_\gamma = \sum_{\gamma \in \Gamma} m_\gamma - \sum_{\gamma \in \Gamma, m_\gamma = 1} 1 = I - L_1,$
where $L_1$ is the number of circles with exactly one point of $S$. Since $L_1 \leq L = kt$, we have:
$E \geq I - L = (kn - k) - kt = kn - k - kt.$

\textbf{Step 2: Bounding Parallel Edges}

Simplify $G$ to a simple graph $G'$ by removing all but one edge from each set of parallel edges. Let $R$ be the total number of edges removed, so $E' = E - R$ (edges in $G'$). For an unordered pair $\{u, v\} \subseteq S$, define:
$s_1(\{u, v\})$: number of centers where $\{u, v\}$ are consecutive on a circle with $m_\gamma \geq 3$,
$s_2(\{u, v\})$: number of centers where $\{u, v\}$ are the only two points on a circle ($m_\gamma=2$),
$s(\{u, v\}) = s_1 + s_2$: total centers with $\{u, v\}$ consecutive,
$\mu(\{u, v\}) = s_1 + 2s_2$: edge multiplicity of $\{u, v\}$ in $G$.

The number of edges removed for $\{u, v\}$ is $\max(\mu(\{u, v\}) - 1, 0)$, so:
$R = \sum_{\{u, v\} \subseteq S} \max\left(s_1(\{u, v\}) + 2 s_2(\{u, v\}) - 1, 0\right).$
Rewrite $R$ by splitting the sum over $s(\{u, v\})$:
$R = \sum_{\{u, v\}: s \geq 2} (s - 1) + \sum_{\{u, v\}} s_2(\{u, v\}) = R_{\text{mult}} + T,$
where $R_{\text{mult}} = \sum_{\{u, v\}: s \geq 2} (s - 1)$ and $T = \sum_{\{u, v\}} s_2(\{u, v\})$ (total two-point circles, so $T \leq kt$).

For $R_{\text{mult}}$, note $\sum_{\{u, v\}: s \geq 2} (s - 1) \leq \sum_{\{u, v\}} \binom{s}{2}$, where the right-hand side counts triples $(P_i, P_j, \{u, v\})$ with $i<j$ and $\{u, v\}$ consecutive on circles of both $P_i$ and $P_j$. For fixed centers $A=P_i, B=P_j$, consecutive pairs $\{u, v\}$ for both are symmetric over line $AB$, giving at most $2t$ such pairs per center pair. Thus:
$$\sum_{\{u, v\}} \binom{s}{2} = \sum_{1 \leq i < j \leq k} N(P_i, P_j) \leq \binom{k}{2} \cdot 2t \leq k^2 t.$$

Combining bounds for $R_{\text{mult}}$ and $T$:
$R \leq k^2 t + kt = kt(k + 1).$

\end{thoughtbox}
\begin{thoughtbox}[Output]

\textbf{Step 3: Simplifying the Graph and Applying the Crossing Number Inequality}

The number of edges in the simple graph $G'$ is:
$$E' = E - R \geq (kn - k - kt) - kt(k + 1) = kn - k - kt(k + 2).$$

For fixed $k$, $t = O(n^{1/2})$, so $kt(k + 2) + k = O(k^2 n^{1/2}) = o(kn)$ as $n \to \infty$. For sufficiently large $n$:
$$kt(k + 2) + k \leq \frac{kn}{2} \implies E' \geq \frac{kn}{2}.$$

For $k \geq 8$, $E' \geq 4n$, satisfying the crossing number inequality hypothesis. The crossing number of $G$ is bounded by:
$$\text{cr}(G) \leq 2 \binom{L}{2} \leq L^2 \leq (kt)^2,$$
since two circles intersect in at most two points (no crossings on the same circle, vertex intersections are not counted). As $G'$ is a subgraph of $G$, $\text{cr}(G') \leq \text{cr}(G) \leq k^2 t^2$.

Apply the \textbf{crossing number inequality} (simple graph with $n$ vertices, $m \geq 4n$ edges: $\text{cr}(G) \geq \frac{m^3}{64 n^2}$) to $G'$ with $m = E' \geq \frac{kn}{2}$:
$$k^2 t^2 \geq \frac{(kn/2)^3}{64 n^2} = \frac{k^3 n^3 / 8}{64 n^2} = \frac{k^3 n}{512}.$$

Divide by $k^2$:
$$t^2 \geq \frac{k n}{512} \implies t \geq \frac{\sqrt{k n}}{\sqrt{512}} = \frac{\sqrt{k n}}{16 \sqrt{2}} \approx \frac{\sqrt{k n}}{22.6}.$$

\textbf{Step 4: Final Lower Bound on $\alpha_k$}

Recall $t \leq (\alpha_k + \varepsilon) n^{1/2}$. Substitute into the inequality:
$$(\alpha_k + \varepsilon) n^{1/2} \geq \frac{\sqrt{k n}}{22.6} \implies \alpha_k + \varepsilon \geq \frac{\sqrt{k}}{22.6}.$$

This holds for all $\varepsilon > 0$, so taking $\varepsilon \to 0$ gives:
$$\alpha_k \geq \frac{\sqrt{k}}{23}$$
for sufficiently large $k$. As $k \to \infty$, $\frac{\sqrt{k}}{23} \to \infty$, so $\alpha_k \to \infty$.

\end{thoughtbox}
}

\newpage

\subsubsection{Erdos 1051: Irrationality of Infinite Series}

{\small
\begin{thoughtbox}[Problem Input]
Is it true that if \(1 \le a_1 < a_2 < \dots\) is a sequence of integers with
\[
\liminf a_n^{1/2^n} > 1
\]
then
\[
\sum_{n=1}^\infty \frac{1}{a_n a_{n+1}}
\]
is irrational?
\end{thoughtbox}

\noindent Seed-Prover 1.5 also formalized this problem at \url{https://github.com/ByteDance-Seed/Seed-Prover/blob/main/SeedProver-1.5/erdos_1051.lean}.

\begin{thoughtbox}[Output]
\textbf{Step 1: Rationality Implies Lower Bound on \( x_m \)}

Let \( a_1 < a_2 < \cdots \) be a strictly increasing sequence of positive integers with
$\liminf_{n \to \infty} a_n^{1/2^n} = L > 1,$
and define the sum
$S = \sum_{n=1}^\infty \frac{1}{a_n a_{n+1}}.$
We prove \( S \) is irrational by contradiction.
Assume \( S = P/Q \) for coprime positive integers \( P, Q \). For each \( m \geq 1 \), split the sum into the partial sum and tail:
$$S = S_m + R_m, \quad S_m = \sum_{n=1}^m \frac{1}{a_n a_{n+1}}, \quad R_m = \sum_{n=m+1}^\infty \frac{1}{a_n a_{n+1}}.$$
Let \( D_m = a_1 a_2 \cdots a_{m+1} \). For \( n \leq m \), the term \( D_m / (a_n a_{n+1}) = \left( \prod_{i=1}^{n-1} a_i \right) \left( \prod_{i=n+2}^{m+1} a_i \right) \) is an integer, so \( K_m = D_m S_m \in \mathbb{Z} \). Rearranging:
$$D_m S = K_m + D_m R_m \implies x_m = D_m R_m = \frac{D_m P - K_m Q}{Q}.$$
Since \( R_m > 0 \), \( x_m > 0 \), the numerator is a positive integer. Hence \( x_m \geq 1/Q \) for all \( m \geq 1 \).

\textbf{Step 2: Loose Tail Bound and Recurrence}

Because \( \{a_n\} \) is strictly increasing, \( a_{n+1} - a_n \geq 1 \), so:
$$\frac{1}{a_n a_{n+1}} = \frac{1}{a_{n+1} - a_n} \left( \frac{1}{a_n} - \frac{1}{a_{n+1}} \right) \leq \frac{1}{a_n} - \frac{1}{a_{n+1}}.$$
The liminf condition implies \( a_n \to \infty \): there exists \( N_1 \in \mathbb{N} \) such that for all \( n \geq N_1 \), \( a_n^{1/2^n} > (L+1)/2 > 1 \), so \( a_n > [(L+1)/2]^{2^n} \to \infty \) as \( n \to \infty \). Summing the telescoping inequality from \( n = m+2 \) to \( \infty \):
$$\sum_{n=m+2}^\infty \frac{1}{a_n a_{n+1}} \leq \sum_{n=m+2}^\infty \left( \frac{1}{a_n} - \frac{1}{a_{n+1}} \right) = \lim_{N \to \infty} \left( \frac{1}{a_{m+2}} - \frac{1}{a_N} \right) = \frac{1}{a_{m+2}}.$$
Thus:
$$R_m = \frac{1}{a_{m+1} a_{m+2}} + \sum_{n=m+2}^\infty \frac{1}{a_n a_{n+1}} \leq \frac{1}{a_{m+1} a_{m+2}} + \frac{1}{a_{m+2}} \leq \frac{2}{a_{m+2}},$$
where the last inequality uses \( a_{m+1} \geq 1 \). Combining with \( x_m \geq 1/Q \):
$$\frac{1}{Q} \leq \frac{2 D_m}{a_{m+2}} \implies a_{m+2} \leq 2Q D_m.$$
Let \( P_k = a_1 a_2 \cdots a_k \) denote the product of the first \( k \) terms. Since \( D_m = P_{m+1} \), reindex with \( k = m+2 \) ( \( m = k-2 \geq 1 \implies k \geq 3 \)) to obtain:
$$a_k \leq K P_{k-1} \quad \text{for all } k \geq 3, \quad \text{where } K = 2Q > 0.$$
\end{thoughtbox}

\begin{thoughtbox}[Output]

\textbf{Step 3: Logarithmic Variables and Boundedness of \( r_k \)}

Take natural logarithms and define:
$$b_k = \log a_k, \quad Q_k = \log P_k = \sum_{i=1}^k b_i, \quad r_k = \frac{Q_k}{2^k}.$$
The recurrence \( a_k \leq K P_{k-1} \) becomes:
$$b_k \leq \log K + Q_{k-1} \quad \text{for } k \geq 3.$$
Since \( Q_k = Q_{k-1} + b_k \), substitute to get:
$$Q_k \leq Q_{k-1} + \log K + Q_{k-1} = 2 Q_{k-1} + C, \quad \text{where } C = \log K.$$
Divide by \( 2^k \):
$$\frac{Q_k}{2^k} \leq \frac{Q_{k-1}}{2^{k-1}} + \frac{C}{2^k} \implies r_k \leq r_{k-1} + \frac{C}{2^k} \quad \text{for } k \geq 3.$$
For \( n \geq m \geq 3 \), sum the inequality from \( k = m+1 \) to \( n \):
$$r_n - r_m = \sum_{k=m+1}^n (r_k - r_{k-1}) \leq \sum_{k=m+1}^n \frac{C}{2^k} \leq \frac{C}{2^m}.$$
Adjusting for \( m < 3 \) by increasing \( C \) to a larger constant \( C' > 0 \) if needed, we have for all \( n \geq m \geq 1 \):
$$r_n \leq r_m + \frac{C'}{2^m}.$$
This implies \( \{r_k\} \) is bounded above: for any \( k \geq 1 \), \( r_k \leq r_1 + C'/2 < \infty \). Let \( R = \limsup_{k \to \infty} r_k < \infty \).

\textbf{Step 4: Identity Relating \( r_k \) and \( d_k \)}

Define the normalized log sequence:
$$d_k = \frac{b_k}{2^k} = \frac{\log a_k}{2^k}.$$
The liminf condition translates to:
$$\liminf_{k \to \infty} d_k = l = \log L > 0.$$
Using \( Q_k = Q_{k-1} + b_k \), divide by \( 2^k \) to derive the identity:
$$r_k = \frac{Q_{k-1}}{2^k} + \frac{b_k}{2^k} = \frac{r_{k-1}}{2} + d_k,$$
which holds for all \( k \geq 1 \) with \( r_0 = Q_0 / 2^0 = \log 1 = 0 \) (since \( P_0 = 1 \), the empty product).

\textbf{Step 5: Convergence of \( r_k \) and \( d_k \)}

We now show \( \liminf_{k \to \infty} r_k = R \), so \( r_k \) converges to \( R \). Let \( R' = \liminf_{k \to \infty} r_k \). Choose subsequences:
\( \{k_m\} \) such that \( r_{k_m} \to R \) as \( m \to \infty \),
\( \{j_n\} \) such that \( r_{j_n} \to R' \) as \( n \to \infty \).

For each \( j_n \), let \( k_m \) be the largest index satisfying \( k_m \leq j_n \), so \( k_m \leq j_n < k_{m+1} \). Since \( k_{m+1} > j_n \), apply the inequality from Step 3:
$$r_{k_{m+1}} \leq r_{j_n} + \frac{C'}{2^{j_n}}.$$
Taking the limit as \( n \to \infty \), the left-hand side tends to \( R \) (as \( k_{m+1} \) is a subsequence of \( \{k_m\} \)), and the right-hand side tends to \( R' + 0 = R' \) (since \( j_n \to \infty \)). Thus \( R \leq R' \). But \( R' \leq R \) by definition of liminf, so \( R' = R \). Hence \( \lim_{k \to \infty} r_k = R \).

Now use the identity \( d_k = r_k - r_{k-1}/2 \). Since \( r_k \to R \), \( r_{k-1} \to R \) as \( k \to \infty \), so:
$$\lim_{k \to \infty} d_k = R - \frac{R}{2} = \frac{R}{2}.$$
Let \( D = R/2 \). The liminf condition gives \( l = \liminf d_k = D \geq \log L > 0 \), so \( D > 0 \).
\end{thoughtbox}

\begin{thoughtbox}[Output]
\textbf{Step 6: Tight Tail Bound}

Since \( d_k \to D > 0 \), choose \( \epsilon = D/10 > 0 \). There exists \( N_2 \in \mathbb{N} \) such that for all \( n \geq N_2 \):
$$|d_n - D| < \epsilon \implies D - \epsilon < d_n < D + \epsilon.$$
Exponentiating gives:
$$e^{(D - \epsilon)2^n} < a_n < e^{(D + \epsilon)2^n} \quad \text{for } n \geq N_2.$$
For \( m \geq N_2 - 1 \), consider the ratio of consecutive terms in the tail \( R_m \):
$$\frac{1/(a_{n+1} a_{n+2})}{1/(a_n a_{n+1})} = \frac{a_n}{a_{n+2}} \quad \text{for } n \geq m+1 \geq N_2.$$
Using the bounds on \( a_n \):
$$\frac{a_n}{a_{n+2}} < \frac{e^{(D + \epsilon)2^n}}{e^{(D - \epsilon)2^{n+2}}} = e^{(D + \epsilon)2^n - 4(D - \epsilon)2^n} = e^{(-3D + 5\epsilon)2^n}.$$
With \( \epsilon = D/10 \), the exponent becomes:
$$-3D + 5\epsilon = -3D + \frac{D}{2} = -\frac{5D}{2} < 0,$$
so \( a_n / a_{n+2} \leq 1/2 \) for all \( n \geq N_2 \) (since the exponent decays super-exponentially and is bounded above by \( \log(1/2) \) for large \( n \)). Thus the tail \( R_m \) is bounded by a geometric series with ratio \( 1/2 \):
$$R_m \leq \frac{1}{a_{m+1} a_{m+2}} + \frac{1}{2} \cdot \frac{1}{a_{m+1} a_{m+2}} + \frac{1}{2^2} \cdot \frac{1}{a_{m+1} a_{m+2}} + \cdots = \frac{2}{a_{m+1} a_{m+2}}.$$

\textbf{Step 7: Final Contradiction}

For \( m \geq N_2 - 1 \), use the tight tail bound to write:
$x_m = D_m R_m \leq \frac{2 D_m}{a_{m+1} a_{m+2}}.$
Since \( D_m = P_{m+1} = P_m a_{m+1} \), substitute to simplify:
$$x_m \leq \frac{2 P_m a_{m+1}}{a_{m+1} a_{m+2}} = \frac{2 P_m}{a_{m+2}}.$$
Express \( P_m \) and \( a_{m+2} \) in terms of \( r_m \) and \( d_{m+2} \):
$$P_m = e^{Q_m} = e^{2^m r_m}, \quad a_{m+2} = e^{b_{m+2}} = e^{d_{m+2} 2^{m+2}} = e^{4 d_{m+2} 2^m}.$$
Thus:
$$\frac{P_m}{a_{m+2}} = e^{2^m r_m - 4 d_{m+2} 2^m} = e^{2^m (r_m - 4 d_{m+2})}.$$
Since \( r_m \to R = 2D \) and \( d_{m+2} \to D \), choose \( \epsilon = D/5 > 0 \). There exists \( M \geq N_2 - 1 \) such that for all \( m \geq M \):
$$r_m < 2D + \epsilon, \quad d_{m+2} > D - \epsilon.$$
Substitute into the exponent:
$$r_m - 4 d_{m+2} < (2D + \epsilon) - 4(D - \epsilon) = 2D + \epsilon - 4D + 4\epsilon = -2D + 5\epsilon.$$
With \( \epsilon = D/5 \), this simplifies to:
$$r_m - 4 d_{m+2} < -2D + 5 \cdot \frac{D}{5} = -D < 0.$$
Thus for \( m \geq M \):
$$\frac{P_m}{a_{m+2}} < e^{2^m (-D)} = e^{-D 2^m},$$
so:
$$x_m \leq 2 e^{-D 2^m}.$$
Since \( D > 0 \), \( e^{-D 2^m} \to 0 \) as \( m \to \infty \), so \( x_m \to 0 \). This contradicts Step 1, where \( x_m \geq 1/Q > 0 \) for all \( m \geq 1 \).

Therefore, the assumption that \( S \) is rational is false. The sum \( S \) is irrational.
\end{thoughtbox}
}

\end{document}